\documentclass[10pt]{article} 
\usepackage[preprint]{tmlr}

\usepackage{amsmath,amsfonts,bm}

\def\eqref#1{equation~\ref{#1}}

\def\1{\bm{1}}

\DeclareMathAlphabet{\mathsfit}{\encodingdefault}{\sfdefault}{m}{sl}
\SetMathAlphabet{\mathsfit}{bold}{\encodingdefault}{\sfdefault}{bx}{n}

\usepackage{hyperref}
\usepackage{url}
\usepackage{tikz}
\usepackage{algorithm2e}
\usepackage{pgfplots}
\usepackage{subcaption}
\usepackage{multirow}
\usepackage{adjustbox}
\usepackage{makecell}

\usetikzlibrary{shapes}
\usetikzlibrary{shapes.symbols}
\usetikzlibrary{shapes.multipart}

\usepackage{pifont}
\newcommand{\cmark}{\ding{51}}%
\newcommand{\xmark}{\ding{55}}%
 
\title{Towards interpretable-by-design deep learning algorithms}

\author{Plamen Angelov$^{+, *}$, Dmitry Kangin$^{+, *}$, Ziyang Zhang$^{+}$\\$^+$ LIRA Centre, School of Computing and Communications, Lancaster University, UK; \\* Equal contribution}

\def\month{MM}  
\def\year{YYYY} 
\def\openreview{\url{https://openreview.net/forum?id=XXXX}} 

\newcommand{\xb}{\mathbf{x}}

\begin{document}

\maketitle

\begin{abstract}
 Most of the existing deep learning (DL) methods rely on parametric tuning and lack explainability. The few methods that claim to offer explainable DL solutions, such as ProtoPNet and xDNN, do require end-to-end training and finetuning. The proposed framework named IDEAL (Interpretable-by-design DEep learning ALgorithms) recasts the standard supervised classification problem into a function of similarity to a set of prototypes derived from the training data, while taking advantage of existing latent spaces of large neural networks forming so-called Foundation Models (FM). This decomposes the overall problem into two inherently connected stages: A) feature extraction (FE), which maps the raw features of the real world problem into a latent space, and B) identifying representative prototypes and decision making based on similarity and association between the query and the prototypes. This addresses the issue of explainability (stage B) while retaining the benefits from the tremendous achievements offered by DL models (e.g., visual transformers, ViT) pre-trained on huge data sets such as IG-3.6B + ImageNet-1K or LVD-142M (stage A). We show that one can turn such DL models into conceptually simpler, explainable-through-prototypes ones. 
 The key findings can be summarized as follows: (1) the proposed models are interpretable through prototypes, mitigating the issue of confounded interpretations, (2) the proposed IDEAL framework circumvents the issue of catastrophic forgetting allowing efficient class-incremental learning, and (3) the proposed IDEAL approach demonstrates that ViT architectures narrow the gap between finetuned and non-finetuned models allowing for transfer learning in a fraction of time \textbf{without} finetuning of the feature space on a target dataset with iterative supervised methods. Furthermore, we show that the proposed approach \textbf{without} finetuning improves the performance on confounded data over finetuned counterparts avoidong overfitting. On a range of datasets (CIFAR-10, CIFAR-100, CalTech101, STL-10, Oxford-IIIT Pet, EuroSAT), we demonstrate, through an extensive set of experiments, how the choice of the latent space, prototype selection, and finetuning of the latent space affect the performance. Building upon this knowledge, we demonstrate that the proposed models have an edge over state-of-the-art baselines in class-incremental learning. Finally, we analyse the interpretations provided by the proposed IDEAL framework, as well as the impact of confounding on the interpretations.
\end{abstract}

\section{Background}

\begin{figure}
  \centering{
\begin{tikzpicture}
\node at (1,4.75) {\includegraphics[width=1.5cm]{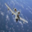}};
\draw[thick,->]  (1,4) -- (1,3.75);

\draw[fill={rgb,255:red,210; green,255; blue,230}] (0,3.25) rectangle (2,3.75);
\node at (1,3.5) {$f_1(\cdot | \boldsymbol{\theta_1})$};
\draw[thick,->]  (1,3.25) -- (1,3); 
 
\draw[fill={rgb,255:red,210; green,255; blue,230}] (0,2.5) rectangle (2,3);
\node at (1,2.75) {$f_2(\cdot | \boldsymbol{\theta_1})$};
\draw[thick,->]  (1,2.5) -- (1,2.25); 

\node at (1, 2) {$\cdots$};

\draw[thick,->]  (1,1.75) -- (1,1.5); 
\draw[fill={rgb,255:red,210; green,255; blue,230}] (0,1.5) rectangle (2,1);
\node at (1,1.25) {$f_n(\cdot | \boldsymbol{\theta_n})$};
\draw[thick,->]  (1,1) -- (1,0.75); 

\node at (1,0.5) {''airplane''};

\draw (2.5,0) -- (2.5,5.5);

\node at (6,3) {\includegraphics[width=6cm]{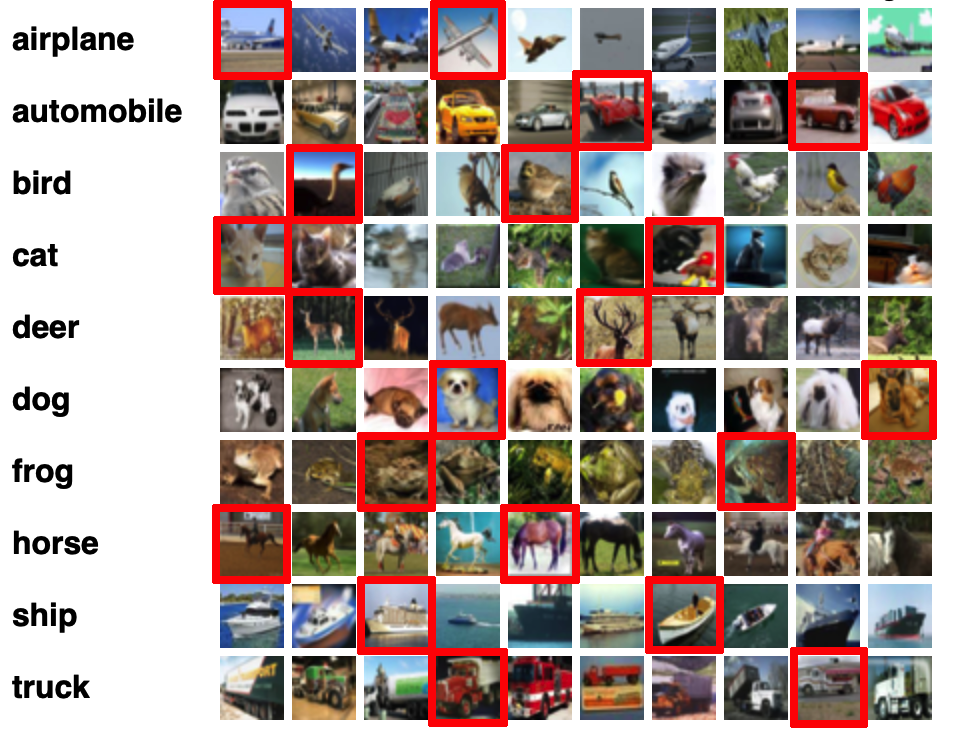}};

\node at(6.5, 5.5) {Prototype selection};

\draw[thick,->] (8.7,2) -- (9.5,2);

\draw[fill={rgb,255:red,210; green,230; blue,255}] (10.25,2) circle (0.75);
\node at (10.25,2) {$d(\cdot, \cdot | \boldsymbol{\theta_1})$};

\draw[thick,->] (10.25,5) -- (10.25,2.75);

\node at (10.25,4.75) {\includegraphics[width=1.5cm]{airplane2.png}};

\draw[thick,->] (11,2) -- (11.25,2);

\node at(13, 2) {$h\left(\left.\left[ \begin{array}{c}
d(\cdot, \mathbf{p}_1)\\
d(\cdot, \mathbf{p}_2)\\
\ldots\\
d(\cdot, \mathbf{p}_n)\\
\end{array}
\right] \right| \boldsymbol{\theta_2}\right)$};
\draw[thick,->] (13,1) -- (13,0.5);
\node at (13,0.25) {''airplane''};

\node at (1,0) {(a)};
\node at (8,0) {(b)};
\end{tikzpicture}
}
\caption{Difference between (a) a standard deep-learning model, and (b) the proposed prototype-based approach, IDEAL; the example is shown for CIFAR-10 dataset (\cite{krizhevskylearning})}
\end{figure}

Deep-learning (DL) models can be formulated as deeply embedded functions of functions (\cite{angelov2019empirical}, \cite{rosenblatt1962principles}): 
\begin{equation}
    \hat{y} (\mathbf{x}) = f_n(\ldots(f_1 (\xb | \boldsymbol{\theta}_1)\ldots)|\boldsymbol{\theta}_n),
    \label{StandardDL}
\end{equation}
where $f_n(\ldots(f_1 (\xb | \boldsymbol{\theta}_1)\ldots)|\boldsymbol{\theta}_n)$ is a layered function of the input $\xb$, which has a generic enough, fixed parameterisation $\boldsymbol{\theta}_{\cdot}$ to predict desirable outputs $\hat{y}$. 

However, this problem statement has the following limitations:

(1) transfer learning typically requires finetuning (\cite{kornblith2019better}) using error back-propagation (EBP) on the target, "downstream" problem/data of interest

(2) such formulation does not depend upon training data, so the contribution of these samples towards the output $\hat{y}$ is unclear, which hinders interpretability; for the interpretable architectures, such as ProtoPNet (\cite{chen2019looks}), finetuning leads to confounding interpretations (\cite{bontempelli2022concept})

(3) finally, for lifelong learning problems, it creates obstacles such as catastrophic forgetting (\cite{parisi2019continual}) 

We follow an alternative solution centered around prototypes inspired by xDNN (\cite{angelov2020towards}), which, at its core, is using a different formulation:
\begin{equation}
     \hat{y} = g (\xb | \boldsymbol{\theta}, \mathbb{P}),
    \label{InterpretableDL}   
\end{equation}
where $\mathbb{P}$ is a set of prototypes. In fact, we consider a more restricted version of function $g(\cdot)$:
\begin{equation}
    \hat{y} = g (\xb | \boldsymbol{\theta}_{\{d,h\}}, \mathbb{X}) = h (d(\xb, \mathbf{p} | \boldsymbol{\theta}_d)| _{\mathbf{p} \in \mathbb{P}} | \boldsymbol{\theta}_h),
\end{equation}
where $d$ is some form of (dis)similarity function (which can include DL FE), $\boldsymbol{\theta}_d$ and $\boldsymbol{\theta}_h$ are parameterisations of functions $d$ and $h$.

The idea takes its roots from cognitive science and the way humans learn, namely using examples of previous observations and experiences (\cite{zeithamova2008dissociable}). Prototype-based models have long been used in different learning systems: $k$ nearest neighbours (\cite{radovanovic2010hubs}); decision trees (\cite{nauta2021neural}); rule-based systems (\cite{angelov2008evolving}); case-based reasoning (\cite{kim2014bayesian}); sparse kernel machines (\cite{tipping1999relevance}). The advantages of prototype-based models has been advocated, for example, in \cite{bien2011prototype}; the first prototypical architecture, learning both distances and prototypes, was proposed in \cite{snell2017prototypical}; more recently, they have been successfully used in \cite{chen2019looks,angelov2020towards} and \cite{wang2023visual}. 

In this paper, we demonstrate the efficiency of the proposed compact, easy to interpret by humans, fast to train and adapt in lifelong learning models that benefit from a latent data space learnt from a generic data set transferred to a different, more specific domain. 

This can be summarised through following contributions:
\begin{itemize}
\item we propose a conceptually simple yet efficient framework, IDEAL, which transforms a given non-interpretable DL model into an interpretable one based on prototypes, derived from the training set.
\item we demonstrate the benefits of the proposed framework on transfer and lifelong learning scenarios: in a fraction of training time, \textbf{without finetuning} of latent features, the proposed models achieve  performance competitive with standard DL techniques.
\item we demonstrate the model's interpretability, on classification and lifelong learning tasks, and show that \textbf{without} finetuning, the resulting models achieves \textbf{better} performance on confounded CUB data comparing to finetuned counterparts (\cite{wah2011caltech, bontempelli2022concept}); yet, for big ViT models the gap decreases. 
\end{itemize}
We apply this generic new IDEAL framework to a set of standard DL architectures such as ViT (\cite{dosovitskiy2020image,singh2022revisiting}), VGG (\cite{simonyan2014very}), ResNet (\cite{he2016deep}) and xDNN (\cite{angelov2020towards}) on a range of data sets such as CIFAR-10, CIFAR-100, CalTech101, EuroSAT, Oxford-IIIT Pet, and STL-10.

\section{Related work}
\paragraph {Explainability} The ever more complicated DL models (\cite{krizhevsky2012imagenet, dosovitskiy2020image}) do not keep pace with the demands for human understandable explainability (\cite{rudin2019stop}). The spread of use of complex DL models prompted pursuit of ways to explain such models. Explainability of deep neural networks is especially important in a number of applications in automotive (\cite{kim2017interpretable}), medical (\cite{ahmad2018interpretable}), Earth observation (\cite{zhang2022interpretable}) problems alongside others. Demand in such models is necessitated by the pursuit of safety (\cite{wei2022safety}), as well as ethical concerns (\cite{peters2022explainable}). 
Some of the pioneering approaches to explaining deep neural networks involve \textit{post hoc} methods; these include saliency models such as saliency map visualisation method (\cite{simonyan2014deep}) as well as Grad-CAM (\cite{selvaraju2017grad}). However, saliency-based explanations may be misleading and not represent the causal relationship between the inputs and outputs (\cite{atrey2019exploratory}), representing instead the biases of the model (\cite{adebayo2018sanity}). An arguably better approach is to construct interpretable-by-design (\textit{ante hoc}) models (\cite{rudin2019stop}). These models could use different principles: interpretable-by-design architectures (\cite{bohle2022b}), which are designed to provide interpretations at every step of the architecture, as well as prototype-based models, which perform decision making as a function of (dis)similarity to existing prototypes (\cite{angelov2020towards}). One of the limitations of the prototype based methods is that they are often still based on non-interpretable similarity metrics; this can be considered an orthogonal open problem which can be addressed by providing interpretable-by-design DL architectures (\cite{bohle2022b}). 

\paragraph{Symbolic and sparse learning machines} The idea of prototype-based machine learning is closely related to the symbolic methods (\cite{newell1959report}), and draws upon the sparse learning machines (\cite{poggio1998sparse}) and case based reasoning (\cite{kim2014bayesian}). The idea of sparse learning machines (\cite{poggio1998sparse}) is to learn a linear (with respect to parameters) model, which is (in general, nonlinearly) dependent on a subset of training data samples (hence, the notion of sparsity). At the centre of many such methods is the kernel trick (\cite{scholkopf2001generalized}), which involves mapping of training and inference data into a space with different inner product within a reproducing Hilbert space (\cite{aronszajn1950theory}). Such models include support vector machines (SVMs) for classification (\cite{boser1992training}) and support vector regression (SVR) models (\cite{smola2004tutorial}) for regression, as well as relevance vector machines (RVMs), which demonstrated improvements in sparsity (\cite{tipping2001sparse}).

\paragraph{Prototype-based models} (\cite{snell2017prototypical}) proposed to use a single prototype per class in a few-shot learning supervised scenario. \cite{li2018deep} proposed prototype-based learning for interpretable case-based reasoning. ProtoPNet (\cite{chen2019looks}) extend this idea to classify an image through dissecting it into a number of patches, which are then compared to  prototypes for decision making using end-to-end supervised training. xDNN (\cite{angelov2020towards}) considers whole images as prototypes resulting from the data density distribution resulting in possibly multiple prototypes per class in a non-iterative online procedure. It does consider, though finetuned on the "downstream"/target data set model for feature extraction for a better performance owing largely to the fact that weak backbone models such as VGG-16 were used. Versions of xDNN offering prototypes in a form of segments (\cite{soares2021explainable}) or even pixels (\cite{zhang2022interpretable}) as prototypes were also reported. The concept of xDNN was used in the end-to-end prototype-based learning method DNC (\cite{wang2023visual}). In contrast to xDNN and DNC, we consider the \textbf{lifelong learning }scenario and investigate the properties of models, trained on generic and \textbf{not finetuned} datasets. 
 
 \paragraph{Large deep-learning classifiers} In contrast to DNC (\cite{wang2023visual}) and ProtoPNet (\cite{chen2019looks}), the proposed framework goes beyond the end-to-end learning concept. Instead, it takes advantage of the feature space of large classifiers such as ResNet (\cite{he2016deep}), VGG (\cite{simonyan2014very}), SWAG-ViT (\cite{singh2022revisiting}), and shows that with carefully selected prototypes one can achieve, on a number of datasets, a performance comparable to end-to-end trained models, in offline and online (lifelong) learning scenarios with or even \textbf{without finetuning and end-to-end learning}, thus very fast and computationally efficient, yet interpretable.

\paragraph{Continual learning} Continual learning models solve different related problems (\cite{van2022three}). \textit{Task-incremental learning} addresses the problem of incrementally learning known tasks, with the intended task explicitly input into the algorithm (\cite{ruvolo2013ella, li2017learning,kirkpatrick2017overcoming}). \textit{Domain-incremental learning} (\cite{wang2022s,lamers2023clustering}) addresses the problem of learning when the domain is changing and the algorithm is not informed about these changes. This includes such issues as \textit{concept drift} when the input data distribution is non-stationary (\cite{widmer1996learning}). Finally, \textit{class-incremental learning} (\cite{yan2021dynamically, wang2022online}) is a problem of ever expanding number of classes of data. In this paper, we only focus on this last problem; however, one can see how the prototype-based approaches could help solve the other two problems by circumventing catastrophic forgetting (\cite{french1999catastrophic}) through incremental update of the prototypes (\cite{baruah2012evolving}). 

\paragraph{Clustering} Critically important for enabling continual learning is to break the iterative nature of the end-to-end learning and within the proposed concept which offers to employ clustering to determine prototypes. Therefore, we are using both online (ELM (\cite{baruah2012evolving}), which is an online version of mean-shift (\cite{comaniciu2002mean})) and offline (\cite{macqueen1967some}) methods.  Although there are a number of online clustering methods, e.g. the stochastic Chinese restaurant process Bayesian non-parametric approach (\cite{aldous2006ecole}), they usually require significant amount of time to run and therefore we did not consider those.

\section{Methodology}

\subsection{Problem statement}
Two different definitions of the problem statement are considered: offline and online (lifelong) learning. In the experimental section, we discuss the implementations of the framework and the experimental results. 

\paragraph{Offline learning} Consider the following optimisation problem:
\begin{equation}
    \arg \min_{\substack{\mathbb{P}=\mathbb{P}(\mathbb{X}),\\ \boldsymbol{\theta}_{\{d,h\}}}} \sum_{(\mathbf{x}, y) \in (\mathbb{X}, \mathbb{Y})} l (h (d (\mathbf{x}, \mathbf{p} | \boldsymbol{\theta}_d)|_{\mathbf{p} \in \mathbb{P}} | \boldsymbol{\theta}_h), y),
\label{equation_offline}
\end{equation}
where $(\mathbb{X}, \mathbb{Y})$ are a tuple of inputs and labels, respectively, and $\mathbb{P}$ is a list of prototypes derived from data $\mathbb{X}$ (e.g., by selecting a set of representative examples or by clustering). 

Brute force optimisation for the problem of selecting a set of representative examples is equivalent to finding a solution of the best subset selection problem, which is an NP-hard problem (\cite{natarajan1995sparse}).  While there are methods solving such subset selection problems in limited cases such as sparse linear regression (\cite{bertsimas2016best}), it still remains computationally inefficient in general case (polynomial complexity is claimed in \cite{zhu2020polynomial})  and/or solving it only in a limited (i.e. linear) setting. 

The common approach is to replace the original optimisation problem (equation (\ref{equation_offline})) with a surrogate one, where the prototypes $\mathbb{P}$ are provided by a data distribution (\cite{angelov2020towards}) or a geometric, e.g. clustering (\cite{wang2023visual}) technique. Then, once the prototypes are selected, the optimisation problem becomes: 
\begin{equation}
    \arg \min_{\boldsymbol{\theta}_{\{d,h\}}} \sum_{(\mathbf{x}, y) \in (\mathbb{X}, \mathbb{Y})} l (h (d (\mathbf{x}, \mathbf{p} | \boldsymbol{\theta}_d)|_{p \in \mathbb{P}} | \boldsymbol{\theta}_h), y).
\label{equation_offline_surrogate}
\end{equation}

\paragraph{Online (lifelong) learning}
Instead of solving a single objective for a fixed dataset, the problem is transformed into a series of optimisation problems for progressively growing set $\mathbb{X}$:
\begin{equation}
    \{ \arg \min_{\boldsymbol{\theta}_{\{d,h\}}} \sum_{(\xb, y) \in (\mathbb{X}_n, \mathbb{Y}_n)} l (h (d (\xb, \mathbf{p} | \theta_d)|_{\mathbf{p} \in \mathbb{P}_n} | \boldsymbol{\theta}_h), y) \}_{n=1}^N, \mathbb{X}_n = \mathbb{X}_{n-1}+\{\xb_n\}, \mathbb{X}_1 = \{\xb_1\}.
\label{equation_online}
\end{equation}
Once the prototypes are found, the problem would only require only light-weight optimisation steps as described in Algorithms \ref{alg:one} and \ref{alg:two}.  

\begin{algorithm}
\caption{Training and testing (offline)}\label{alg:one}
\KwData{Training data $\mathbb{X}=\{\xb_1\ldots \xb_N\}$;   }
\KwResult{Prototype-based classifier $c(\xb | \mathbb{P}, \boldsymbol{\theta})$}
$P \gets \mathrm{FindPrototypes}(\{\xb_1\ldots \xb_N\}$)\;
$\theta \gets \mathrm{SelectParameters} (\mathbb{X}, \mathbb{Y}, \boldsymbol{\theta})$\;
$\hat{\mathbb{Y}_T} \gets \{h (d (\mathbf{x}, \mathbf{p} | \boldsymbol{\theta}_d)|_{\mathbf{p} \in \mathbb{P}} | \boldsymbol{\theta}_h)\}_{\xb \in \mathbb{X}_T} $\;
\end{algorithm}

\begin{algorithm}
\caption{Training and testing (online)}\label{alg:two}
\KwData{Training data $\mathbb{X}=\{\xb_1\ldots \xb_N\}$;  }
\KwResult{Prototype-based classifier $h (d (\xb, \mathbf{p} | \boldsymbol{\theta}_1)|_{p \in \mathbb{P}} | \boldsymbol{\theta}_2)$}
$ \mathbb{P} \gets \{\}$\;
\For{$\{\xb, y\} \in \mathbb{X}$}
{
$\hat{y} = h (d (\xb, \mathbf{p} | \boldsymbol{\theta}_d)|_{\mathbf{p} \in \mathbb{P}} | \boldsymbol{\theta}_h)$\;
$\theta \gets \mathrm{UpdateParameters} (\mathbb{X}, \mathbb{Y}, \boldsymbol{\theta})$\;
$ \mathbb{P} \gets \mathrm{UpdatePrototypes}(\mathbb{P}, \mathbf{x}))$\;
}
\end{algorithm}

\subsection{Prototype selection through clustering} 
\begin{figure}
\centering{
\resizebox{0.6\textwidth}{!}{
\begin{tikzpicture}[domain=-15:15]

\node at (-10.5,4.5) {\includegraphics[width=2cm]{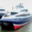}};
\draw[thick,->] (-10.5,3.5) -- (-10.5,2.5);
\draw (-11.5,2.5)[fill=black] rectangle (-9.5,-3);
\node at (-10.5,0) [rotate=90, text=white] {\Large{Black box classifier}};

\draw[thick,->] (-10.5,-3) -- (-10.5,-4);
\node at (-10.5,-4.7) {\huge{''ship''}};

\draw (-9.25,-5.5) -- (-9.25,5.5);

\draw[decoration={brace,raise=2pt},decorate] (-9,5) -- node[above=3pt] {similar} (-2,5); 
\node at (-8,4.7) {''ship''};
\node[cloud,
    fill = gray!30,
    minimum width = 2.25cm,
    minimum height = 2cm] (c) at (-8, 3.5) {};
\node at (-8,3.5) {\includegraphics[width=0.5cm]
{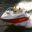}};
\node at (-8.5,3) {\includegraphics[width=0.5cm]
{interpretation_example/39278.png}};
\node at (-7.5,3) {\includegraphics[width=0.5cm]
{interpretation_example/39278.png}};
\node at (-8.5,4) {\includegraphics[width=0.5cm]
{interpretation_example/39278.png}};
\node at (-7.5,4) {\includegraphics[width=0.5cm]
{interpretation_example/39278.png}};
\node at (-8,2.25) {$\ell^2$};
\draw (-8,2) -- (-5.5,1);

\node at (-5.5,4.7) {''ship''};
\node[cloud,
    fill = gray!30,
    minimum width = 2.25cm,
    minimum height = 2cm] (c) at (-5.5, 3.5) {};   
\node at (-5.5,3.5) {\includegraphics[width=0.5cm]{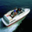}};
\node at (-6,3) {\includegraphics[width=0.5cm]{interpretation_example/15416.png}};
\node at (-6,4) {\includegraphics[width=0.5cm]{interpretation_example/15416.png}};
\node at (-5,3) {\includegraphics[width=0.5cm]{interpretation_example/15416.png}};
\node at (-5,4) {\includegraphics[width=0.5cm]{interpretation_example/15416.png}};
\node at (-5.5,2.25) {$\ell^2$};
\draw (-5.5,2) -- (-5.5,1);

\node at (-3,4.7) {''ship''};
\node[cloud,
    fill = gray!30,
    minimum width = 2.25cm,
    minimum height = 2cm] (c) at (-3, 3.5) {};   
\node at (-3,3.5) {\includegraphics[width=0.5cm]{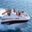}};
\node at (-3.5,4) {\includegraphics[width=0.5cm]{interpretation_example/19357.png}};
\node at (-2.5,4) {\includegraphics[width=0.5cm]{interpretation_example/19357.png}};
\node at (-3.5,3) {\includegraphics[width=0.5cm]{interpretation_example/19357.png}};
\node at (-2.5,3) {\includegraphics[width=0.5cm]{interpretation_example/19357.png}};

\node at (-3,2.25) {$\ell^2$};
\draw (-3,2) -- (-5.5,1);

\node at (-7.5,0) {\Huge{?}};
\node at (-8.5,1) {\Huge{$\mathcal{\cdot}$}};
\node at (-8.5,0) {\Huge{$\mathcal{\cdot}$}};
\node at (-8.5,-1) {\Huge{$\mathcal{\cdot}$}};
\node at (-3.5,0) {\huge{$\longrightarrow$ ''ship''}};

\node at (-5.5,0) {\includegraphics[width=2cm]{interpretation_example/1.png}};
\node[cloud,
    fill = gray!30,
    minimum width = 2.25cm,
    minimum height = 2cm] (c) at (-8, -3.5) {};   
\node at (-8,-3.5) {\includegraphics[width=0.5cm]{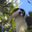}};  
\node at (-8.5,-3) {\includegraphics[width=0.5cm]{interpretation_example/19287.png}};  
\node at (-8.5,-4) {\includegraphics[width=0.5cm]{interpretation_example/19287.png}};  
\node at (-7.5,-3) {\includegraphics[width=0.5cm]{interpretation_example/19287.png}};  
\node at (-7.5,-4) {\includegraphics[width=0.5cm]{interpretation_example/19287.png}};
\draw (-8,-2) -- (-5.5,-1);
\node at (-8,-2.25) {$\ell^2$};
\node at (-8,-4.7) {''bird''};

\node[cloud,
    fill = gray!30,
    minimum width = 2.25cm,
    minimum height = 2cm] (c) at (-5.5, -3.5) {};  
\node at (-5.5,-3.5) {\includegraphics[width=0.5cm]{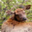}};
\node at (-6,-4) {\includegraphics[width=0.5cm]{interpretation_example/28509.png}};
\node at (-6,-3) {\includegraphics[width=0.5cm]{interpretation_example/28509.png}};
\node at (-5,-4) {\includegraphics[width=0.5cm]{interpretation_example/28509.png}};
\node at (-5,-3) {\includegraphics[width=0.5cm]{interpretation_example/28509.png}};
\draw (-5.5,-2) -- (-5.5,-1);
\node at (-5.5,-2.25) {$\ell^2$};
\node at (-5.5,-4.7) {''deer''};

\node[cloud,
    fill = gray!30,
    minimum width = 2.25cm,
    minimum height = 2cm] (c) at (-3, -3.5) {}; 
\node at (-3,-3.5) {\includegraphics[width=0.5cm]{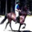}};
\node at (-3.5,-3) {\includegraphics[width=0.5cm]{interpretation_example/14561.png}};
\node at (-3.5,-4) {\includegraphics[width=0.5cm]{interpretation_example/14561.png}};
\node at (-2.5,-3) {\includegraphics[width=0.5cm]{interpretation_example/14561.png}};
\node at (-2.5,-4) {\includegraphics[width=0.5cm]{interpretation_example/14561.png}};
\node at (-3,-2.25) {$\ell^2$};
\node at (-3,-4.7) {''horse''};
\draw (-3,-2) -- (-5.5,-1);
\draw[decoration={brace,raise=2pt,mirror},decorate] (-9,-5) -- node[below=3pt]{dissimilar} (-2,-5);

\draw (-1.75,-5.5) -- (-1.75,5.5);

\draw[decoration={brace,raise=2pt},decorate] (-1.5,5) -- node[above=3pt] {similar} (5.5,5); 
\node at (-0.5,4.7) {''ship''};
\node at (-0.5,3.5) {\includegraphics[width=2cm]{interpretation_example/39278.png}};
\node at (-0.5,2.25) {$\ell^2$: $28.145$};
\draw (-0.5,2) -- (2,1);

\node at (2,4.7) {''ship''};
\node at (2,3.5) {\includegraphics[width=2cm]{interpretation_example/15416.png}};
\node at (2,2.25) {$\ell^2$: $28.272$};
\draw (2,2) -- (2,1);

\node at (4.5,4.7) {''ship''};
\node at (4.5,3.5) {\includegraphics[width=2cm]{interpretation_example/19357.png}};
\node at (4.5,2.25) {$\ell^2$: $28.735$};
\draw (4.5,2) -- (2,1);

\node at (0,0) {\Huge{?}};
\node at (-1,1) {\Huge{$\mathcal{\cdot}$}};
\node at (-1,0) {\Huge{$\mathcal{\cdot}$}};
\node at (-1,-1) {\Huge{$\mathcal{\cdot}$}};
\node at (4,0) {\huge{$\longrightarrow$ ''ship''}};

\node at (2,0) {\includegraphics[width=2cm]{interpretation_example/1.png}};

\node at (-0.5,-3.5) {\includegraphics[width=2cm]{interpretation_example/19287.png}};
\draw (-0.5,-2) -- (2,-1);
\node at (-0.5,-2.25) {$\ell^2$: $52.952$};
\node at (-0.5,-4.7) {''bird''};

\node at (2,-3.5) {\includegraphics[width=2cm]{interpretation_example/28509.png}};
\draw (2,-2) -- (2,-1);
\node at (2,-2.25) {$\ell^2$: $52.960$};
\node at (2,-4.7) {''deer''};

\node at (4.5,-3.5) {\includegraphics[width=2cm]{interpretation_example/14561.png}};
\node at (4.5,-2.25) {$\ell^2$: $52.960$};
\node at (4.5,-4.7) {''horse''};
\draw (4.5,-2) -- (2,-1);
\draw[decoration={brace,raise=2pt,mirror},decorate] (-1.5,-5) -- node[below=3pt]{dissimilar} (5.5,-5);
\end{tikzpicture}
}
}
\caption{Black-box, $k$-means centroid prototypes, and interpretable prototypes (CIFAR-10)}
\label{black_box_centroid_interpretable}
\end{figure}

Selection of prototypes through many standard methods of clustering, such as $k$-means (\cite{steinhaus1956division}), is used by methods such as (\cite{zhang2022interpretable}), DCN (\cite{wang2023visual}), however, has one serious limitation: they utilise the averaging of cluster values, so the prototypes $\mathbb{P}$ do not, in general, belong to the original training dataset $\mathbb{X}$. It is still possible, however, to attribute the prediction to the set of the cluster members. This can create, as we show in the experimental section, a trade-off between interpretability and performance (see Section \ref{classification_section}). The available options are summarised in Figure \ref{black_box_centroid_interpretable}. Standard \textit{black-box} classifiers do not offer interpretability through prototypes. Prototypes, selected through $k$-means, are non-interpretable by their own account as discussed above; however, it is possible to attribute such similarity to the members of the clusters. Finally, one can select real prototypes as cluster centroids; this way it is possible to attribute the decision to a number of real image prototypes ranked by their similarity to the query image.

\section{Experiments}
\begin{figure}
\centering
{
\resizebox{0.8\textwidth}{!}{
\begin{tikzpicture}[domain=0:5]
\draw (-3,-1.25) rectangle (9, 1.75);
\node at (-2,0.25)
{\includegraphics[width=1.5cm]{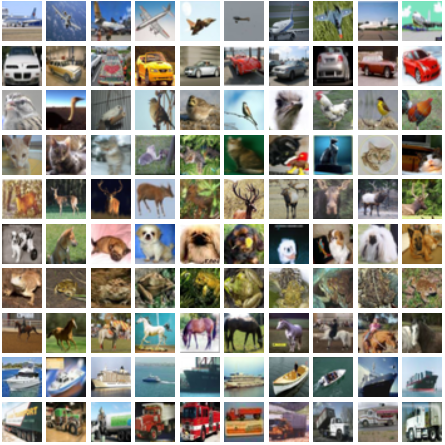}};
\node at (-2, -0.9) {\textbf{\textbf{\begin{tabular}{c} \small{Downstream}\\ \small{dataset} \end{tabular}}}};

\draw[->] (-1,0.25) -- (0.25,0.25);
\node at (1.95, 1.6) {\textbf{Backprop.}};
\draw[->] (2.95,1.45) -- (0.9,1.45);

\draw[->] (0.75, 1.25) -- (1.05,1.25);
\draw[->] (1.35, 1.25) -- (1.8,1.25);
\draw[->] (1.95, 1.25) -- (2.55,1.25);
\draw[->] (2.75, 1.25) -- (3.0,1.25);

\draw[->] (1.35, 1.25) -- (1.8,0.5);
\draw[->] (1.95, 1.25) -- (2.55,0.5);

\draw[->] (1.35, 0.5) -- (1.8,1.25);
\draw[->] (1.95, 0.5) -- (2.55,1.25);

\draw[->] (1.35, -0.25) -- (1.8,1.25);
\draw[->] (1.95, -0.25) -- (2.55,1.25);

\draw[->] (1.35, 1.25) -- (1.8,-0.25);
\draw[->] (1.95, 1.25) -- (2.55,-0.25);

\draw[fill={rgb,255:red,210; green,230; blue,255}] (1.2,1.25) circle (0.15);
\draw[fill={rgb,255:red,210; green,230; blue,255}] (1.95,1.25) circle (0.15);
\draw[fill={rgb,255:red,210; green,230; blue,255}] (2.7,1.25) circle (0.15);

\draw[->] (0.75, 0.5) -- (1.05,0.5);
\draw[->] (1.35, 0.5) -- (1.8,0.5);
\draw[->] (1.95, 0.5) -- (2.55,0.5);
\draw[->] (2.75, 0.5) -- (3.0,0.5);

\draw[->] (1.35, 0.5) -- (1.8,-0.25);
\draw[->] (1.95, 0.5) -- (2.55,-0.25);

\draw[->] (1.35, -0.25) -- (1.8,0.5);
\draw[->] (1.95, -0.25) -- (2.55,0.5);

\draw[fill={rgb,255:red,210; green,230; blue,255}] (1.2,0.5) circle (0.15);
\draw[fill={rgb,255:red,210; green,230; blue,255}] (1.95,0.5) circle (0.15);
\draw[fill={rgb,255:red,210; green,230; blue,255}] (2.7,0.5) circle (0.15);

\draw[->] (0.75, -0.25) -- (1.05,-0.25);
\draw[->] (1.35, -0.25) -- (1.8,-0.25);
\draw[->] (1.95, -0.25) -- (2.55,-0.25);
\draw[->] (2.75, -0.25) -- (3.0,-0.25);

\draw[fill={rgb,255:red,210; green,230; blue,255}] (1.2,-0.25) circle (0.15);
\draw[fill={rgb,255:red,210; green,230; blue,255}] (1.95,-0.25) circle (0.15);
\draw[fill={rgb,255:red,210; green,230; blue,255}] (2.7,-0.25) circle (0.15);

\node at (1.95, -0.75) {\textbf{Finetuning}};

 \draw[->] (3.6,0.25) -- (5.6,0.25);

\node at (7,1.45)
{\includegraphics[width=0.5cm]{airplane2.png}};

\draw[->] (6.25, 1.20) -- (6.25,1.0);
\draw[->] (7, 1.20) -- (7,1.0);
\draw[->] (7.75, 1.20) -- (7.75,1.0);

\draw[->] (6.25, 0.85) -- (6.25,0.35);
\draw[->] (6.25, 0.85) -- (7,0.35);
\draw[->] (6.25, 0.85) -- (7.75,0.35);

\draw[->] (7, 0.85) -- (6.25,0.35);
\draw[->] (7, 0.85) -- (7,0.35);
\draw[->] (7, 0.85) -- (7.75,0.35);

\draw[->] (7.75, 0.85) -- (6.25,0.35);
\draw[->] (7.75, 0.85) -- (7,0.35);
\draw[->] (7.75, 0.85) -- (7.75,0.35);

\draw[fill={rgb,255:red,210; green,230; blue,255}] (6.25,0.85) circle (0.15);
\draw[fill={rgb,255:red,210; green,230; blue,255}] (7,0.85) circle (0.15);
\draw[fill={rgb,255:red,210; green,230; blue,255}] (7.75,0.85) circle (0.15);

\draw[->] (6.25, 0.2) -- (6.25,-0.3);
\draw[->] (6.25, 0.2) -- (7,-0.3);
\draw[->] (6.25, 0.2) -- (7.75,-0.3);

\draw[->] (7, 0.2) -- (6.25,-0.3);
\draw[->] (7, 0.2) -- (7,-0.3);
\draw[->] (7, 0.2) -- (7.75,-0.3);

\draw[->] (7.75, 0.2) -- (6.25,-0.3);
\draw[->] (7.75, 0.2) -- (7,-0.3);
\draw[->] (7.75, 0.2) -- (7.75,-0.3);

\draw[fill={rgb,255:red,210; green,230; blue,255}] (6.25,0.2) circle (0.15);
\draw[fill={rgb,255:red,210; green,230; blue,255}] (7,0.2) circle (0.15);
\draw[fill={rgb,255:red,210; green,230; blue,255}] (7.75,0.2) circle (0.15);

\draw[->] (6.25, -0.45) -- (6.25,-0.75);
\draw[->] (7, -0.45) -- (7,-0.75);
\draw[->] (7.75, -0.45) -- (7.75,-0.75);

\draw[fill={rgb,255:red,210; green,230; blue,255}] (6.25,-0.45) circle (0.15);
\draw[fill={rgb,255:red,210; green,230; blue,255}] (7,-0.45) circle (0.15);
\draw[fill={rgb,255:red,210; green,230; blue,255}] (7.75,-0.45) circle (0.15);

\node at (7, -1) {\textbf{Classification}};
\draw (-3,-1.5) -- (9,-1.5);
\end{tikzpicture}
}
}
\centering
{
\resizebox{0.8\textwidth}{!}{
\begin{tikzpicture}[domain=0:5]
\draw (-2.5,-1.45) rectangle (10.25, 1.75);
\node at (-1.5,0.5)
{\includegraphics[width=1.5cm]{cifar10.png}};
\node at (-1.5, -0.9) {\textbf{\textbf{\begin{tabular}{c} \small{Downstream}\\ \small{dataset} \end{tabular}}}};

\draw[->] (-0.75,0.5) -- (-0.3,0.5);

\draw[->] (-0.25, 1.25) -- (0.05,1.25);
\draw[->] (0.35, 1.25) -- (0.8,1.25);
\draw[->] (0.95, 1.25) -- (1.55,1.25);
\draw[->] (1.75, 1.25) -- (2.0,1.25);

\draw[->] (0.35, 1.25) -- (0.8,0.5);
\draw[->] (0.95, 1.25) -- (1.55,0.5);

\draw[->] (0.35, 0.5) -- (0.8,1.25);
\draw[->] (0.95, 0.5) -- (1.55,1.25);

\draw[->] (0.35, -0.25) -- (0.8,1.25);
\draw[->] (0.95, -0.25) -- (1.55,1.25);

\draw[->] (0.35, 1.25) -- (0.8,-0.25);
\draw[->] (0.95, 1.25) -- (1.55,-0.25);

\draw[fill={rgb,255:red,210; green,230; blue,255}] (0.2,1.25) circle (0.15);
\draw[fill={rgb,255:red,210; green,230; blue,255}] (0.95,1.25) circle (0.15);
\draw[fill={rgb,255:red,210; green,230; blue,255}] (1.7,1.25) circle (0.15);

\draw[->] (-0.25, 0.5) -- (0.05,0.5);
\draw[->] (0.35, 0.5) -- (0.8,0.5);
\draw[->] (0.95, 0.5) -- (1.55,0.5);
\draw[->] (1.75, 0.5) -- (2.0,0.5);

\draw[->] (0.35, 0.5) -- (0.8,-0.25);
\draw[->] (0.95, 0.5) -- (1.55,-0.25);

\draw[->] (0.35, -0.25) -- (0.8,0.5);
\draw[->] (0.95, -0.25) -- (1.55,0.5);

\draw[fill={rgb,255:red,210; green,230; blue,255}] (0.2,0.5) circle (0.15);
\draw[fill={rgb,255:red,210; green,230; blue,255}] (0.95,0.5) circle (0.15);
\draw[fill={rgb,255:red,210; green,230; blue,255}] (1.7,0.5) circle (0.15);

\draw[->] (-0.25, -0.25) -- (0.05,-0.25);
\draw[->] (0.35, -0.25) -- (0.8,-0.25);
\draw[->] (0.95, -0.25) -- (1.55,-0.25);
\draw[->] (1.75, -0.25) -- (2.0,-0.25);

\draw[fill={rgb,255:red,210; green,230; blue,255}] (0.2,-0.25) circle (0.15);
\draw[fill={rgb,255:red,210; green,230; blue,255}] (0.95,-0.25) circle (0.15);
\draw[fill={rgb,255:red,210; green,230; blue,255}] (1.7,-0.25) circle (0.15);

\node at (1.25, -1) {\textbf{Feature extraction}};

\draw (2.05, -0.5) rectangle (2.35, 1.5);
\node[rotate=90] at (2.2, 0.5) {\textbf{Features}};

\draw[->] (2.3,0.5) -- (2.75,0.5);

\begin{scope}
\clip (4.5, 0.5) -- (6.15,0.5) arc (0:360:1.75) -- cycle;
\node [cloud, cloud puffs=12, draw, minimum width=3.3cm, minimum height=2.5cm,fill={rgb,255:red,255; green,210; blue,230}]
    at (4.5, 0.5) {};
\end{scope}
\begin{scope}
\clip (4.5, 0.5) -- (6.15,0.5) arc (0:220:1.75) -- cycle;
\node [cloud, cloud puffs=12, draw, minimum width=3.3cm, minimum height=2.5cm,fill={rgb,255:red,210; green,230; blue,255}]
    at (4.5, 0.5) {};
\end{scope}
\begin{scope}
\clip (4.5, 0.5) -- (6.15,0.5) arc (0:100:1.75) -- cycle;
\node [cloud, cloud puffs=12, draw, minimum width=3.3cm, minimum height=2.5cm,fill={rgb,255:red,230; green,255; blue,210}]
    at (4.5, 0.5) {};
\end{scope}

\node[star,star point ratio=2.25,minimum size=6pt,
          inner sep=0pt,draw=black,solid,fill=red] at (4.75, 0) {};

\node[circle,minimum size=3pt,
          inner sep=0pt,draw=black,solid,fill=red] at (4.3, 0.15) {};

\node[circle,minimum size=3pt,
          inner sep=0pt,draw=black,solid,fill=red] at (3.6, -0.35) {};

\node[circle,minimum size=3pt,
          inner sep=0pt,draw=black,solid,fill=red] at (5.0, -0.25) {};

\node[circle,minimum size=3pt,
          inner sep=0pt,draw=black,solid,fill=red] at (5.0, 0.35) {};
          
\node[circle,minimum size=3pt,
          inner sep=0pt,draw=black,solid,fill=red] at (5.6, 0.25) {};
          
\node[star,star point ratio=2.25,minimum size=6pt,
          inner sep=0pt,draw=black,solid,fill=blue] at (3.75, 0.75) {};

\node[circle,minimum size=3pt,
          inner sep=0pt,draw=black,solid,fill=blue] at (4.3, 0.65) {};

\node[circle,minimum size=3pt,
          inner sep=0pt,draw=black,solid,fill=blue] at (3.2, 0.25) {};

\node[circle,minimum size=3pt,
          inner sep=0pt,draw=black,solid,fill=blue] at (4.0, 1.25) {};

\node[circle,minimum size=3pt,
          inner sep=0pt,draw=black,solid,fill=blue] at (4.1, 0.85) {};
          
\node[circle,minimum size=3pt,
          inner sep=0pt,draw=black,solid,fill=blue] at (3.6, 1.05) {};

\node[star,star point ratio=2.25,minimum size=6pt,
          inner sep=0pt,draw=black,solid,fill=green] at (5.0, 0.85) {};

\node[circle,minimum size=3pt,
          inner sep=0pt,draw=black,solid,fill=green] at (5.3, 0.65) {};

\node[circle,minimum size=3pt,
          inner sep=0pt,draw=black,solid,fill=green] at (4.7, 0.65) {};

\node[circle,minimum size=3pt,
          inner sep=0pt,draw=black,solid,fill=green] at (5.0, 1.25) {};

\node[circle,minimum size=3pt,
          inner sep=0pt,draw=black,solid,fill=green] at (5.7, 0.85) {};
          
\node[circle,minimum size=3pt,
          inner sep=0pt,draw=black,solid,fill=green] at (4.6, 1.05) {};

\node at (4.5, -1.025) {\textbf{Clustering}};

\draw[->] (6.15, 0.5) -- (7,0.5);

\begin{scope}
\clip (8.5, 0.5) -- (10.15,0.5) arc (0:360:1.75) -- cycle;
\node [cloud, cloud puffs=12, draw, minimum width=3.3cm, minimum height=2.5cm,fill={rgb,255:red,255; green,210; blue,230}]
    at (8.5, 0.5) {};
\end{scope}
\begin{scope}
\clip (8.5, 0.5) -- (10.15,0.5) arc (0:220:1.75) -- cycle;
\node [cloud, cloud puffs=12, draw, minimum width=3.3cm, minimum height=2.5cm,fill={rgb,255:red,210; green,230; blue,255}]
    at (8.5, 0.5) {};
\end{scope}
\begin{scope}
\clip (8.5, 0.5) -- (10.15,0.5) arc (0:100:1.75) -- cycle;
\node [cloud, cloud puffs=12, draw, minimum width=3.3cm, minimum height=2.5cm,fill={rgb,255:red,230; green,255; blue,210}]
    at (8.5, 0.5) {};
\end{scope}

\node[star,star point ratio=2.25,minimum size=6pt,
          inner sep=0pt,draw=black,solid,fill=red] at (8.75, 0) {};
          
\node[star,star point ratio=2.25,minimum size=6pt,
          inner sep=0pt,draw=black,solid,fill=blue] at (7.75, 0.75) {};

\node[star,star point ratio=2.25,minimum size=6pt,
          inner sep=0pt,draw=black,solid,fill=green] at (9, 0.85) {};

\node at (8.5, -1.025) {\textbf{\begin{tabular}{c} \small{Classification}\\ \small{Interpretation} \end{tabular}}};

\node at (6.5,1.45)
{\includegraphics[width=0.5cm]{airplane2.png}};
\draw [->] (6.75, 1.45) -- (6.95,1.45) -- (7.25,0.55);

\node[circle,minimum size=3pt,
          inner sep=0pt,draw=black,solid,fill=gray] at (7.25,0.55) {};

\draw [<->] (7.30,0.58) -- (7.66, 0.72);

\end{tikzpicture}
}
}
\centering
{
\resizebox{0.8\textwidth}{!}{
\begin{tikzpicture}[domain=0:5]
\draw (-2.5,-1.45) rectangle (12.7, 1.75);
\node at (-1.5,0.5)
{\includegraphics[width=1.5cm]{cifar10.png}};

\node at (-1.5,0.5)
{\includegraphics[width=1.5cm]{cifar10.png}};
\node at (-1.5, -0.9) {\textbf{\textbf{\begin{tabular}{c} \small{Downstream}\\ \small{dataset} \end{tabular}}}};

\draw[->] (-0.75,0.5) -- (-0.3,0.5);

\node at (0.95, 1.6) {\textbf{Backprop.}};
\draw[->] (1.95,1.45) -- (-0.1,1.45);

\draw[->] (-0.25, 1.25) -- (0.05,1.25);
\draw[->] (0.35, 1.25) -- (0.8,1.25);
\draw[->] (0.95, 1.25) -- (1.55,1.25);
\draw[->] (1.75, 1.25) -- (2.0,1.25);

\draw[->] (0.35, 1.25) -- (0.8,0.5);
\draw[->] (0.95, 1.25) -- (1.55,0.5);

\draw[->] (0.35, 0.5) -- (0.8,1.25);
\draw[->] (0.95, 0.5) -- (1.55,1.25);

\draw[->] (0.35, -0.25) -- (0.8,1.25);
\draw[->] (0.95, -0.25) -- (1.55,1.25);

\draw[->] (0.35, 1.25) -- (0.8,-0.25);
\draw[->] (0.95, 1.25) -- (1.55,-0.25);

\draw[fill={rgb,255:red,210; green,230; blue,255}] (0.2,1.25) circle (0.15);
\draw[fill={rgb,255:red,210; green,230; blue,255}] (0.95,1.25) circle (0.15);
\draw[fill={rgb,255:red,210; green,230; blue,255}] (1.7,1.25) circle (0.15);

\draw[->] (-0.25, 0.5) -- (0.05,0.5);
\draw[->] (0.35, 0.5) -- (0.8,0.5);
\draw[->] (0.95, 0.5) -- (1.55,0.5);
\draw[->] (1.75, 0.5) -- (2.0,0.5);

\draw[->] (0.35, 0.5) -- (0.8,-0.25);
\draw[->] (0.95, 0.5) -- (1.55,-0.25);

\draw[->] (0.35, -0.25) -- (0.8,0.5);
\draw[->] (0.95, -0.25) -- (1.55,0.5);

\draw[fill={rgb,255:red,210; green,230; blue,255}] (0.2,0.5) circle (0.15);
\draw[fill={rgb,255:red,210; green,230; blue,255}] (0.95,0.5) circle (0.15);
\draw[fill={rgb,255:red,210; green,230; blue,255}] (1.7,0.5) circle (0.15);

\draw[->] (-0.25, -0.25) -- (0.05,-0.25);
\draw[->] (0.35, -0.25) -- (0.8,-0.25);
\draw[->] (0.95, -0.25) -- (1.55,-0.25);
\draw[->] (1.75, -0.25) -- (2.0,-0.25);

\draw[fill={rgb,255:red,210; green,230; blue,255}] (0.2,-0.25) circle (0.15);
\draw[fill={rgb,255:red,210; green,230; blue,255}] (0.95,-0.25) circle (0.15);
\draw[fill={rgb,255:red,210; green,230; blue,255}] (1.7,-0.25) circle (0.15);

\node at (0.95, -1) {\textbf{Finetuning}};

\draw[->] (2.1,0.5) -- (2.4,0.5);

\draw[->] (2.45, 1.25) -- (2.75,1.25);
\draw[->] (3.05, 1.25) -- (3.5,1.25);
\draw[->] (3.65, 1.25) -- (4.25,1.25);
\draw[->] (4.55, 1.25) -- (4.7,1.25);

\draw[->] (3.05, 1.25) -- (3.5,0.5);
\draw[->] (3.65, 1.25) -- (4.25,0.5);

\draw[->] (3.05, 0.5) -- (3.5,1.25);
\draw[->] (3.65, 0.5) -- (4.25,1.25);

\draw[->] (3.05, -0.25) -- (3.5,1.25);
\draw[->] (3.65, -0.25) -- (4.25,1.25);

\draw[->] (3.05, 1.25) -- (3.5,-0.25);
\draw[->] (3.65, 1.25) -- (4.25,-0.25);

\draw[fill={rgb,255:red,210; green,230; blue,255}] (2.9,1.25) circle (0.15);
\draw[fill={rgb,255:red,210; green,230; blue,255}] (3.65,1.25) circle (0.15);
\draw[fill={rgb,255:red,210; green,230; blue,255}] (4.4,1.25) circle (0.15);

\draw[->] (2.45, 0.5) -- (2.75,0.5);
\draw[->] (3.05, 0.5) -- (3.5,0.5);
\draw[->] (3.65, 0.5) -- (4.25,0.5);
\draw[->] (4.55, 0.5) -- (4.7,0.5);

\draw[->] (3.05, 0.5) -- (3.5,-0.25);
\draw[->] (3.65, 0.5) -- (4.25,-0.25);

\draw[->] (3.05, -0.25) -- (3.5,0.5);
\draw[->] (3.65, -0.25) -- (4.25,0.5);

\draw[fill={rgb,255:red,210; green,230; blue,255}] (2.9,0.5) circle (0.15);
\draw[fill={rgb,255:red,210; green,230; blue,255}] (3.65,0.5) circle (0.15);
\draw[fill={rgb,255:red,210; green,230; blue,255}] (4.4,0.5) circle (0.15);

\draw[->] (2.45, -0.25) -- (2.75,-0.25);
\draw[->] (3.05, -0.25) -- (3.5,-0.25);
\draw[->] (3.65, -0.25) -- (4.25,-0.25);
\draw[->] (4.55, -0.25) -- (4.7,-0.25);

\draw[fill={rgb,255:red,210; green,230; blue,255}] (2.9,-0.25) circle (0.15);
\draw[fill={rgb,255:red,210; green,230; blue,255}] (3.65,-0.25) circle (0.15);
\draw[fill={rgb,255:red,210; green,230; blue,255}] (4.4,-0.25) circle (0.15);

\node at (3.85, -1) {\textbf{Feature extraction}};

\draw (4.75, -0.5) rectangle (5.05, 1.5);
\node[rotate=90] at (4.9, 0.5) {\textbf{Features}};

\draw[->] (5.05,0.5) -- (5.35,0.5);

\begin{scope}
\clip (7, 0.5) -- (8.65,0.5) arc (0:360:1.75) -- cycle;
\node [cloud, cloud puffs=12, draw, minimum width=3.3cm, minimum height=2.5cm,fill={rgb,255:red,255; green,210; blue,230}]
    at (7, 0.5) {};
\end{scope}
\begin{scope}
\clip (7, 0.5) -- (8.65,0.5) arc (0:220:1.75) -- cycle;
\node [cloud, cloud puffs=12, draw, minimum width=3.3cm, minimum height=2.5cm,fill={rgb,255:red,210; green,230; blue,255}]
    at (7, 0.5) {};
\end{scope}
\begin{scope}
\clip (7, 0.5) -- (8.65,0.5) arc (0:100:1.75) -- cycle;
\node [cloud, cloud puffs=12, draw, minimum width=3.3cm, minimum height=2.5cm,fill={rgb,255:red,230; green,255; blue,210}]
    at (7, 0.5) {};
\end{scope}

\node[star,star point ratio=2.25,minimum size=6pt,
          inner sep=0pt,draw=black,solid,fill=red] at (7.25, 0) {};

\node[circle,minimum size=3pt,
          inner sep=0pt,draw=black,solid,fill=red] at (6.8, 0.15) {};

\node[circle,minimum size=3pt,
          inner sep=0pt,draw=black,solid,fill=red] at (6.1, -0.35) {};

\node[circle,minimum size=3pt,
          inner sep=0pt,draw=black,solid,fill=red] at (7.5, -0.25) {};

\node[circle,minimum size=3pt,
          inner sep=0pt,draw=black,solid,fill=red] at (7.5, 0.35) {};
          
\node[circle,minimum size=3pt,
          inner sep=0pt,draw=black,solid,fill=red] at (8.1, 0.25) {};
          
\node[star,star point ratio=2.25,minimum size=6pt,
          inner sep=0pt,draw=black,solid,fill=blue] at (6.25, 0.75) {};

\node[circle,minimum size=3pt,
          inner sep=0pt,draw=black,solid,fill=blue] at (6.8, 0.65) {};

\node[circle,minimum size=3pt,
          inner sep=0pt,draw=black,solid,fill=blue] at (5.7, 0.25) {};

\node[circle,minimum size=3pt,
          inner sep=0pt,draw=black,solid,fill=blue] at (6.5, 1.25) {};

\node[circle,minimum size=3pt,
          inner sep=0pt,draw=black,solid,fill=blue] at (6.6, 0.85) {};
          
\node[circle,minimum size=3pt,
          inner sep=0pt,draw=black,solid,fill=blue] at (6.1, 1.05) {};

\node[star,star point ratio=2.25,minimum size=6pt,
          inner sep=0pt,draw=black,solid,fill=green] at (7.5, 0.85) {};

\node[circle,minimum size=3pt,
          inner sep=0pt,draw=black,solid,fill=green] at (7.8, 0.65) {};

\node[circle,minimum size=3pt,
          inner sep=0pt,draw=black,solid,fill=green] at (7.2, 0.65) {};

\node[circle,minimum size=3pt,
          inner sep=0pt,draw=black,solid,fill=green] at (7.5, 1.25) {};

\node[circle,minimum size=3pt,
          inner sep=0pt,draw=black,solid,fill=green] at (8.2, 0.85) {};
          
\node[circle,minimum size=3pt,
          inner sep=0pt,draw=black,solid,fill=green] at (7.1, 1.05) {};

\node at (7, -1.025) {\textbf{Clustering}};

\begin{scope}
\clip (11, 0.5) -- (12.65,0.5) arc (0:360:1.75) -- cycle;
\node [cloud, cloud puffs=12, draw, minimum width=3.3cm, minimum height=2.5cm,fill={rgb,255:red,255; green,210; blue,230}]
    at (11, 0.5) {};
\end{scope}
\begin{scope}
\clip (11, 0.5) -- (12.65,0.5) arc (0:220:1.75) -- cycle;
\node [cloud, cloud puffs=12, draw, minimum width=3.3cm, minimum height=2.5cm,fill={rgb,255:red,210; green,230; blue,255}]
    at (11, 0.5) {};
\end{scope}
\begin{scope}
\clip (11, 0.5) -- (12.65,0.5) arc (0:100:1.75) -- cycle;
\node [cloud, cloud puffs=12, draw, minimum width=3.3cm, minimum height=2.5cm,fill={rgb,255:red,230; green,255; blue,210}]
    at (11, 0.5) {};
\end{scope}

\node[star,star point ratio=2.25,minimum size=6pt,
          inner sep=0pt,draw=black,solid,fill=red] at (11.25, 0) {};
          
\node[star,star point ratio=2.25,minimum size=6pt,
          inner sep=0pt,draw=black,solid,fill=blue] at (10.25, 0.75) {};

\node[star,star point ratio=2.25,minimum size=6pt,
          inner sep=0pt,draw=black,solid,fill=green] at (11.5, 0.85) {};

\node at (11, -1.025) {\textbf{\begin{tabular}{c} \small{Classification}\\ \small{Interpretation} \end{tabular}}};

\draw[->] (8.75, 0.5) -- (9.5,0.5);

\node at (9,1.5)
{\includegraphics[width=0.5cm]{airplane2.png}};
\node at (9,1.15) {\textbf{Query}};
\draw [->] (9.25, 1.45) -- (9.45,1.45) -- (9.75,0.55);

\node[circle,minimum size=3pt,
          inner sep=0pt,draw=black,solid,fill=gray] at (9.75,0.55) {};

\draw [<->] (9.80,0.58) -- (10.16, 0.72);

\end{tikzpicture}
}
}
\caption{Experimental setup: top: standard DL model; middle: proposed framework with \textbf{no} finetuning; bottom: proposed framework \textbf{with} finetuning}
\label{fig:experiment_scheme}
\end{figure}

Throughout the experimental scenarios, we contrast three settings (see Figure \ref{fig:experiment_scheme}): 
\begin{itemize}
    \item A) Standard DL pipeline involving training on generic data sets as well as finetuning on target/"downstream" task/data - both with iterative error backpropagation
    \item B) IDEAL \textbf{without finetuning}: the proposed prototype-based IDEAL method involving clustering in the latent feature space with subsequent decision making process such as using winner-takes-all analysis or $k$ nearest neighbours as outlined in Algorithms \ref{alg:one} and \ref{alg:two}
    \item C) IDEAL \textbf{with finetuning}: Same as B) with the only difference that the clustering is performed in a latent feature space formed by finetuned on target data set (from the "downstream" task) using iterative error backpropagation. The main difference between the settings A) and C) is that setting C) does provide interpretable prototypes unlike setting A) 
\end{itemize}

In an extensive set of experiments, we demonstrate that with state-of-the-art models, such as ViT, the proposed IDEAL framework can provide interpretable results even \textbf{without finetuning}, which are competitive and extend to the lifelong learning, and mitigate confounding bias. For reproducibility, the full parameterisation is described in Section \ref{experimental_setup} of the Appendix.

The outline of the empirical questions is presented below. Questions 1 and 2 confirm that the method delivers competitive results \textbf{even without finetuning}; building upon this initial intuition we develop the key questions 3, 4 and 5, analysing the performance for lifelong learning scenarios and interpretations proposed by IDEAL respectively. 

\textbf{Question 1}. \textit{How does the performance of the IDEAL framework \textbf{without} finetuning compare with the well-known deep learning frameworks?}

Section \ref{classification_section} and Appendix \ref{Complete_experimental_results} show, with a concise summary in Figure \ref{fig:comparison_cifar10_feature_extractors} and Figure \ref{fig:comparison_vit}, that the gaps between finetuned and non-finetuned IDEAL framework are consistently much smaller (tens of percent vs a few percentage points) for vision transformer backbones comparing to ResNets and VGG. Furthermore, Figure \ref{fig:time_comparison_with_without_finetuning} shows that the training time expenditure is more than an order of magnitude smaller comparing to the original finetuning. 

\textbf{Question 2}. \textit{To what extent does finetuning of the feature space for the target problem lead to overfitting?}

In Section \ref{overfitting_demonstration}, figures \ref{tSNE_plots_ResNet101}, \ref{tSNE_plots_ViT}, \ref{IDEAL_vit_cifar100_cifar10_finetuning}, we demonstrate the issue of overfitting on the target spaces by finetuning on CIFAR-10 and testing on CIFAR-100 in both performance and through visualising the feature space. Interestingly, we also show in Table \ref{tab:cifar10_classification_fine-tuning} of the Appendix that, while the choice of prototypes greatly influences the performance of the IDEAL framework \textbf{without} finetuning of the backbone, it does not make any significant impact for the finetuned models (i.e., does not improve upon random selection).

\textbf{Question 3} \textit{How does the IDEAL framework \textbf{without} finetuning compare in the class-incremental learning setting?}

In Section \ref{continual_learning} we build upon questions 1 and 2 and demonstrate: the small gap between pretrained and finetuned ViT models ultimately enables us to solve class-incremental learning scenarios, improving upon well-known baseline methods. IDEAL framework \textbf{without} finetuning shows performance results on a number of class-incremental learning problems, comparable to task-level finetuning. Notably, in CIFAR-100 benchmark, the proposed method provides $83.2\%$ and $69.93\%$ on ViT-L and ResNet-101 respectively, while the state-of-the-art method from (\cite{wang2022online}) only reports $65.86\%$. 

\textbf{Question 4} \textit{How does the IDEAL framework provide insight and interpretation?}

In Section \ref{study_of_interpretability}, we present the analysis of interpretations provided by the method. In Figure \ref{fig:visual_interpretability_ideal} we demonstrate the qualitative experiments showing the human-readable interpretations provided by the model for both lifelong learning and offline scenarios. 
 
\textbf{Question 5}. \textit{Can models \textbf{without} finetuning bring advantage over the finetuned ones in terms of accuracy and help identify misclassifications due to confounding (spurious correlations in the input)?} 

While, admittedly, the model only approaches but does not reach the same level of accuracy for the same backbone without finetuning in the standard benchmarks such as CIFAR-10, it delivers better performance in cases with confounded data (with spurious correlations in the input). In Section \ref{confounding_interpretations}, Table \ref{tab:confounding_experiment} we demonstrate, building upon the intuition from Question 2, that finetuning leads to overfitting on confounded data, and leads to confounded predictions and interpretations. We also demonstrate that in this setting, IDEAL \textbf{without finetuning} improves, against the finetuned baseline, upon both F1 score, as well as providing the interpretations for wrong predictions due to the confounding. 

\subsection{Experimental setting}
We use the negative Euclidean distance between the feature vectors for our experiments:
\begin{equation}
    d (\mathbf{x}, \mathbf{p} | \theta_d) = -\ell^2 (\phi(\mathbf{x} | \mathbf{\theta}_d), \phi(\mathbf{p} | \mathbf{\theta}_d)),
\end{equation}
where $\phi$ is the normalised feature extractor output. The similarities bounded between $(0, 1]$ could be obtained by taking the exponential of the similarity function and normalising it.

Except from the experiment in Figure \ref{fig:k_nearest_neighbours}, the function $h$ is a winner-takes-all operator:
\begin{equation}
    h (\cdot)  = \mathrm{CLASS} (\arg \min_{p\in \mathbb{P}} d(\cdot, \mathbf{p} | \theta_d))
    \label{winner_takes_all}
\end{equation}

\paragraph {Datasets} CIFAR-10 and CIFAR-100 (\cite{krizhevskylearning}), STL-10 (\cite{coates2011analysis}), Oxford-IIIT Pet (\cite{parkhi12a}), EuroSAT (\cite{helber2018introducing,helber2019eurosat}), CalTech101 (\cite{li2006one}).
\paragraph {Feature extractors} We consider a number of feature extractor networks such as \textsc{VGG-16} (\cite{simonyan2014very}), \textsc{ResNet50} (\cite{he2016deep}), \textsc{ResNet101} (\cite{he2016deep}), \textsc{ViT-B/16} (\cite{dosovitskiy2020image}, referenced further as \textsc{ViT}), \textsc{ViT-L/16}  (\cite{dosovitskiy2020image}) (referenced further as ViT-L) with or without finetuning; the pre-trained latent spaces for ViT models were obtained using SWAG methodology  (\cite{singh2022revisiting}); the computations for feature extractors has been conducted using a single V100 GPU.
\paragraph {Clustering techniques} We include the results for such clustering techniques as $k$-means, $k$-means with a nearest data point (referred to as $k$-means (nearest)), and two online clustering methods: xDNN (\cite{angelov2020towards}) and ELM (\cite{baruah2012evolving}).
\paragraph {Baselines} We explore trade-offs between standard deep neural networks, different architectural choices  (averaged prototypes vs real-world examples), and, in Appendix \ref{Complete_experimental_results}, also compare the results with another prototype-based approach DNC (\cite{wang2023visual}).

\subsection{Offline classification}
\label{classification_section}
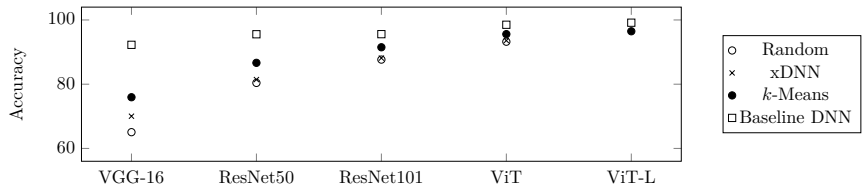
\begin{figure}
\centering
{
\resizebox{0.7\textwidth}{!}{
\begin{tikzpicture}
\begin{axis}[%
legend style={at={(1.3,0.5)},anchor=east},
scatter/classes={%
    random={mark=o,draw=black}, xDNN={mark=x,draw=black}, kMeans={mark=*,draw=black},baseline={mark=square,draw=black}},ymin=60,ymax=100,enlargelimits=true,ylabel=Accuracy,symbolic x coords={VGG-16, ResNet50,ResNet101,ViT,ViT-L},xtick={VGG-16, ResNet50,ResNet101,ViT,ViT-L},width=0.8\textwidth, 
        typeset ticklabels with strut,height=0.2\textheight]
\addplot[scatter,only marks,%
    scatter src=explicit symbolic]%
table[meta=label,row sep=crcr] {
x y label\\
VGG-16 65.06 random\\
VGG-16 70.03 xDNN\\
VGG-16 75.94 kMeans\\
VGG-16 92.26 baseline\\
ResNet50 80.40 random\\
ResNet50 81.44 xDNN\\
ResNet50 86.65 kMeans\\
ResNet50 95.55 baseline\\
ResNet101 87.66 random\\
ResNet101 88.13 xDNN\\
ResNet101 91.50 kMeans\\
ResNet101 95.58 baseline\\
ViT 93.23 random\\
ViT 93.59 xDNN\\
ViT 95.59 kMeans\\
ViT 98.51 baseline\\
ViT-L 96.48 kMeans\\
ViT-L 99.15 baseline\\
    };
\legend{Random,xDNN,$k$-Means, Baseline DNN}
\end{axis}
\end{tikzpicture}
}
}
\caption{Comparison of the proposed IDEAL framework (\textbf{without} finetuning) on the CIFAR-10 data set with different clustering methods 
(random, the clustering used in xDNN (\cite{soares2021explainable}) and $k$-means method) vs the baseline DNN}
\label{fig:comparison_cifar10_feature_extractors}
\end{figure}

We found that the gap between the models on a range of tasks decreases for the modern, high performance, architectures, such as ViT (\cite{dosovitskiy2020image}). For CIFAR-10, these findings are highlighted in Figure \ref{fig:comparison_cifar10_feature_extractors}: while finetuned VGG-16's accuracy is close to the one of ViT and other recent models, different prototype selection techniques (the one used in xDNN, $k$-means clustering, and random selection) all have accuracy between $60$ and $80\%$. The picture is totally different for ViT, where $k$-means prototype selection \textbf{without finetuning} provides accuracy of $95.59\%$ against finetuned ViT's own performance of $98.51\%$. 

\begin{figure}
\resizebox{0.45\textwidth}{!}{
\begin{subfigure}{0.5\textwidth}
\begin{tikzpicture}
\begin{axis}[legend pos=north east,log ticks with fixed point,enlargelimits=true,xlabel={Clustering methods},ylabel={Time},point meta=rawy, symbolic x coords={ViT,random,xDNN,$k$-means},nodes near coords, nodes near coords style={font=\tiny}, xtick=data,ybar=2*\pgflinewidth,width=\textwidth, height=0.15\textheight]
\addplot%
table[row sep=crcr] {
x y label\\
random 597 nofinetuning\\
xDNN 606 nofinetuning\\
$k$-means 925 nofinetuning\\
    };
\addplot%
table[row sep=crcr] {
x y label\\
random 15880 finetuning\\
xDNN 15889 finetuning\\
$k$-means 16220 finetuning\\
    };
\end{axis}
\end{tikzpicture}
\end{subfigure}
}
\resizebox{0.45\textwidth}{!}{
\begin{subfigure}{0.5\textwidth}
\begin{tikzpicture}
\begin{axis}[legend style={at={(1.3,0.5)},anchor=north},log ticks with fixed point,enlargelimits=true,xlabel={Clustering methods},point meta=rawy, symbolic x coords={ViT,random,xDNN,$k$-means},nodes near coords, nodes near coords style={font=\tiny}, xtick=data,ybar=2*\pgflinewidth,width=\textwidth, height=0.15\textheight]
\addplot%
table[row sep=crcr] {
x y label\\
random 604 nofinetuning\\
xDNN 613 nofinetuning\\
$k$-means 673 nofinetuning\\
    };
\addlegendentry{No finetuning};
\addplot%
table[row sep=crcr] {
x y label\\
random 16157 finetuning\\
xDNN 16166 finetuning\\
$k$-means 16231 finetuning\\
    };
\addlegendentry{finetuning};
\end{axis}
\end{tikzpicture}
\end{subfigure}
}

\caption{Comparison of training time expenditure on CIFAR-10 (left) and CIFAR-100 (right) with and without funetuning (ViT)}
\label{fig:time_comparison_with_without_finetuning}
\end{figure}
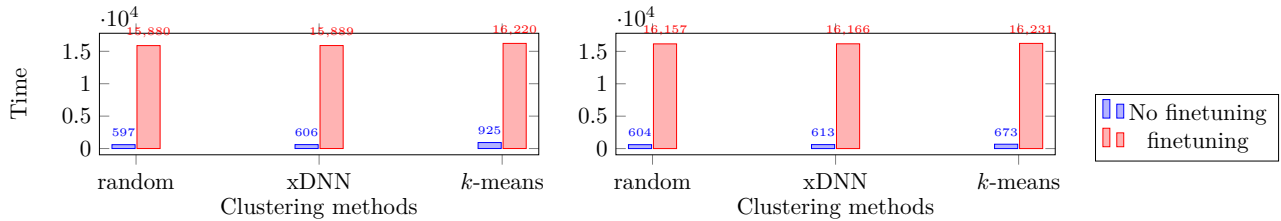

While the results above report on performance of the $k$-means clustering used as a prototype selection technique, the experimental results in Figure \ref{fig:k_nearest_neighbours} explore choosing the nearest prototype to $k$-means cluster centroid for interpretability reasons. While it is clear (with further evidence presented in Appendix \ref{Complete_experimental_results}) that the performance when selecting the nearest to the $k$-means centroids prototypes is lagging slightly behind the direct use of the centorids (denoted simply as $k$-means), it is possible to bring this performance closer by replacing the winner-takes-all decision making approach (Equation (\ref{winner_takes_all})) with the $k$ nearest neighbours method. For this purpose, we utilise the sklearn's \texttt{KNeighborsClassifier} function. 

The abridged results for classification \textbf{without} finetuning for different tasks are presented in Figure \ref{fig:comparison_vit} (one can find a full version for different methods in Section \ref{Complete_experimental_results}). 

\begin{figure}
    \centering
\resizebox{0.49\textwidth}{!}{
\begin{subfigure}{0.4\textwidth}
\begin{tikzpicture}[domain=0:5]
\begin{axis}[legend style={at={(0.5,1.8)},anchor=north},log ticks with fixed point,ymin=90,ymax=100,enlargelimits=true,xlabel={Methods (CIFAR-10)},point meta=rawy, symbolic x coords={ViT,random,xDNN,$k$-means},nodes near coords, nodes near coords style={font=\tiny}, xtick = data,
    ylabel={Accuracy},error bars/y dir=both, error bars/y explicit,ybar=2*\pgflinewidth,width=\textwidth, height=0.15\textheight]
  \addplot coordinates {
                (ViT,0) 
                (random,93.23) += (0,0.11) -= (0,0.11)
                (xDNN,93.59) += (0,0.12) -= (0,0.12)
                ($k$-means,95.59) += (0,0.08) -= (0,0.08)
            };         
 \addlegendentry{No finetuning};
  \addplot coordinates {
                (ViT,98.51)
                (random,98.56) += (0,0.02) -= (0,0.02)
                (xDNN,98.00)  += (0,0.14) -= (0,0.14)
                ($k$-means,98.53) += (0,0.04) -= (0,0.04)
            }; 
 \addlegendentry{finetuned};
\end{axis}
\end{tikzpicture}
\end{subfigure}
}
\resizebox{0.49\textwidth}{!}{
\begin{subfigure}{0.4\textwidth}
\begin{tikzpicture}[domain=0:5]
\begin{axis}[log ticks with fixed point,ymin=70,ymax=100,enlargelimits=true,xlabel={Methods (CIFAR-100)},point meta=rawy, symbolic x coords={ViT, random, xDNN, $k$-means},error bars/y dir=both, error bars/y explicit,nodes near coords, nodes near coords style={font=\tiny}, xtick = data,ybar=2*\pgflinewidth,width=\textwidth, height=0.15\textheight]
  \addplot coordinates {
                (ViT,0)
                (random,72.39) += (0,0.21) -= (0,0.21)
                (xDNN,76.24) += (0,0.24) -= (0,0.24)
                ($k$-means,79.12) += (0,0.28) -= (0,0.28)
            };
   \addplot coordinates {
                (ViT,90.29)
                (random,89.90) += (0,0.10) -= (0,0.10)
                (xDNN,89.17) += (0,0.18) -= (0,0.18)
                ($k$-means,90.48) += (0,0.05) -= (0,0.05)
            };
\end{axis}
\end{tikzpicture}
\end{subfigure}
}
\resizebox{0.49\textwidth}{!}{
\begin{subfigure}{0.4\textwidth}
\begin{tikzpicture}[domain=0:15]
\begin{axis}[legend pos=north east,log ticks with fixed point,ymin=98,ymax=100,enlargelimits=true,xlabel={Methods (STL-10)},
    ylabel={Accuracy},point meta=rawy, symbolic x coords={ViT,random, xDNN,$k$-means},error bars/y dir=both, error bars/y explicit,nodes near coords, nodes near coords style={font=\tiny},xtick=data,ybar=2*\pgflinewidth,width=\textwidth, height=0.15\textheight]
  \addplot coordinates {
                (ViT,0)
                (random,98.55) += (0,0.09) -= (0,0.09)
                (xDNN,98.63) += (0,0.12) -= (0,0.12)
                ($k$-means,99.32) += (0,0.03) -= (0,0.03)
            };
  \addplot coordinates {
                (ViT,98.97)
            };
\end{axis}
\end{tikzpicture}
\end{subfigure}
}
\resizebox{0.49\textwidth}{!}{
\begin{subfigure}{0.4\columnwidth}
\begin{tikzpicture}[domain=0:15]
\begin{axis}[legend pos=north east,log ticks with fixed point,ymin=90,ymax=100,enlargelimits=true,xlabel={Methods (Oxford-IIIT Pet)},point meta=rawy, symbolic x coords={ViT,Random,xDNN,$k$-means},error bars/y dir=both, error bars/y explicit,nodes near coords, nodes near coords style={font=\tiny}, xtick=data, ybar=2*\pgflinewidth, width=\textwidth, height=0.15\textheight]
  \addplot coordinates {
                (ViT,0)
                (Random,90.83) += (0,0.51) -= (0,0.51)
                (xDNN,94.30) += (0,0.23) -= (0,0.23)
                ($k$-means,94.07) += (0,0.20) -= (0,0.20)
            };  
   \addplot coordinates {
                (ViT,94.41)
            };
\end{axis}
\end{tikzpicture}
\end{subfigure}
}
\resizebox{0.49\textwidth}{!}{
\begin{subfigure}{0.4\columnwidth}
\begin{tikzpicture}[domain=0:15]
\begin{axis}[legend pos=north east,log ticks with fixed point,ymin=85,ymax=100,enlargelimits=true,xlabel={Methods (Caltech-101)},ylabel={Accuracy},point meta=rawy, symbolic x coords={ViT,random,xDNN,$k$-means},error bars/y dir=both, error bars/y explicit,nodes near coords, nodes near coords style={font=\tiny}, xtick=data,ybar=2*\pgflinewidth,width=\textwidth, height=0.15\textheight]
  \addplot coordinates {
                (ViT,0)
                (random,89.42) += (0,0.32) -= (0,0.32)
                (xDNN,94.61) += (0,0.94) -= (0,0.94)
                ($k$-means,94.46) += (0,0.44) -= (0,0.44)
            };  
   \addplot coordinates {
                (ViT,96.26)
            };
\end{axis}
\end{tikzpicture}
\end{subfigure}
}
\resizebox{0.49\textwidth}{!}{
\begin{subfigure}{0.4\columnwidth}
\begin{tikzpicture}[domain=0:15]
\begin{axis}[legend pos=north east,log ticks with fixed point,ymin=80,ymax=100,enlargelimits=true,xlabel={Methods (EuroSAT)},point meta=rawy, symbolic x coords={ViT,random,xDNN,$k$-means},error bars/y dir=both, error bars/y explicit,nodes near coords, nodes near coords style={font=\tiny}, xtick=data,ybar=2*\pgflinewidth,width=\textwidth, height=0.15\textheight]
  \addplot coordinates {
                (ViT,0)
                (random,82.76) += (0,0.54) -= (0,0.54)
                (xDNN,85.24) += (0,1.05) -= (0,1.05)
                ($k$-means,91.43) += (0,0.16) -= (0,0.16)
            };  
   \addplot coordinates {
                (ViT,95.17)
            };
\end{axis}
\end{tikzpicture}
\end{subfigure}
}
    \caption{Comparison of results with 
    ViT (\cite{dosovitskiy2020image}) as a feature extractor; \{Random,xDNN, $k$-means\}=Proposed (\{Random, xDNN, $k$-means\} prototype selection) }
    \label{fig:no_finetuning_vs_fine-tuned_results}
\end{figure}
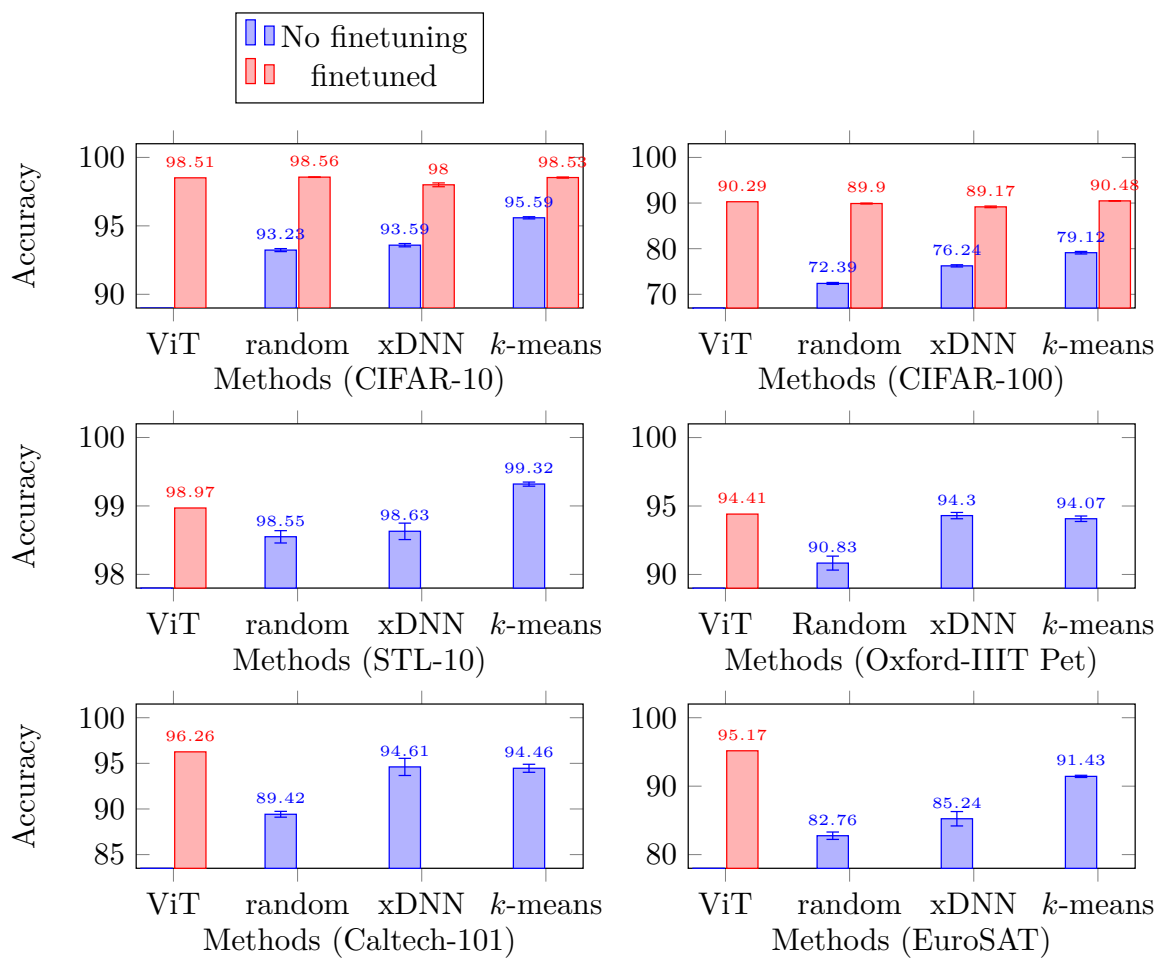

\begin{figure}
  \centering{
\includegraphics[width=0.39\textwidth]{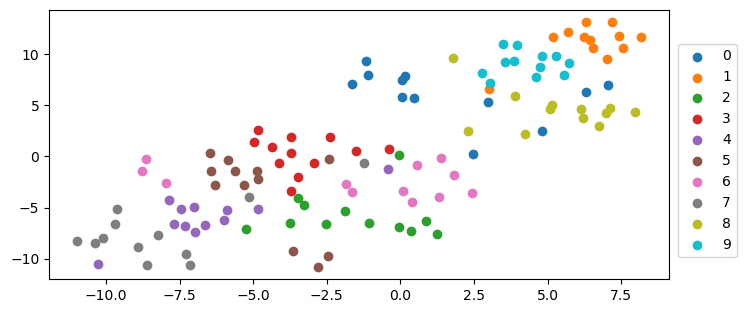}
\includegraphics[width=0.39\textwidth]{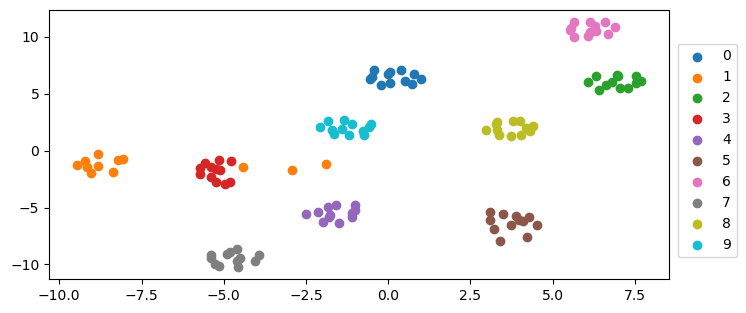}\\
\includegraphics[width=0.39\textwidth]{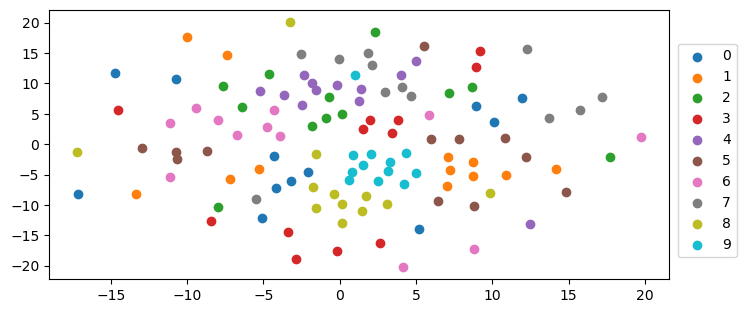}
\includegraphics[width=0.39\textwidth]{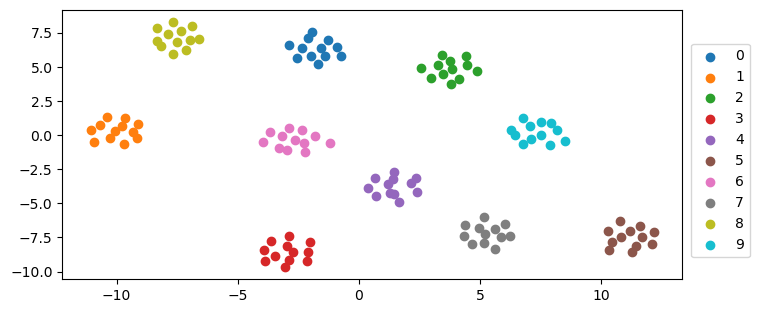}
}
\caption{tSNE plots for original (top-left) vs finetuned (top-right) features of ResNet101, k-means prototypes;  original (bottom left) vs finetuned (bottom right), ResNet101, random prototype selection, CIFAR-10}
\label{tSNE_plots_ResNet101}
\end{figure}

 \begin{figure}
  \centering{
\includegraphics[width=0.39\textwidth]{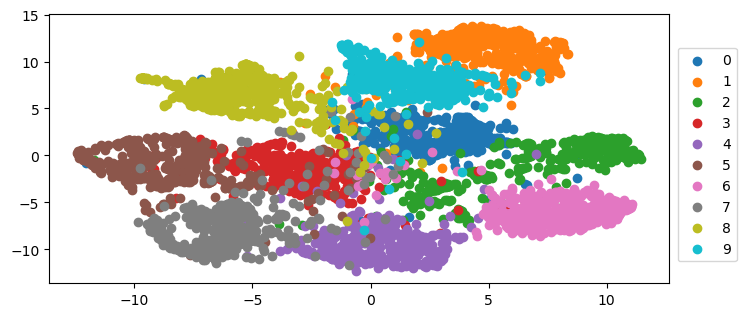}
\includegraphics[width=0.39\textwidth]{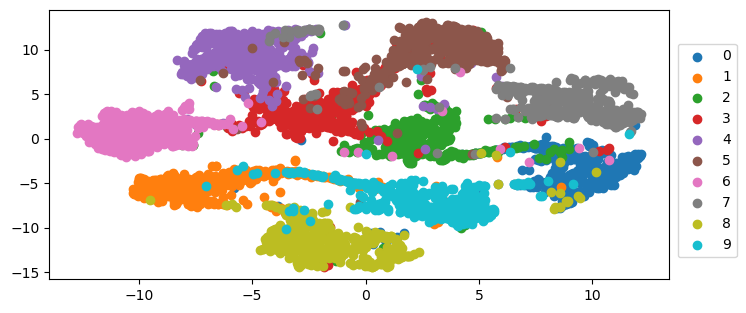}\\
\includegraphics[width=0.39\textwidth]{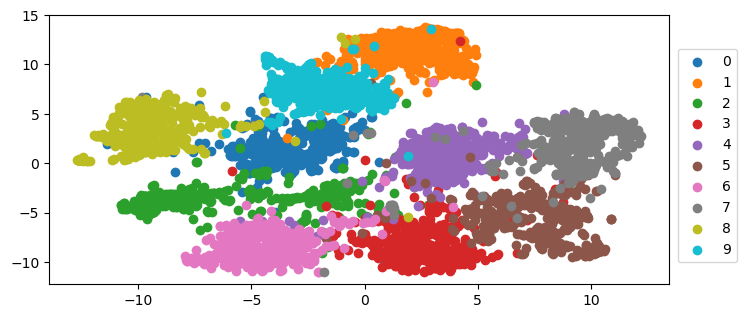}
\includegraphics[width=0.39\textwidth]{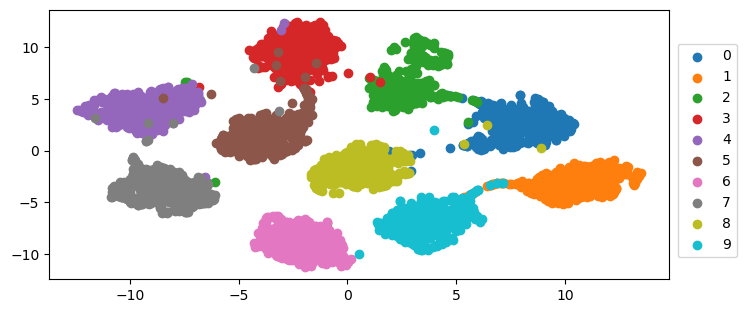}
}
\caption{tSNE plots for original (top-left) vs finetuned (top-right) features of ViT, $k$-means prototypes;  original (bottom left) vs finetuned (bottom right), ViT, random prototype selection, CIFAR-10}
\label{tSNE_plots_ViT}
\end{figure}

\begin{figure}
\centering
{
\resizebox{0.9\textwidth}{!}{
\begin{subfigure}{0.8\columnwidth}
\begin{tikzpicture}[domain=0:15]
\begin{axis}[legend style={at={(0.5,1.8)},anchor=north},log ticks with fixed point,ymin=50,ymax=100,enlargelimits=true,xlabel={Methods (CIFAR-100)},point meta=rawy, symbolic x coords={Random,xDNN,$k$-means},nodes near coords, nodes near coords style={font=\tiny}, xtick = data,
    ylabel={Accuracy},ybar=2*\pgflinewidth,width=\textwidth, height=0.15\textheight]
  \addplot coordinates {
                 (Random,72.39) (xDNN,76.24)  ($k$-means,79.12)
            };         
 \addlegendentry{Proposed (ViT backbone \textbf{without} finetuning)};
  \addplot coordinates {
                 (Random,63.86) (xDNN,64.55)  ($k$-means,73.49)
            }; 
 \addlegendentry{Proposed (ViT backbone with finetuning on CIFAR-10)};
\end{axis}
\end{tikzpicture}
\end{subfigure}
}
}
\caption{Comparison between the model performance on CIFAR-100 \textbf{without} finetuning and finetuning on CIFAR-10}
\label{IDEAL_vit_cifar100_cifar10_finetuning}
\end{figure}
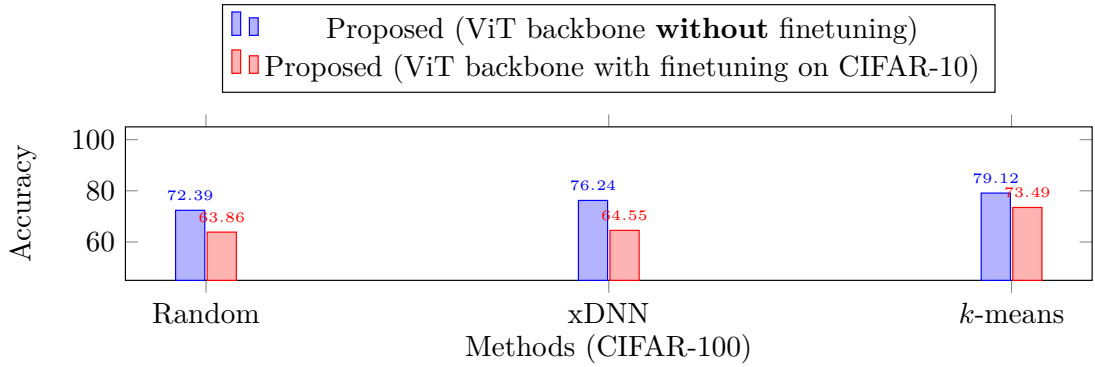

\begin{figure}
\centering
{
\resizebox{0.8\textwidth}{!}{
\begin{tikzpicture}[domain=0:15]
\begin{axis}[legend pos=north east,ymin=93,ymax=100,enlargelimits=true,xlabel={$k$ in $k$-NN},point meta=rawy, symbolic x coords={k-means,1,3,5,7,9,11,13,15},nodes near coords, nodes near coords style={font=\tiny}, xtick = data,
    ylabel={Accuracy, \%},error bars/y dir=both, error bars/y explicit,ybar=2*\pgflinewidth,width=\textwidth, height=0.2\textheight]
  \addplot coordinates {
                (k-means,95.59) += (0,0.08) -= (0,0.08)
                (1,94.04) += (0,0.10) -= (0,0.10)
                (3,94.91) += (0,0.10) -= (0,0.10)
                (5,95.22) += (0,0.09) -= (0,0.09)
                (7,95.32) += (0,0.12) -= (0,0.12)
                (9,95.34) += (0,0.08) -= (0,0.08)
                (11,95.32)  += (0,0.06) -= (0,0.06)
                (13,95.32)   += (0,0.06) -= (0,0.06)
                (15,95.30)  += (0,0.06) -= (0,0.06)
            };
 \addlegendentry{Proposed ($k$-means)};
 \addlegendentry{ResNet101};
\end{axis}
\end{tikzpicture}
}
}
\caption{Comparison of results on CIFAR-10 ($k$ nearest neighbours)}
\label{fig:k_nearest_neighbours}
\end{figure}
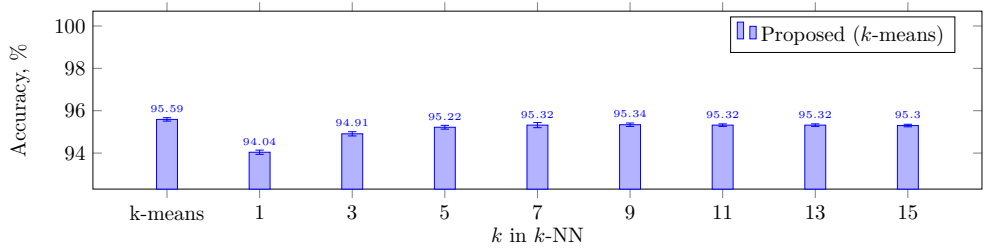

Below, we analyse closer just the results with using ViT as a feature extractor forming the latent data space. One can see in Figure  \ref{fig:no_finetuning_vs_fine-tuned_results} that: 
(1) \textbf{without finetuning}, on a number of tasks the model shows competitive performance, and 
(2) with finetuning of the backbone, the difference between the standard backbone and the proposed model is insignificant within the confidence interval.
In Figure \ref{fig:time_comparison_with_without_finetuning}, one can see the comparison of the time expenditure between the finetuned and \textbf{non-finetuned} model.

We also conducted (see Appendix \ref{sensitivity_number_of_prototypes}) an experiment to vary the selected number of prototypes for CIFAR-10 on ResNet101 backbone and the value $k$ for the $k$-means method. It is a well-known specific of the $k$-means approach that it does require the number of clusters, $k$ to be pre-defined. The online clustering method ELM, for example, does not require the number of clusters to be pre-defined, though it requires a single meta-parameter, called radius of the cluster to be pre-defined which can be related to the expected granulation level considering all data being normalized (\cite{baruah2012evolving}). Therefore, in Appendix \ref{Complete_experimental_results}, we include results for ELM.

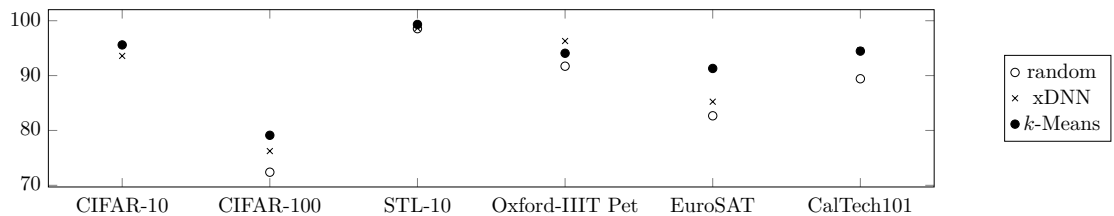
\begin{figure}
\centering
{
\resizebox{0.8\textwidth}{!}{
\begin{subfigure}{\columnwidth}
\begin{tikzpicture}
\begin{axis}[%
scatter/classes={%
    random={mark=o,draw=black}, xDNN={mark=x,draw=black}, kMeans={mark=*,draw=black}},symbolic x coords={CIFAR-10,CIFAR-100,STL-10,Oxford-IIIT\ Pet,EuroSAT,CalTech101},xtick={CIFAR-10,CIFAR-100,STL-10,Oxford-IIIT\ Pet,EuroSAT,CalTech101},width=\textwidth, 
        typeset ticklabels with strut,height=0.2\textheight,legend style={at={(1.2,0.5)},anchor=east}]
\addplot[scatter,only marks,%
    scatter src=explicit symbolic]%
table[meta=label,row sep=crcr] {
x y label
CIFAR-10 93.23 random\\
CIFAR-10 93.59 xDNN\\
CIFAR-10 95.59 kMeans\\
CIFAR-100 72.39 random\\
CIFAR-100 76.24 xDNN\\
CIFAR-100 79.12 kMeans\\
STL-10 98.55 random\\
STL-10 98.63 xDNN\\
STL-10 99.32 kMeans\\
Oxford-IIIT\ Pet 91.71 random\\
Oxford-IIIT\ Pet 96.30 xDNN\\
Oxford-IIIT\ Pet 94.07 kMeans\\
EuroSAT 82.67 random\\
EuroSAT 85.24 xDNN\\
EuroSAT 91.30 kMeans\\
CalTech101 89.42 random\\
CalTech101 94.61 xDNN\\
CalTech101 94.46 kMeans\\
    };
\legend{random,xDNN,$k$-Means}
\end{axis}
\end{tikzpicture}
\end{subfigure}
}
}
\caption{Results \textbf{without} finetuning for various problems (ViT)}
\label{fig:comparison_vit}
\end{figure}

\subsection{Demonstration of overfitting in the feature spaces}
\label{overfitting_demonstration}
 One clear advantage of transfer learning without finetuning is the dramatically lower computational costs reflected in the time expenditure. However, there is also another advantage: the evidence shows that the finetuned feature space shows less generalisation. In Figures \ref{tSNE_plots_ResNet101} and \ref{tSNE_plots_ViT}, one can see the comparison of the tSNE plots between the finetuned and \textbf{non-finetuned} version of the method. While the finetuned method achieves clear separation on this task, using the same features to transfer to another task (from CIFAR-10 to CIFAR-100) leads to sharp decrease in performance (see Figure \ref{IDEAL_vit_cifar100_cifar10_finetuning}). Despite the time consumption and limited generalisation, the finetuned version of the proposed framework, see setting C), section 4 and also Tables 3 and 5. has one advantage: it demonstrates that with a small computational cost additional to finetuning, a standard DL classifier can be transformed into interpretable through prototypes ones with difference in performance within the confidence interval. While for the finetuned backbone, predictably, the results are not far off the standard DL models, they also show no significant difference between different types of prototype selection, including random (see Figure \ref{fig:no_finetuning_vs_fine-tuned_results}); however, for the non-finetuned results, the difference in top-1 accuracy between random and non-random prototype selection is drastic, reaching around 24\% for VGG16. 

The choice of prototypes greatly influences performance of a model when it is not finetuned as witnessed in a number of tasks for a number of backbone models. In Figure \ref{fig:comparison_cifar10_feature_extractors} and Appendix \ref{Complete_experimental_results}, one can see that simple $k$-means prototype selection in the latent space can improve the performance by tens of percentage points; with the increase of the number of prototypes this difference decreases, but is still present. Furthermore, one can see that the proposed framework without finetuning and online prototype selection algorithm can be competitive with the state-of-the-art, especially when working in latent feature space defined by powerful DNNs such as ViT on large data sets. When using finetuning, it is seen that the choice of prototypes, including random, does not make significant difference. This can be explained by the previous discussion of Figures \ref{tSNE_plots_ResNet101} and \ref{tSNE_plots_ViT}: finetuning gives clear separation of features, so the features of the same class stay close; that makes the prototype choice practically unimportant for decision making.

\subsection{Continual learning}
\label{continual_learning}
\begin{figure}
\begin{subfigure}{\textwidth}
\centering{
\resizebox{\textwidth}{!}{
\begin{tikzpicture}[domain=0:15]
\begin{axis}[legend pos=north east, legend columns=-1, legend style={draw=none, column sep=1ex},ymin=60,ymax=100,enlargelimits=true,xlabel={Number of training classes},point meta=rawy,,nodes near coords=\rotatebox{90}{\pgfmathprintnumber\pgfplotspointmeta}, nodes near coords style={font=\fontsize{5}{8}}, xtick = data,
    ylabel={Accuracy, \%},error bars/y dir=both, error bars/y explicit,ybar=\pgflinewidth,width=\textwidth, height=0.28\textheight]
  \addplot coordinates {
                (10,97.34) += (0,0.20) -= (0,0.20)
                (20,93.73) += (0,0.21) -= (0,0.21)
                (30,91.63) += (0,0.21) -= (0,0.21)
                (40,89.58) += (0,0.19) -= (0,0.19)
                (50,88.00) += (0,0.19) -= (0,0.19)
                (60,86.11)  += (0,0.26) -= (0,0.26)
                (70,84.11)   += (0,0.25) -= (0,0.25)
                (80,83.66)  += (0,0.21) -= (0,0.21)
                (90,83.59)  += (0,0.20) -= (0,0.20)
                (100,82.20)  += (0,0.23) -= (0,0.23)
            };
  \addlegendentry{ViT-L};
  \addplot coordinates {
                     (10,94.29) += (0,0.31) -= (0,0.31)
                     (20,88.27) += (0,0.26) -= (0,0.26)
                     (30,84.06) += (0,0.41) -= (0,0.41)
                     (40,81.06) += (0,0.41) -= (0,0.41)
                     (50,78.67) += (0,0.22) -= (0,0.22)
                     (60,76.06) += (0,0.27) -= (0,0.27)
                     (70,74.57) += (0,0.24) -= (0,0.24)
                     (80,72.41) += (0,0.25) -= (0,0.25)
                     (90,71.91) += (0,0.26) -= (0,0.26)
                     (100,69.93) += (0,0.23) -= (0,0.23)
                     };
  \addlegendentry{ResNet-101};
  \addplot coordinates {
                     (10,92.20) += (0,0.62) -= (0,0.62)
                     (20,83.66) += (0,0.48) -= (0,0.48)
                     (30,79.42) += (0,0.38) -= (0,0.38)
                     (40,76.34) += (0,0.25) -= (0,0.25)
                     (50,73.49) += (0,0.32) -= (0,0.32)
                     (60,70.84) += (0,0.32) -= (0,0.32)
                     (70,69.24) += (0,0.30) -= (0,0.30)
                     (80,66.94) += (0,0.24) -= (0,0.24)
                     (90,66.14) += (0,0.19) -= (0,0.19)
                     (100,64.18) += (0,0.16) -= (0,0.16)
                     };
  \addlegendentry{ResNet-50};
\end{axis}
\end{tikzpicture}
}
}
\caption{iCIFAR-100}
\label{fig:incremental_learning_cifar100}
\end{subfigure}

\begin{subfigure}{\textwidth}
\centering{
\resizebox{\textwidth}{!}{
\begin{tikzpicture}[domain=0:15]
\begin{axis}[legend pos=north east, legend columns=-1, legend style={draw=none, column sep=1ex},ymin=85,ymax=100,enlargelimits=true,xlabel={Number of training classes},point meta=rawy,nodes near coords=\rotatebox{90}{\pgfmathprintnumber\pgfplotspointmeta}, nodes near coords style={font=\fontsize{5}{8}}, xtick = data,
    ylabel={Accuracy, \%},error bars/y dir=both, error bars/y explicit,ybar=\pgflinewidth,width=\textwidth, height=0.25\textheight]
  \addplot coordinates {
                (10,94.53) += (0,0.61) -= (0,0.61)
                (20,95.35) += (0,0.49) -= (0,0.49)
                (30,94.91) += (0,0.43) -= (0,0.43)
                (40,95.31) += (0,0.40) -= (0,0.40)
                (50,95.67) += (0,0.37) -= (0,0.37)
                (60,95.84) += (0,0.33) -= (0,0.33)
                (70,96.00) += (0,0.30) -= (0,0.30)
                (80,95.90)  += (0,0.26) -= (0,0.26)
                (90,96.08)  += (0,0.24) -= (0,0.24)
                (101,96.05)  += (0,0.24) -= (0,0.24)
            };
  \addlegendentry{ViT-L};
  
  \addplot coordinates {
                     (10,90.77) += (0,0.45) -= (0,0.45)
                     (20,91.40) += (0,0.41) -= (0,0.41)
                     (30,90.70) += (0,0.49) -= (0,0.49)
                     (40,91.21) += (0,0.49) -= (0,0.49)
                     (50,91.07) += (0,0.48) -= (0,0.48)
                     (60,91.01) += (0,0.48) -= (0,0.48)
                     (70,90.81) += (0,0.43) -= (0,0.43)
                     (80,90.51) += (0,0.42) -= (0,0.42)
                     (90,90.57) += (0,0.41) -= (0,0.41)
                     (101,90.47) += (0,0.41) -= (0,0.41)
                     };
  \addlegendentry{ResNet-101};

  \addplot coordinates {
                     (10,91.62) += (0,0.66) -= (0,0.66)
                     (20,92.26) += (0,0.57) -= (0,0.57)
                     (30,91.53) += (0,0.52) -= (0,0.52)
                     (40,91.91) += (0,0.46) -= (0,0.46)
                     (50,92.22) += (0,0.39) -= (0,0.39)
                     (60,91.94) += (0,0.28) -= (0,0.28)
                     (70,91.86) += (0,0.31) -= (0,0.31)
                     (80,91.44) += (0,0.26) -= (0,0.26)
                     (90,91.54) += (0,0.29) -= (0,0.29)
                     (101,91.21) += (0,0.38) -= (0,0.38)
                     };
  \addlegendentry{ResNet-50};
\end{axis}
\end{tikzpicture}
}
}
\caption{iCaltech101}
\label{fig:incremental_learning_caltech101}
\end{subfigure}

\begin{subfigure}{\textwidth}
\centering{
\resizebox{\textwidth}{!}{
\begin{tikzpicture}[domain=0:15]
\begin{axis}[legend pos=north east, legend columns=-1, legend style={draw=none, column sep=1ex},ymin=94,ymax=100,enlargelimits=true,xlabel={Number of training classes},point meta=rawy,nodes near coords, nodes near coords style={font=\fontsize{7}{8}}, xtick = data,
    ylabel={Accuracy, \%},error bars/y dir=both, error bars/y explicit,ybar=\pgflinewidth,width=\textwidth, height=0.17\textheight]
  \addplot coordinates {
                (2,99.65) += (0,0.09) -= (0,0.09)
                (4,98.14) += (0,0.08) -= (0,0.08)
                (6,96.06) += (0,0.10) -= (0,0.10)
                (8,95.54) += (0,0.10) -= (0,0.10)
                (10,95.01) += (0,0.08) -= (0,0.08)
            };
  \addlegendentry{ViT-L};

\end{axis}
\end{tikzpicture}
}
}
\caption{iCIFAR10}
\label{fig:incremental_learning_cifar10}
\end{subfigure}
\caption{Accuracy of IDEAL in class-incremental learning experiments for different backbones (ViT-L, ResNet-101 and 50). }
\label{fig:incremental_learning}
\end{figure}
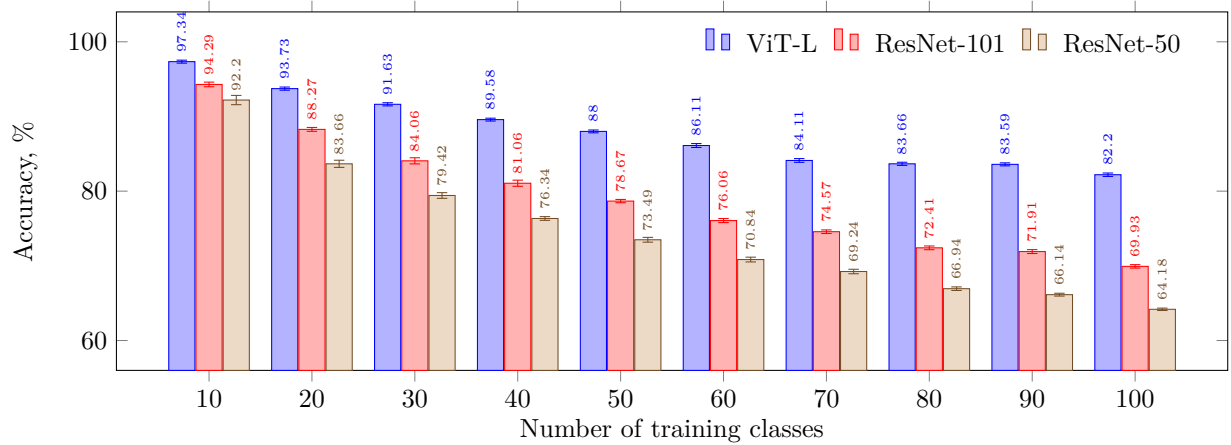
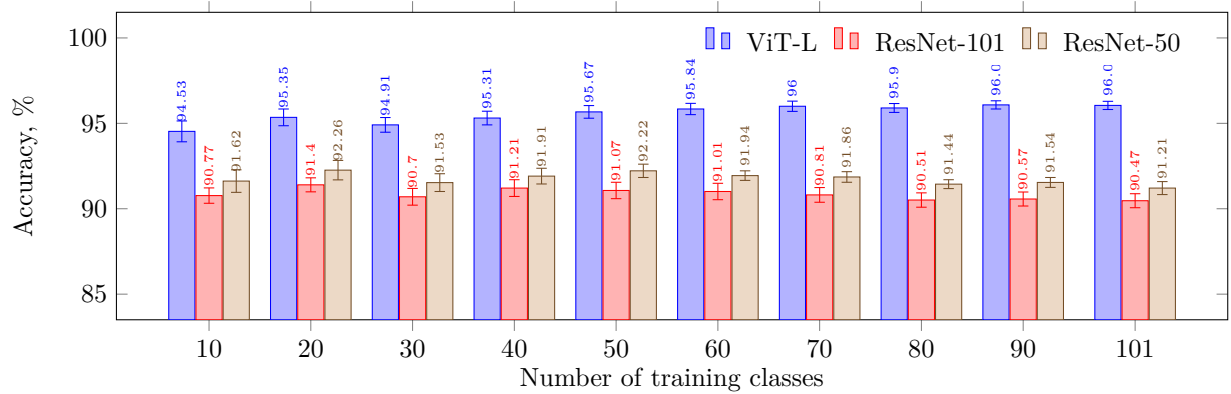
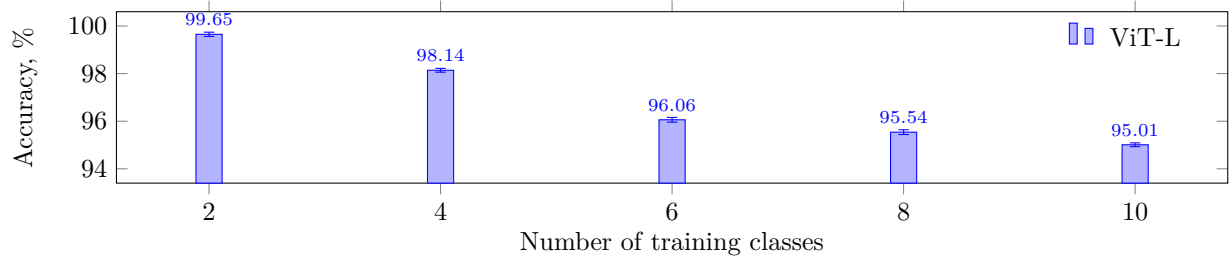

The evidence from the previous sections motivates us to extend the analysis to continual learning problems: given a much smaller gap between the finetuned and non-finetuned models, can the IDEAL framework \textbf{without} finetuning compete with the state-of-the-art class-incremental learning baselines? It turns out the answer is affirmative. We repeat the setting from \cite{rebuffi2017icarl} (Section 4, iCIFAR-100 benchmark) using IDEAL without finetuning the latent space of the ViT-L model. This benchmark gradually adds the new classes with a class increment of 10, until it reaches 100 classes. The results, shown in Figure \ref{fig:incremental_learning_cifar100}, highlight excellent performance of the proposed method (the number of prototypes is $10000$ or $100$ per class on average, however, as one can see in Appendix \ref{sensitivity_number_of_prototypes}, much lower number of prototypes, below $1000$ or just $10$ per class on average can still lead to competitive results). While we report $64.18 \pm 0,0.16$, $69.93 \pm 0.23\%$, $82.20\pm 0.23$ for ResNet-50, ResNet-101, and ViT-L respectively, \cite{wang2022online} reports in its Table 1 for the best performing method for class-incremental learning, based on ViT architecture and contrastive learning, accuracy of just $65.86 \pm 1.24\%$ (with the size of the buffer - $1000$), while the original proposed benchmark \cite{rebuffi2017icarl} (iCarl) reports,  according to \cite{wang2022online}, only $50.49 \pm 0.18\%$. 

To demonstrate the consistent performance, we expanded iCIFAR-100 protocol to other datasets, referred to as iCaltech101 and iCIFAR-10. Figure \ref{fig:incremental_learning} shows robust performance on iCaltech101 and iCIFAR-10. We use the class increment value of ten (eleven for the last step) and two for iCaltech101 and iCIFAR-10, respectively.  The hyperparameters of the proposed methods are given in Appendix \ref{experimental_setup}. We see that for iCaltech101, the model performance changes insignificantly with adding the training classes, and all three datasets demonstrate performance similar to the offline classification performance (see Section \ref{classification_section}).

\subsection{Study of Interpretability}
\label{study_of_interpretability}

\begin{figure}
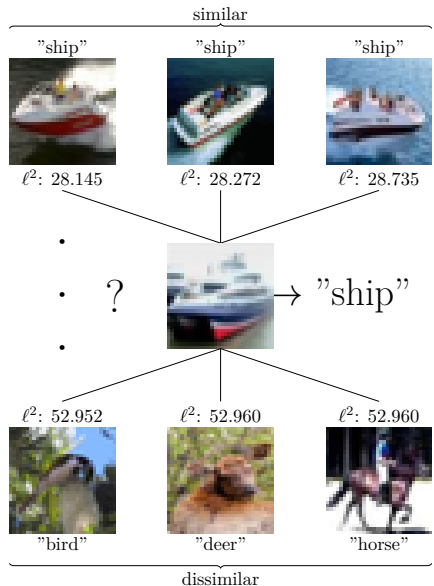

\centering{
\resizebox{0.36\textwidth}{!}{
\begin{tikzpicture}[domain=0:15]
\draw[decoration={brace,raise=2pt},decorate] (-4,5) -- node[above=3pt] {similar} (4,5); 
\node at (-3,4.7) {''ship''};
\node at (-3,3.5) {\includegraphics[width=2cm]{interpretation_example/39278.png}};
\node at (-3,2.25) {$\ell^2$: $28.145$};
\draw (-3,2) -- (0,1);

\node at (0,4.7) {''ship''};
\node at (0,3.5) {\includegraphics[width=2cm]{interpretation_example/15416.png}};
\node at (0,2.25) {$\ell^2$: $28.272$};
\draw (0,2) -- (0,1);

\node at (3,4.7) {''ship''};
\node at (3,3.5) {\includegraphics[width=2cm]{interpretation_example/19357.png}};
\node at (3,2.25) {$\ell^2$: $28.735$};
\draw (3,2) -- (0,1);

\node at (-2,0) {\Huge{?}};
\node at (-3,1) {\Huge{$\mathcal{\cdot}$}};
\node at (-3,0) {\Huge{$\mathcal{\cdot}$}};
\node at (-3,-1) {\Huge{$\mathcal{\cdot}$}};
\node at (2,0) {\huge{$\longrightarrow$ ''ship''}};

\node at (0,0) {\includegraphics[width=2cm]{interpretation_example/1.png}};

\node at (-3,-3.5) {\includegraphics[width=2cm]{interpretation_example/19287.png}};
\draw (3,-2) -- (0,-1);
\node at (-3,-2.25) {$\ell^2$: $52.952$};
\node at (-3,-4.7) {''bird''};

\node at (0,-3.5) {\includegraphics[width=2cm]{interpretation_example/28509.png}};
\draw (0,-2) -- (0,-1);
\node at (0,-2.25) {$\ell^2$: $52.960$};
\node at (0,-4.7) {''deer''};

\node at (3,-3.5) {\includegraphics[width=2cm]{interpretation_example/14561.png}};
\draw (-3,-2) -- (0,-1);
\node at (3,-2.25) {$\ell^2$: $52.960$};
\node at (3,-4.7) {''horse''};

\draw[decoration={brace,raise=2pt,mirror},decorate] (-4,-5) -- node[below=3pt]{dissimilar} (4,-5);
\end{tikzpicture}
}
}
\caption{Interpreting the predictions of the proposed model ($k$-means (nearest), CIFAR-10, ViT)}
\label{visual_interpretability_ideal}
\end{figure}

In Figure \ref{visual_interpretability_ideal_3_offline}, we demonstrate the visual interpretability of the proposed model, through both most similar and most dissimilar prototypes. In addition, the results could be interpreted linguistically (see Appendix \ref{linguistic_interpretability_of_IDEAL_outputs}). Figure \ref{visual_interpretability_ideal_3_offline} shows a number of quantitative examples for a number of datasets: Caltech101, STL-10, Oxford-IIIT Pets, all corresponding to the non-finetuned feature space scenario according to the experimental setup from Appendix \ref{experimental_setup}. We see that on a range of datasets, without any finetuning, the proposed IDEAL approach provides semantically meaningful interpretations. Furthermore, as there has been no finetuning, the $\ell^2$ distances are defined in exact the same feature space and, hence, can be compared like-for-like between datasets (see subfigures \ref{oxford_iiit_pets_correct_interpretation}-\ref{caltech101_incorrect_interpretation}). This strengthens the evidence of the benefits of our approach \textbf{without finetuning}. This experiment demonstrates that the incorrectly classified data tend to have larger distance to the closest prototypes than the correctly classified ones. Finally, Figure \ref{continual_learning_ranking_evolution} outlines the evolution of predictions for the online scenario. For the sake of demonstration, we used the same setting as the one for the class-incremental lifelong learning detailed in Appendix \ref{experimental_setup} and Section \ref{continual_learning}, except from taking CIFAR-10 for class-incremental learning using ViT model with the increment batch of two classes. We trace the best and the worst matching and selected middle prototypes (according to the $\ell^2$ metric) through the stages of class-incremental learning. For the successful predictions, while the best matching prototypes tend to be constant, the worst matching ones change over time when the class changes.

\subsection{Impact of confounding on interpretations}
\label{confounding_interpretations}

The phenomenon of confounding takes its origin in causal modelling and is informally described, as per \cite{greenland1999confounding}, as \textit{'a mixing of effects of extraneous factors (called confounders) with the effects of interest'}. In many real-world scenarios, images contain confounding features, such as watermarks or naturally occurring spurious correlations ('seagulls always appear with the sea on the background'). The challenge for the interpretable models is therefore multi-fold: (1) these models need to be resistant to such confounders (2) should these confounders interfere with the performance of the model, the model should highlight them in the interpretations. 

To model confounding, we use the experimental setup from \cite{bontempelli2022concept}, which involves inpainting training images of three out of five selected classes of the CUB dataset with geometric figures (squares) which correlate with, but not caused by, the original data (e.g., every image of the \texttt{Crested Auklet} class is marked in the training data with a blue square). In Table \ref{tab:confounding_experiment}, we compare the experimental results between the original (\cite{wah2011caltech}) and confounded (\cite{bontempelli2022concept}) CUB dataset. We use the same original pre-trained feature spaces as  stated in Appendix \ref{experimental_setup}. The finetuned spaces are obtained through finetuning on confounded CUB data from \cite{bontempelli2022concept} for $15$ epochs. 

The results in Table \ref{tab:confounding_experiment} demonstrate clear advantage of models \textbf{without finetuning} on the confounded dataset for both $k$-means and $k$-means (nearest),  in the case of ViT. Such gap, however, is much narrower for $k$-means prototype selection, VGG-16 and ResNet-50. It is consistent with the results of Question 1 on the larger performance gap of these models compared with the ViT. Furthermore, $k$-means (nearest) does not show improvements over finetuning in a $k$-means (nearest) scenario, in a stark contrast with the ViT results.

\begin{figure}
\begin{subfigure}{0.49\textwidth}
\centering{
\resizebox{\textwidth}{!}{
\begin{tikzpicture}[domain=0:15]
\draw[decoration={brace,raise=2pt},decorate] (-4,5) -- node[above=3pt] {similar} (4,5); 
\node at (-3,4.7) {Black-footed A'ross};
\node at (-3,3.5) {\includegraphics[height=2cm]{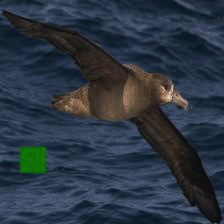}};
\node at (-3,2.25) {$\ell^2$: $39.120$};
\draw (-3,2) -- (0,1);

\node at (0,4.7) {Sooty A'ross};
\node at (0,3.5) {\includegraphics[height=2cm]{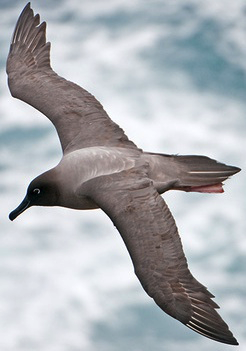}};
\node at (0,2.25) {$\ell^2$: $39.138$};
\draw (0,2) -- (0,1);

\node at (3,4.7) {Black-footed A'ross};
\node at (3,3.5) {\includegraphics[height=2cm]{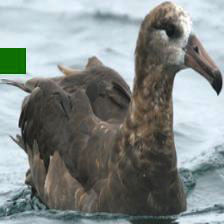}};
\node at (3,2.25) {$\ell^2$: $39.164$};
\draw (3,2) -- (0,1);

\node at (-2,0) {\Huge{?}};
\node at (-3,1) {\Huge{$\mathcal{\cdot}$}};
\node at (-3,0) {\Huge{$\mathcal{\cdot}$}};
\node at (-3,-1) {\Huge{$\mathcal{\cdot}$}};
\node at (2.1,0) {\huge{$\longrightarrow$} \textbf{\begin{tabular}{c} \small{''Black-footed} \\ \small{Albatross''} \end{tabular}}};

\node at (0,0) {\includegraphics[height=2cm]{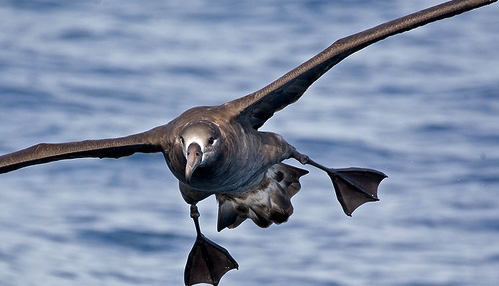}};

\node at (-3,-3.5) {\includegraphics[height=2cm]{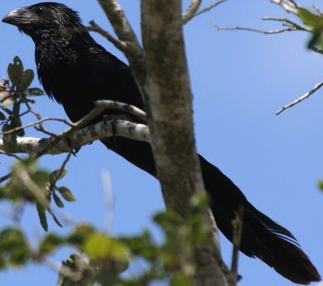}};
\draw (3,-2) -- (0,-1);
\node at (-3,-2.25) {$\ell^2$: $39.837$};
\node at (-3,-4.7) {Groove-billed Ani};

\node at (0,-3.5) {\includegraphics[height=2cm]{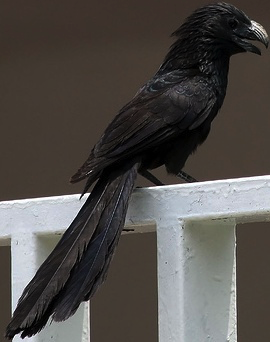}};
\draw (0,-2) -- (0,-1);
\node at (0,-2.25) {$\ell^2$: $39.888$};
\node at (0,-4.7) {Groove-billed Ani};

\node at (3,-3.5) {\includegraphics[height=2cm]{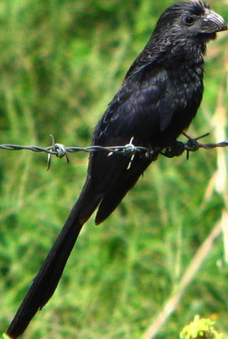}};
\draw (-3,-2) -- (0,-1);
\node at (3,-2.25) {$\ell^2$: $39.925$};
\node at (3,-4.7) {Groove-billed Ani};

\draw[decoration={brace,raise=2pt,mirror},decorate] (-4,-5) -- node[below=3pt]{dissimilar} (4,-5);
\end{tikzpicture}
}
}
\caption{Non-finetuned model interpretation (A'ross denotes 'Albatross')}
\end{subfigure}
\begin{subfigure}{0.49\textwidth}
\resizebox{\textwidth}{!}{
\begin{tikzpicture}[domain=0:15]
\draw[decoration={brace,raise=2pt},decorate] (-4,5) -- node[above=3pt] {similar} (4,5); 
\node at (-3,4.7) {Sooty Albatross};
\node at (-3,3.5) {\includegraphics[height=2cm]{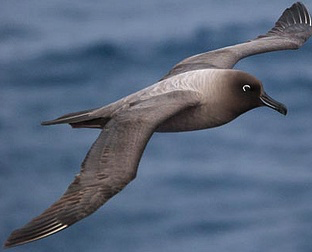}};
\node at (-3,2.25) {$\ell^2$: $39.182$};
\draw (-3,2) -- (0,1);

\node at (0,4.7) {Sooty Albatross};
\node at (0,3.5) {\includegraphics[height=2cm]{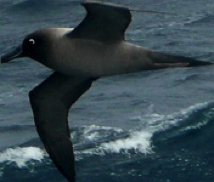}};
\node at (0,2.25) {$\ell^2$: $39.346$};
\draw (0,2) -- (0,1);

\node at (3,4.7) {Sooty Albatross};
\node at (3,3.5) {\includegraphics[height=2cm]{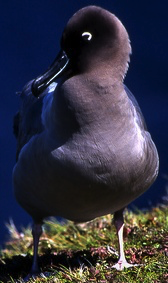}};
\node at (3,2.25) {$\ell^2$: $39.436$};
\draw (3,2) -- (0,1);

\node at (-2,0) {\Huge{?}};
\node at (-3,1) {\Huge{$\mathcal{\cdot}$}};
\node at (-3,0) {\Huge{$\mathcal{\cdot}$}};
\node at (-3,-1) {\Huge{$\mathcal{\cdot}$}};
\node at (2.1,0) {\huge{$\longrightarrow$ \textbf{\begin{tabular}{c} \small{''Black-footed} \\ \small{Albatross''} \end{tabular}}}};

\node at (0,0) {\includegraphics[height=2cm]{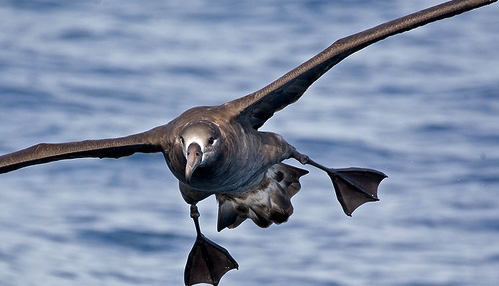}};

\node at (-3,-3.5) {\includegraphics[height=2cm]{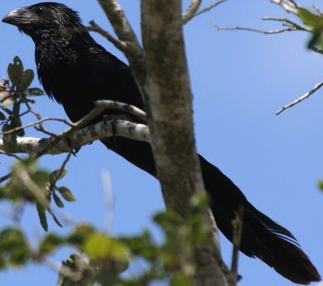}};
\draw (3,-2) -- (0,-1);x
\node at (-3,-2.25) {$\ell^2$: $39.802$};
\node at (-3,-4.7) {Groove-billed Ani};

\node at (0,-3.5) {\includegraphics[height=2cm]{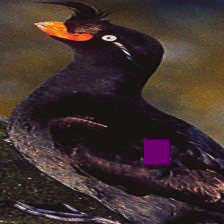}};
\draw (0,-2) -- (0,-1);
\node at (0,-2.25) {$\ell^2$: $39.834$};
\node at (0,-4.7) {Crested Auklet};

\node at (3,-3.5) {\includegraphics[height=2cm]{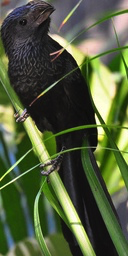}};
\draw (-3,-2) -- (0,-1);
\node at (3,-2.25) {$\ell^2$: $39.890$};
\node at (3,-4.7) {Groove-billed Ani};

\draw[decoration={brace,raise=2pt,mirror},decorate] (-4,-5) -- node[below=3pt]{dissimilar} (4,-5);
\end{tikzpicture}
}
\caption{Finetuned model interpretation\newline}
\end{subfigure}
\caption{Comparing the interpretations of the non-finetuned and finetuned model with confounding on confounded CUB (\cite{bontempelli2022concept}) dataset}
\label{confounding_interpretations_figure}
\end{figure}
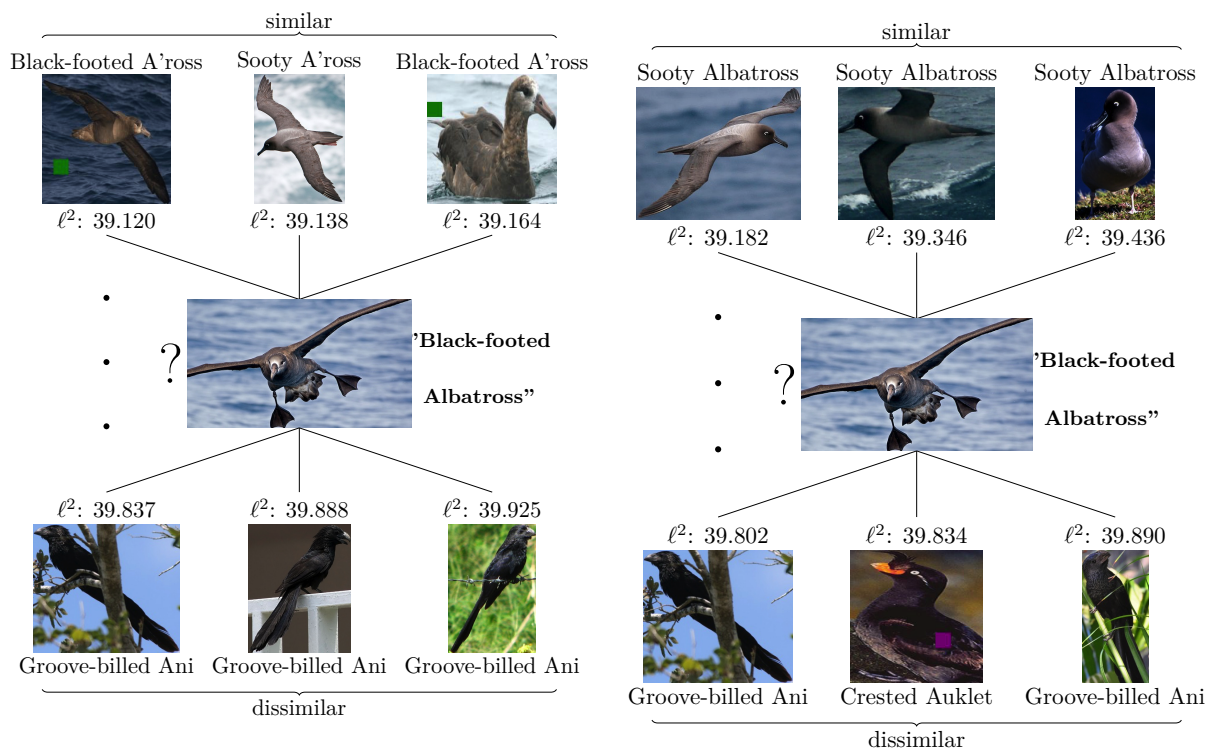

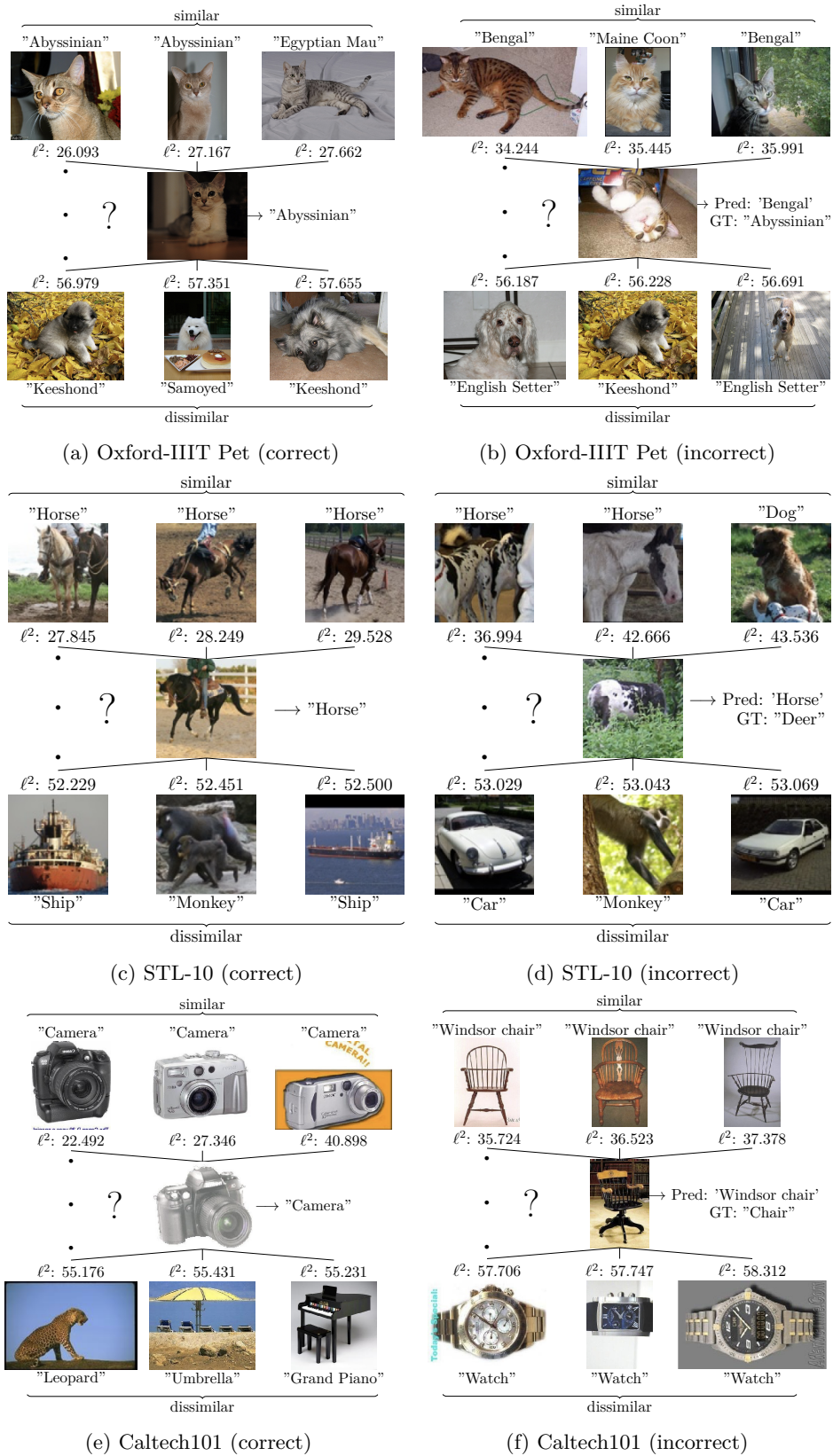
\begin{figure}
\begin{subfigure}{0.37\textwidth}
\centering{
\resizebox{\textwidth}{!}{
\begin{tikzpicture}[domain=0:15]
\draw[decoration={brace,raise=2pt},decorate] (-4,4.25) -- node[above=3pt] {similar} (4,4.25); 
\node at (-3,3.95) {''Abyssinian''};
\node at (-3,2.75) {\includegraphics[height=2cm]{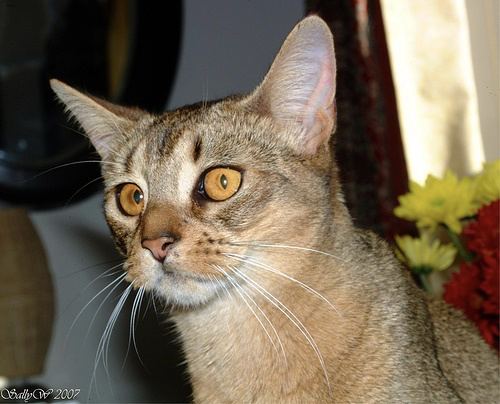}};
\node at (-3,1.5) {$\ell^2$: $26.093$};
\draw (-3,1.25) -- (0,1);

\node at (0,3.95) {''Abyssinian''};
\node at (0,2.75) {\includegraphics[height=2cm]{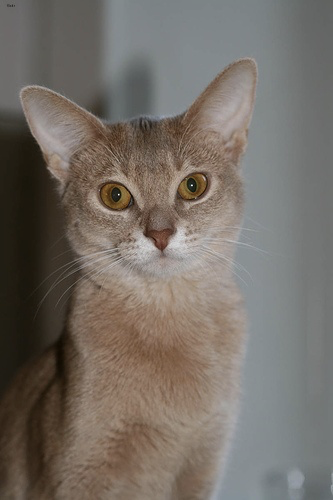}};
\node at (0,1.5) {$\ell^2$: $27.167$};
\draw (0,1.25) -- (0,1);

\node at (3,3.95) {''Egyptian Mau''};
\node at (3,2.75) {\includegraphics[height=2cm]{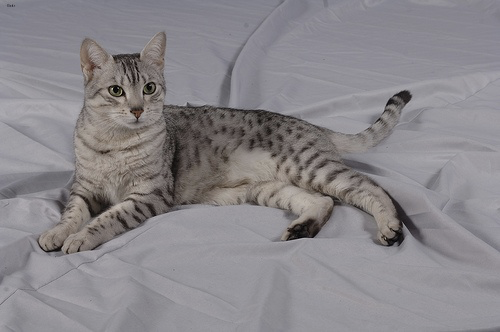}};
\node at (3,1.5) {$\ell^2$: $27.662$};
\draw (3,1.25) -- (0,1);

\node at (-2,0) {\Huge{?}};
\node at (-3,1) {\Huge{$\mathcal{\cdot}$}};
\node at (-3,0) {\Huge{$\mathcal{\cdot}$}};
\node at (-3,-1) {\Huge{$\mathcal{\cdot}$}};
\node at (2.3,0) {{$\longrightarrow$ ''Abyssinian''}};

\node at (0,0) {\includegraphics[height=2cm]{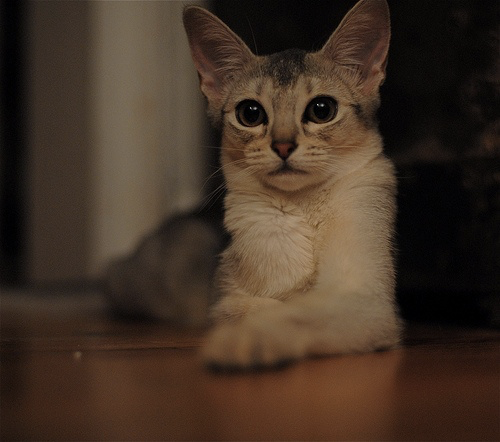}};

\node at (-3,-2.75) {\includegraphics[height=2cm]{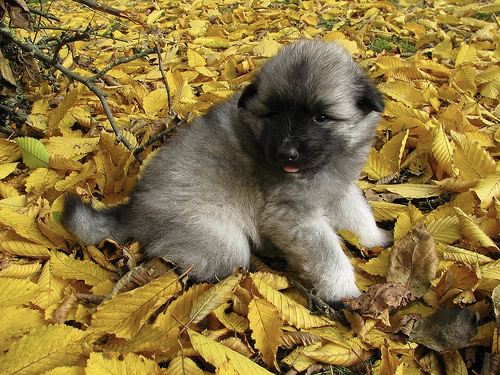}};
\draw (3,-1.25) -- (0,-1);
\node at (-3,-1.5) {$\ell^2$: $56.979$};
\node at (-3,-3.95) {''Keeshond''};

\node at (0,-2.75) {\includegraphics[height=2cm]{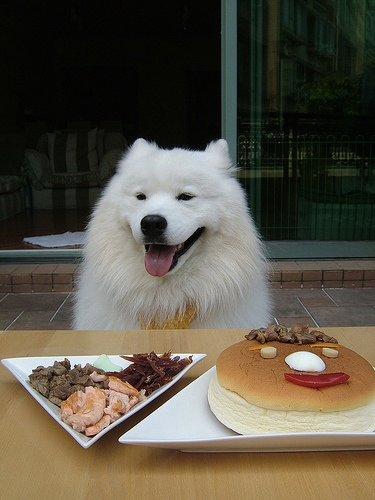}};
\draw (0,-1.25) -- (0,-1);
\node at (0,-1.5) {$\ell^2$: $57.351$};
\node at (0,-3.95) {''Samoyed''};

\node at (3,-2.75) {\includegraphics[height=2cm]{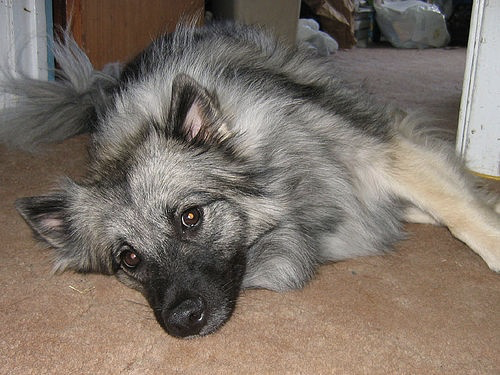}};
\draw (-3,-1.25) -- (0,-1);
\node at (3,-1.5) {$\ell^2$: $57.655$};
\node at (3,-3.95) {''Keeshond''};

\draw[decoration={brace,raise=2pt,mirror},decorate] (-4,-4.25) -- node[below=3pt]{dissimilar} (4,-4.25);
\end{tikzpicture}
}
\caption{Oxford-IIIT Pet (correct)}
\label{oxford_iiit_pets_correct_interpretation}
}
\end{subfigure}
\begin{subfigure}{0.39\textwidth}
\centering{
\resizebox{\textwidth}{!}{
\begin{tikzpicture}[domain=0:15]
\draw[decoration={brace,raise=2pt},decorate] (-4,4.25) -- node[above=3pt] {similar} (4,4.25); 
\node at (-3,3.95) {''Bengal''};
\node at (-3,2.75) {\includegraphics[height=2cm]{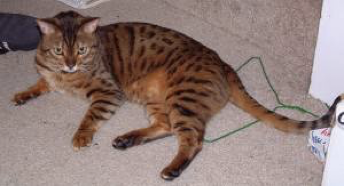}};
\node at (-3,1.5) {$\ell^2$: $34.244$};
\draw (-3,1.25) -- (0,1);

\node at (0,3.95) {''Maine Coon''};
\node at (0,2.75) {\includegraphics[height=2cm]{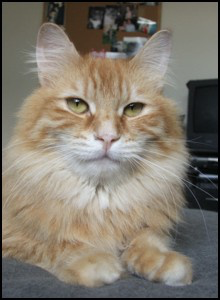}};
\node at (0,1.5) {$\ell^2$: $35.445$};
\draw (0,1.25) -- (0,1);

\node at (3,3.95) {''Bengal''};
\node at (3,2.75) {\includegraphics[height=2cm]{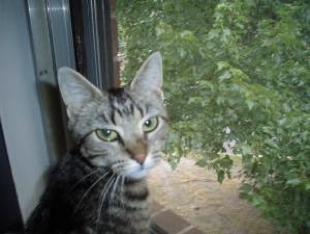}};
\node at (3,1.5) {$\ell^2$: $35.991$};
\draw (3,1.25) -- (0,1);

\node at (-2,0) {\Huge{?}};
\node at (-3,1) {\Huge{$\mathcal{\cdot}$}};
\node at (-3,0) {\Huge{$\mathcal{\cdot}$}};
\node at (-3,-1) {\Huge{$\mathcal{\cdot}$}};
\node at (2.5,0.2) {{$\longrightarrow$ Pred: 'Bengal'}};
\node at (3,-0.2) {{GT: ''Abyssinian''}};

\node at (0,0) {\includegraphics[height=2cm]{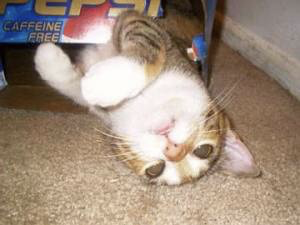}};

\node at (-3,-2.75) {\includegraphics[height=2cm]{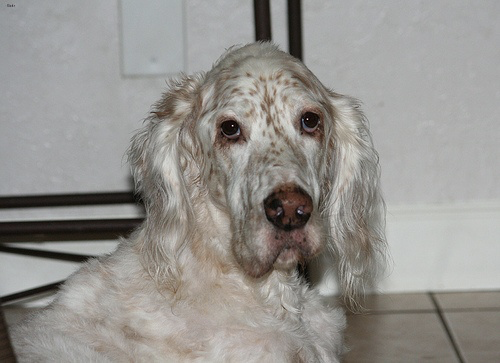}};
\draw (3,-1.25) -- (0,-1);
\node at (-3,-1.5) {$\ell^2$: $56.187$};
\node at (-3,-3.95) {''English Setter''};

\node at (0,-2.75) {\includegraphics[height=2cm]{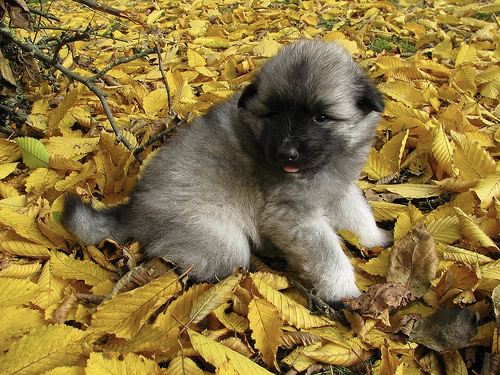}};
\draw (0,-1.25) -- (0,-1);
\node at (0,-1.5) {$\ell^2$: $56.228$};
\node at (0,-3.95) {''Keeshond''};

\node at (3,-2.75) {\includegraphics[height=2cm]{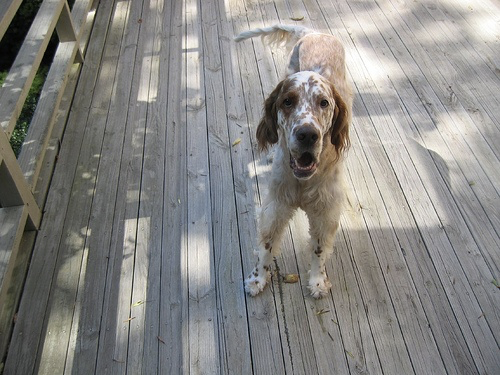}};
\draw (-3,-1.25) -- (0,-1);
\node at (3,-1.5) {$\ell^2$: $56.691$};
\node at (3,-3.95) {''English Setter''};

\draw[decoration={brace,raise=2pt,mirror},decorate] (-4,-4.25) -- node[below=3pt]{dissimilar} (4,-4.25);
\end{tikzpicture}
}

\caption{Oxford-IIIT Pet (incorrect)}}
\end{subfigure}
\begin{subfigure}{0.38\textwidth}
\centering{
\resizebox{\textwidth}{!}{
\begin{tikzpicture}[domain=0:15]
\draw[decoration={brace,raise=2pt},decorate] (-4,4.25) -- node[above=3pt] {similar} (4,4.25); 
\node at (-3,3.95) {''Horse''};
\node at (-3,2.75) {\includegraphics[height=2cm]{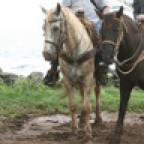}};
\node at (-3,1.5) {$\ell^2$: $27.845$};
\draw (-3,1.25) -- (0,1);

\node at (0,3.95) {''Horse''};
\node at (0,2.75) {\includegraphics[height=2cm]{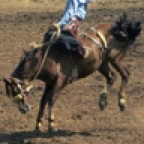}};
\node at (0,1.5) {$\ell^2$: $28.249$};
\draw (0,1.25) -- (0,1);

\node at (3,3.95) {''Horse''};
\node at (3,2.75) {\includegraphics[height=2cm]{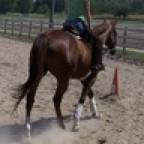}};
\node at (3,1.5) {$\ell^2$: $29.528$};
\draw (3,1.25) -- (0,1);

\node at (-2,0) {\Huge{?}};
\node at (-3,1) {\Huge{$\mathcal{\cdot}$}};
\node at (-3,0) {\Huge{$\mathcal{\cdot}$}};
\node at (-3,-1) {\Huge{$\mathcal{\cdot}$}};
\node at (2.3,0) {{$\longrightarrow$ ''Horse''}};

\node at (0,0) {\includegraphics[height=2cm]{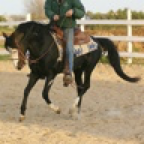}};

\node at (-3,-2.75) {\includegraphics[height=2cm]{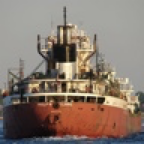}};
\draw (3,-1.25) -- (0,-1);
\node at (-3,-1.5) {$\ell^2$: $52.229$};
\node at (-3,-3.95) {''Ship''};

\node at (0,-2.75) {\includegraphics[height=2cm]{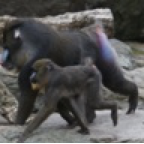}};
\draw (0,-1.25) -- (0,-1);
\node at (0,-1.5) {$\ell^2$: $52.451$};
\node at (0,-3.95) {''Monkey''};

\node at (3,-2.75) {\includegraphics[height=2cm]{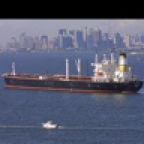}};
\draw (-3,-1.25) -- (0,-1);
\node at (3,-1.5) {$\ell^2$: $52.500$};
\node at (3,-3.95) {''Ship''};

\draw[decoration={brace,raise=2pt,mirror},decorate] (-4,-4.25) -- node[below=3pt]{dissimilar} (4,-4.25);
\end{tikzpicture}
}
\caption{STL-10 (correct)}}
\end{subfigure}
\begin{subfigure}{0.38\textwidth}
\centering{
\resizebox{\textwidth}{!}{
\begin{tikzpicture}[domain=0:15]
\draw[decoration={brace,raise=2pt},decorate] (-4,4.25) -- node[above=3pt] {similar} (4,4.25); 
\node at (-3,3.95) {''Horse''};
\node at (-3,2.75) {\includegraphics[height=2cm]{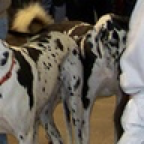}};
\node at (-3,1.5) {$\ell^2$: $36.994$};
\draw (-3,1.25) -- (0,1);

\node at (0,3.95) {''Horse''};
\node at (0,2.75) {\includegraphics[height=2cm]{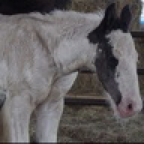}};
\node at (0,1.5) {$\ell^2$: $42.666$};
\draw (0,1.25) -- (0,1);

\node at (3,3.95) {''Dog''};
\node at (3,2.75) {\includegraphics[height=2cm]{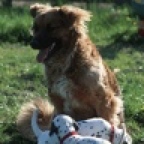}};
\node at (3,1.5) {$\ell^2$: $43.536$};
\draw (3,1.25) -- (0,1);

\node at (-2,0) {\Huge{?}};
\node at (-3,1) {\Huge{$\mathcal{\cdot}$}};
\node at (-3,0) {\Huge{$\mathcal{\cdot}$}};
\node at (-3,-1) {\Huge{$\mathcal{\cdot}$}};
\node at (2.5,0.2) {{$\longrightarrow$ Pred: 'Horse'}};
\node at (3,-0.2) {{GT: ''Deer''}};

\node at (0,0) {\includegraphics[height=2cm]{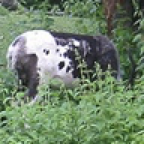}};

\node at (-3,-2.75) {\includegraphics[height=2cm]{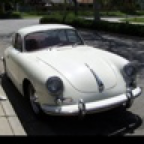}};
\draw (3,-1.25) -- (0,-1);
\node at (-3,-1.5) {$\ell^2$: $53.029$};
\node at (-3,-3.95) {''Car''};

\node at (0,-2.75) {\includegraphics[height=2cm]{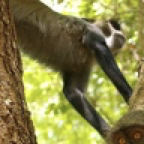}};
\draw (0,-1.25) -- (0,-1);
\node at (0,-1.5) {$\ell^2$: $53.043$};
\node at (0,-3.95) {''Monkey''};

\node at (3,-2.75) {\includegraphics[height=2cm]{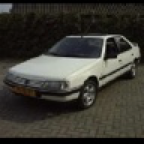}};
\draw (-3,-1.25) -- (0,-1);
\node at (3,-1.5) {$\ell^2$: $53.069$};
\node at (3,-3.95) {''Car''};

\draw[decoration={brace,raise=2pt,mirror},decorate] (-4,-4.25) -- node[below=3pt]{dissimilar} (4,-4.25);
\end{tikzpicture}
}
\caption{STL-10 (incorrect)}}
\end{subfigure}
\begin{subfigure}{0.37\textwidth}
\centering{
\resizebox{\textwidth}{!}{
\begin{tikzpicture}[domain=0:15]
\draw[decoration={brace,raise=2pt},decorate] (-4,4.25) -- node[above=3pt] {similar} (4,4.25); 
\node at (-3,3.95) {''Camera''};
\node at (-3,2.75) {\includegraphics[height=2cm]{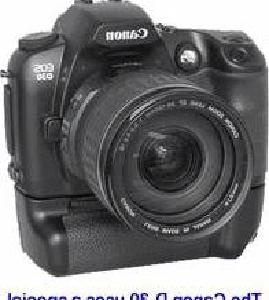}};
\node at (-3,1.5) {$\ell^2$: $22.492$};
\draw (-3,1.25) -- (0,1);

\node at (0,3.95) {''Camera''};
\node at (0,2.75) {\includegraphics[height=2cm]{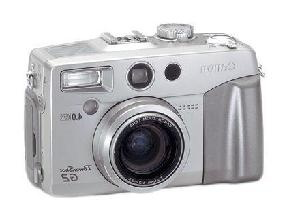}};
\node at (0,1.5) {$\ell^2$: $27.346$};
\draw (0,1.25) -- (0,1);

\node at (3,3.95) {''Camera''};
\node at (3,2.75) {\includegraphics[height=2cm]{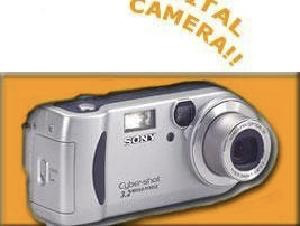}};
\node at (3,1.5) {$\ell^2$: $40.898$};
\draw (3,1.25) -- (0,1);

\node at (-2,0) {\Huge{?}};
\node at (-3,1) {\Huge{$\mathcal{\cdot}$}};
\node at (-3,0) {\Huge{$\mathcal{\cdot}$}};
\node at (-3,-1) {\Huge{$\mathcal{\cdot}$}};
\node at (2.3,0) {{$\longrightarrow$ ''Camera''}};

\node at (0,0) {\includegraphics[height=2cm]{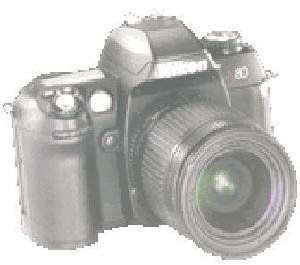}};

\node at (-3,-2.75) {\includegraphics[height=2cm]{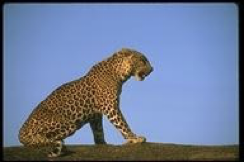}};
\draw (3,-1.25) -- (0,-1);
\node at (-3,-1.5) {$\ell^2$: $55.176$};
\node at (-3,-3.95) {''Leopard''};

\node at (0,-2.75) {\includegraphics[height=2cm]{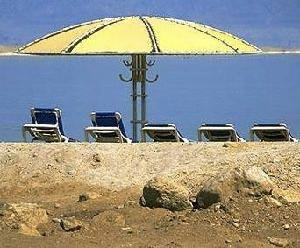}};
\draw (0,-1.25) -- (0,-1);
\node at (0,-1.5) {$\ell^2$: $55.431$};
\node at (0,-3.95) {''Umbrella''};

\node at (3,-2.75) {\includegraphics[height=2cm]{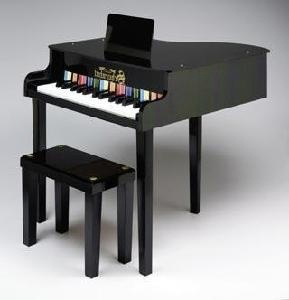}};
\draw (-3,-1.25) -- (0,-1);
\node at (3,-1.5) {$\ell^2$: $55.231$};
\node at (3,-3.95) {''Grand Piano''};

\draw[decoration={brace,raise=2pt,mirror},decorate] (-4,-4.25) -- node[below=3pt]{dissimilar} (4,-4.25);
\end{tikzpicture}
}
\caption{Caltech101 (correct)}
}
\end{subfigure}
\centering{
\begin{subfigure}{0.38\textwidth}
\resizebox{\textwidth}{!}{
\begin{tikzpicture}[domain=0:15]
\draw[decoration={brace,raise=2pt},decorate] (-4,4.25) -- node[above=3pt] {similar} (4,4.25); 
\node at (-3,3.95) {''Windsor chair''};
\node at (-3,2.75) {\includegraphics[height=2cm]{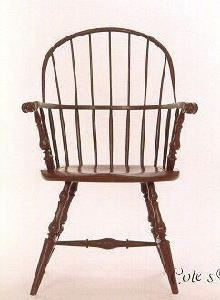}};
\node at (-3,1.5) {$\ell^2$: $35.724$};
\draw (-3,1.25) -- (0,1);

\node at (0,3.95) {''Windsor chair''};
\node at (0,2.75) {\includegraphics[height=2cm]{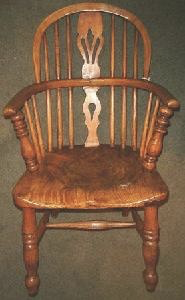}};
\node at (0,1.5) {$\ell^2$: $36.523$};
\draw (0,1.25) -- (0,1);

\node at (3,3.95) {''Windsor chair''};
\node at (3,2.75) {\includegraphics[height=2cm]{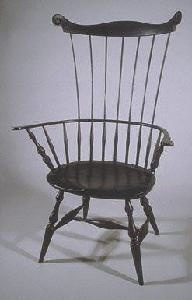}};
\node at (3,1.5) {$\ell^2$: $37.378$};
\draw (3,1.25) -- (0,1);

\node at (-2,0) {\Huge{?}};
\node at (-3,1) {\Huge{$\mathcal{\cdot}$}};
\node at (-3,0) {\Huge{$\mathcal{\cdot}$}};
\node at (-3,-1) {\Huge{$\mathcal{\cdot}$}};
\node at (2.5,0.2) {{$\longrightarrow$ Pred: 'Windsor chair'}};
\node at (3,-0.2) {{GT: ''Chair''}};

\node at (0,0) {\includegraphics[height=2cm]{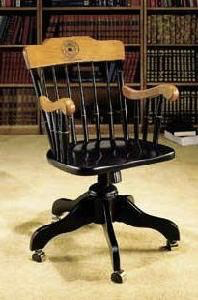}};

\node at (-3,-2.75) {\includegraphics[height=2cm]{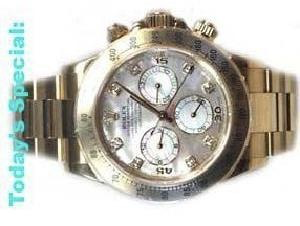}};
\draw (3,-1.25) -- (0,-1);
\node at (-3,-1.5) {$\ell^2$: $57.706$};
\node at (-3,-3.95) {''Watch''};

\node at (0,-2.75) {\includegraphics[height=2cm]{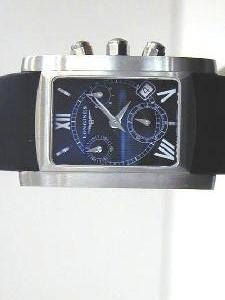}};
\draw (0,-1.25) -- (0,-1);
\node at (0,-1.5) {$\ell^2$: $57.747$};
\node at (0,-3.95) {''Watch''};

\node at (3,-2.75) {\includegraphics[height=2cm]{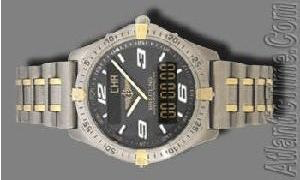}};
\draw (-3,-1.25) -- (0,-1);
\node at (3,-1.5) {$\ell^2$: $58.312$};
\node at (3,-3.95) {''Watch''};

\draw[decoration={brace,raise=2pt,mirror},decorate] (-4,-4.25) -- node[below=3pt]{dissimilar} (4,-4.25);
\end{tikzpicture}
}
\caption{Caltech101 (incorrect)}
\label{caltech101_incorrect_interpretation}
\end{subfigure}
}
\caption{Interpreting the predictions ($k$-means (nearest), OxfordIIITPets/STL-10/Caltech101, ViT)}
\label{visual_interpretability_ideal_3_offline}
\end{figure}

\begin{figure}
\centering
{
\resizebox{0.7\textwidth}{!}{
\begin{tikzpicture}[domain=0:15]

\node at (-0.5,-3) {\includegraphics[height=2cm]{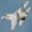}};
\node at (-0.5,-4.25) {''Plane''};

\draw[->] (1,-1) -- (1,-5);
\node[rotate=90] at (0.75,-3) {''Distance''};

\node at (3,2) {''2 classes''};

\node at (3,1) {\includegraphics[height=1cm]{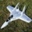}};
\node at (3,0.25) {Plane, $\ell^2$: $30.670$};

\node at (3,-0.5) {\includegraphics[height=1cm]{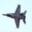}};
\node at (3,-1.25) {Plane, $\ell^2$: $30.790$};

\node at (3,-2) {\includegraphics[height=1cm]{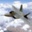}};
\node at (3,-2.75) {Plane, $\ell^2$: $31.063$};
\node at (3,-3.05) {\Large{$\ldots$}};
\node at (3,-3.8) {\includegraphics[height=1cm]{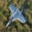}};
\node at (3,-4.5) {Plane, $\ell^2$: $32.026$};
\node at (3,-4.75) {\Large{$\ldots$}};
\node at (3,-5.5) {\includegraphics[height=1cm]{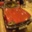}};
\node at (3,-6.25) {Automobile, $\ell^2$: $50.374$};

\node at (3,-7) {\includegraphics[height=1cm]{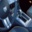}};
\node at (3,-7.75) {Automobile, $\ell^2$: $50.710$};

\node at (3,-8.5) {\includegraphics[height=1cm]{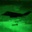}};
\node at (3,-9.25) {Plane, $\ell^2$: $51.601$};

\node at (7,2) {''6 classes''};

\node at (7,1) {\includegraphics[height=1cm]{interpretation_example_3/inc_sim_1_2}};
\node at (7,0.25) {Plane, $\ell^2$: $30.670$};

\node at (7,-0.5) {\includegraphics[height=1cm]{interpretation_example_3/inc_sim_2_2}};
\node at (7,-1.25) {Plane, $\ell^2$: $30.790$};

\node at (7,-2) {\includegraphics[height=1cm]{interpretation_example_3/inc_sim_3_2}};
\node at (7,-2.75) {Plane, $\ell^2$: $31.063$};
\node at (7,-3.05) {\Large{$\ldots$}};
\node at (7,-3.8) {\includegraphics[height=1cm]{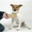}};
\node at (7,-4.5) {Dog, $\ell^2$: $37.949$};
\node at (7,-4.75) {\Large{$\ldots$}};

\node at (7,-5.5) {\includegraphics[height=1cm]{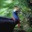}};
\node at (7,-6.25) {Bird, $\ell^2$: $51.614$};

\node at (7,-7) {\includegraphics[height=1cm]{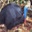}};
\node at (7,-7.75) {Bird, $\ell^2$: $51.738$};

\node at (7,-8.5) {\includegraphics[height=1cm]{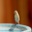}};
\node at (7,-9.25) {Bird, $\ell^2$: $52.579$};

\node at (11,2) {''10 classes''};

\node at (11,1) {\includegraphics[height=1cm]{interpretation_example_3/inc_sim_1_2}};
\node at (11,0.25) {Plane, $\ell^2$: $30.670$};

\node at (11,-0.5) {\includegraphics[height=1cm]{interpretation_example_3/inc_sim_2_2}};
\node at (11,-1.25) {Plane, $\ell^2$: $30.790$};

\node at (11,-2) {\includegraphics[height=1cm]{interpretation_example_3/inc_sim_3_2}};
\node at (11,-2.75) {Plane, $\ell^2$: $31.063$};

\node at (11,-3.05) {\Large{$\ldots$}};
\node at (11,-3.8) {\includegraphics[height=1cm]{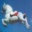}};
\node at (11,-4.5) {Horse, $\ell^2$: $35.329$};

\node at (11,-4.7) {\Large{$\ldots$}};
\node at (11,-5.4) {\includegraphics[height=1cm]{interpretation_example_3/37.94950188329306_5}};
\node at (11,-6.1) {Dog, $\ell^2$: $37.949$};

\node at (11,-6.35) {\Large{$\ldots$}};
\node at (11,-7) {\includegraphics[height=1cm]{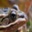}};
\node at (11,-7.75) {Frog, $\ell^2$: $52.010$};

\node at (11,-8.5) {\includegraphics[height=1cm]{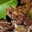}};
\node at (11,-9.25) {Frog, $\ell^2$: $52.034$};

\node at (11,-10) {\includegraphics[height=1cm]{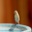}};
\node at (11,-10.75) {Bird, $\ell^2$: $52.579$};
\end{tikzpicture}
}
}
\caption{CIFAR-10 continual learning: evolution of prototype ranking}
\label{continual_learning_ranking_evolution}
\end{figure}

We demonstrate the interpretations for the confounding experiment in Figure \ref{confounding_interpretations_figure}. While the non-finetuned model successfully predicts the correct confounded class, black-footed albatross, the finetuned model fails at this scenario and predicts a similar class Sooty Albatross, which does not containt the confounder mark.

On the other hand, the finetuned model performs similarly or better on the original (not confounded) data. These results further build upon the hypothesis from Question 2 and demonstrate that the use of the proposed framework can help address the phenomenon of confounding.

\begin{table}[]
    \centering
    \begin{tabular}{|c|c|c|c|c|}\hline
         Feature space & Prototype selection & VGG16 & ResNet-50 & ViT \\\hline
        \multicolumn{5}{c}{Confounded data (\cite{bontempelli2022concept})} \\ \hline 
        Finetuned & N/A, backbone network	& $73.99\pm 2.91$ & $70.42\pm 2.68$ & $69.06\pm 4.40$\\ \hline\hline
        Non-finetuned & $k$-means	& $\mathbf{78.52 \pm 1.31}$ & $\mathbf{76.68\pm 1.63}$ & $\mathbf{80.70\pm 2.26}$ \\ 
         Finetuned & $k$-means	& $73.19\pm 1.43$ & $67.16\pm 2.25$ & $66.58\pm 5.81$\\ \hline\hline
         Non-finetuned & $k$-means (nearest)	& $64.13\pm 1.37$ & $67.68\pm 0.90$ &$\mathbf{82.88\pm 2.17}$\\ 
         Finetuned & $k$-means (nearest)	& $\mathbf{71.00\pm 2.92}$ & $69.03\pm 1.19$ & $73.99\pm 5.19$\\ \hline
         \hline\hline
         \multicolumn{5}{c}{Original data} \\ \hline 
        Finetuned & N/A, backbone network	& $83.66\pm 1.16$ & $83.49\pm 1.22$ & $93.92\pm 1.31$ \\ \hline\hline
        Non-finetuned & $k$-means	& $80.01\pm 1.27$ & $80.10\pm 1.66$ & $90.67\pm 1.13$ \\ 
         Finetuned & $k$-means	& $81.98\pm 1.53$ & $79.38\pm 2.87$  & $92.85\pm 1.70$\\ \hline\hline
         Non-finetuned & $k$-means (nearest)	&   $72.11\pm 1.62$ & $72.64\pm 1.87$ & $88.57\pm 0.96$\\ 
         Finetuned & $k$-means (nearest)	& $\mathbf{78.90\pm 2.77}$ & $\mathbf{80.05\pm 2.64}$ & $\mathbf{92.80\pm 1.77}$\\ \hline
    \end{tabular}
    \caption{F1 score comparison for CUB dataset (\cite{wah2011caltech}), confidence interval calculated over five runs; all $k$-means runs are for $10\%$ (15) clusters/prototypes; the better results within its category are highlighted in bold, taking into account the confidence interval. While for the original data finetuning has strong performance benefits, non-finetuned model has an edge over the finetuned one for all architectures; for $k$-means (nearest) the non-finetuned model still performs clearly better with ViT architecture than the finetuned counterpart. }
    \label{tab:confounding_experiment}
\end{table}

\section{Conclusion}
The proposed IDEAL framework considers separately the representations from the latent spaces, learnt on generic large data sets, and learning of an interpretable, prototype-based model within this data space. We confirm an initial intuition that, in offline learning setting, contemporary ViT models drastically narrow the gap between the finetuned and non-finetuned models (Question 1). We justify the architectural choices for the framework such as selection of prototypes (Question 1) and demonstrate the margin of overfitting for finetuned ViTs (Question 2). This insight enables us to demonstrate that the proposed framework can surpass the state-of-the-art class-incremental learning methods (Question 3). We demonstrate interpretations through prototypes provided by the framework in offline and class-incremental learning scenarios  (Question 4). Finally (Question 5), we demonstrate that in non-causal confounding scenarios, for modern architectures, such as ViT, finetuning results in both inferior performance and interpretations.

\section*{Broader Impact Statement}
The proposed approach goes beyond the paradigm of first training and then finetuning complex models to the new tasks, which is standard for the field, where both these stages of the approach use expensive GPU compute to improve the model performance. We show that contemporary architectures, trained with extensive data sets, can deliver competitive performance in a lifelong learning setting even without such expensive finetuning. This can deliver profound impact on democratisation of high-performance machine learning models and implementation on Edge devices, on board of autonomous vehicles, as well as address important problems of environmental sustainability by avoiding using much energy to train new latent representations and finetune, providing instead a way to re-use existing models. Furthermore, the proposed framework can help define a benchmark on how deep-learning latent representations generalise to new tasks. 

This approach also naturally extends to class- and potentially, domain-incremental learning, enabling learning new concepts. It demonstrates that with large and complex enough latent spaces, relatively simple strategies of prototype selection, such as clustering, can deliver results comparable with the state-of-the-art in a fraction of time and compute efforts. Importantly, unlike most of the state-of-the-art approaches, as described in the Related work section of the main paper, the proposed framework directly provides interpretability in linguistic and visual form and provides improved resistance to spurious correlations in input features. 

\section*{Limitations}
This work does not aim for explaining the latent spaces of the deep-learning architecture; instead, it explores explainable-through-prototypes decision making process in terms of similarity to the prototypes in the latent space.

\section*{Acknowledgement}
This work is supported by ELSA – European Lighthouse on Secure and Safe AI funded by the European Union under grant agreement No. 101070617. Views and opinions expressed are however those of the author(s) only and do not necessarily reflect those of the European Union or European Commission.Neither the European Union nor the European Commission can be held responsible.

The computational experiments have been powered by a High-End Computing (HEC) facility of Lancaster University, delivering high-performance and high-throughput computing for research within and across departments.

\bibliography{main}
\bibliographystyle{tmlr}

\appendix
\section{Experimental setup}
\label{experimental_setup}

In this work, all the experiments were conducted in PyTorch 2.0.0. The pre-trained models used in these experiments were obtained from TorchVision \footnote{https://pytorch.org/vision/main/models.html} while the finetuned models have been obtained from three different sources:
\begin{enumerate}
    \item \textit{Models that come from MMPreTrain \footnote{https://github.com/open-mmlab/mmpretrain}}. Specifically, ResNet50 and ResNet101 finetuned on the CIFAR-10, and ResNet 50 finetuned on CIFAR-100.
\item \textit{finetuned TorchVision models}. finetuning was conducted by continuing the EBP across all network layers. Such models include VGG-16 and Vision Transformer (ViT) finetuned on CIFAR-10, as well as ResNet101, VGG-16, and ViT finetuned on CIFAR-100. For ResNet101 and VGG-$16$ models, we ran the training for $200$ epochs, while the Vision Transformer models were trained for $10$ epochs. The Stochastic Gradient Descent (SGD) optimizer was employed for all models, with a learning rate of $0.0005$ and a momentum value of $0.9$.
\item \textit{Linearly finetuned TorchVision models}. In such case, only the linear classifier was trained and all the remaining layers of the network were fixed. For these models, we conducted training for $200$ epochs for ResNet50, ResNet101, and VGG16, and 25 epochs for the ViT models. We adopted the Stochastic Gradient Descent (SGD) optimizer, with a learning rate of 0.001 and a momentum parameter set at 0.9.
\end{enumerate}

We utilized $k$-means clustering and random selection methods, setting the number of prototypes for each class at 10\% of the training data for the corresponding classes. Besides, we also set it to 12 per class and conducted experiments for ResNet50, ResNet101, and VGG-16 on CIFAR-10 and CIFAR-100 datasets, enabling us to evaluate the impact of varying the number of prototypes. 

For ELM online clustering method, we experimented with varying radius values for each specific dataset and backbone network. We selected a radius value that would maintain the number of prototypes within the range of 0-20\% of the training data. In the experiments without finetuning on the CIFAR-10 dataset, we set the radius to 8, 10, 19, and 12 for ResNet50, ResNet101, VGG-16, and Vision Transformer (ViT) models respectively. The radius was adjusted to 8, 11, 19, and 12 for these models when conducting the same tasks without finetuning on CIFAR-100. For STL10, Oxford-IIIT Pets, EuroSAT, and CalTech101 datasets, the radius was set to 13 across all ELM experiments. In contrast, the xDNN model did not require hyper-parameter settings as it is inherently a parameter-free model.

We performed all experiments for Sections  \ref{classification_section} and \ref{continual_learning} of the main paper $5$ times and report mean values and standard deviations for our results, with the exception of the finetuned backbone models where we just performed finetuning once (or sourced finetuned models as detailed above).  The class-incremental learning experiments in Section  \ref{continual_learning} are performed using $k$-means.

The class-incremental lifelong learning experiments (see Figure $11$ of the main paper)  were executed $10$ times to allow a robust comparison with benchmark results. 

To ensure a consistent and stable training environment, for every experiment we used a single NVIDIA V100 GPU from a cluster.

\section{Complete experimental results}
\label{Complete_experimental_results}
Tables \ref{tab:cifar10_classification_no_fine-tuning}-\ref{tab:CalTech101_classification} contain extended experimental results  for multiple benchmarks and feature extractors. These results further demonstrate the findings of the main paper. 

Table \ref{tab:cifar10_classification_no_fine-tuning} demonstrates the data behind Figures 3, 4, 5 of the main paper. It also highlights the performance of the $k$-means model on ViT-L latent space, when the nearest real training data point to the $k$-means cluster centre is selected (labelled as $k$-means (nearest)). One can also see that even with the small number of selected prototypes, the algorithm delivers competitive performance without finetuning. 

Table \ref{tab:cifar10_classification_fine-tuning} compares different latent spaces and gives the number of free (optimised) parameters for the scenario of finetuning of the models. With a small additional number of parameters, which is the number of possible prototypes, one can transform the opaque architectures into ones interpretable through proximity and similarity to prototypes within the latent space (this is highlighted in the interpretability column). 

Tables \ref{tab:cifar100_classification_non_fine-tuned}-\ref{tab:CalTech101_classification} repeat the same analysis, expanded from Figure 5 of the main paper for different data sets. The results show remarkable consistency with the previous conclusions and further back up the claims of generalisation to  different classification tasks.

\begin{table}[]
    \centering
    \begin{tabular}{|c|c|c|c|c|}\hline
         FE & method & accuracy (\%) & \#prototypes & time, s  \\\hline\hline
         \multirow{5}{*}{\rotatebox{90}{\textsc{ResNet50}}} & random & $65.55\pm 1.93$ & $120 (0.24\%)$& $85$  \\
         &random & $80.40\pm 0.37$ & $5,000 (10\%)$ & $85$\\
         &ELM &  $81.17\pm 0.04$ & $5,500 (11\%)$ & $365$ \\
         &xDNN  & $81.44 \pm 0.33$  & $115(0.23\%)$ & $103$ \\
         &$k$-means  & $84.12\pm 0.19$ & $120(0.24\%)$ & $201$ \\
         &$k$-means & $\mathbf{86.65\pm 0.15}$ & $5,000(10\%)$ & $1,138$\\\hline\hline
         \multirow{5}{*}{\rotatebox{90}{\textsc{Resnet101}}} &random & $78.08 \pm 1.38$ & $120(0.24\%)$ & $129$ \\
         &random & $87.66 \pm 0.25$ & $5,000(10\%)$ & $129$ \\
         &ELM & $88.22\pm 0.09$ & $7,154 (14.31\%)$ & $524$\\
         &xDNN & $88.13\pm 0.42$ & $118(0.24\%)$ & $145$\\
         &$k$-means & $90.19 \pm 0.15$ & $120(0.24\%)$ &  $245$ \\
         &$k$-means & $\mathbf{91.50 \pm 0.07}$ & $5,000(10\%)$ & $1,194$\\\hline\hline
         \multirow{5}{*}{\rotatebox{90}{\textsc{VGG-16}}} & random & $50.13\pm 2.37$ & $120(0.24\%)$ & $95$\\       
         &random & $65.06\pm 0.32$ & $5,000(10\%)$ & $95$ \\
         &ELM & $72.31\pm 0.08$ & $1,762 (3.52\%)$ & $215$\\
         &xDNN & $70.03\pm 0.96$ & $103(0.21\%)$ & $132$ \\
         &$k$-means & $74.48\pm 0.16$ & $120(0.24\%$ & $346$\\
         &$k$-means & $\mathbf{75.94\pm 0.15}$ & $5,000(10\%)$ & $2,362$ \\\hline\hline
         \multirow{3}{*}{\rotatebox{90}{\textsc{ViT}}} & random & $93.23\pm 0.11$ & $5,000(10\%)$ & $597$ \\
         &ELM & $90.61\pm 0.14$  & $6,685 (13.37\%)$ & $889$ \\
         &xDNN  & $93.59\pm 0.12$ & $112(0.2\%)$ & $606$\\
         &$k$-means & $\mathbf{95.59 \pm 0.08}$ & $5,000(10\%)$ & $925$\\\hline \multirow{2}{*}{{\textsc{ViT-L}}} & $k$-means & $\mathbf{96.48\pm 0.05}$ & $5,000(10\%)$ & $4,375$ \\
         &$k$-means (nearest) & $95.62 \pm 0.07$ & $5,000(10\%)$ & $4,352$ \\\hline 
    \end{tabular}
    \caption{CIFAR-10 classification task comparison for the case of no finetuning of the feature extractor}
    \label{tab:cifar10_classification_no_fine-tuning}
\end{table}
\begin{table}[]
    \begin{tabular}{|c|c|c|c|c|c|c|}\hline
         FE & method & accuracy (\%) & \#parameters & \#prototypes & time, s & interpretability \\\hline\hline
         \multirow{6}{*}{\rotatebox{90}{\textsc{ResNet50}}} & ResNet50 & $95.55$ $(80.71^*)$ & $\sim 25M$ $(20K)$ & &$36,360$ $(13,122^*)$ & \xmark \\
         & random & $94.92\pm 0.02$ &$\sim 25M+50K$ & $120(0.24\%)$ &$36,360+24$ & \cmark  \\
         & random & $95.32\pm 0.09$ &$\sim 25M+50K$  & $5,000(10\%)$ & $36,360+24$&\cmark \\
         & xDNN  & $95.32\pm 0.12$ & $\sim 25M+50K$ & $111(0.22\%)$ & $36,360+43$&  \cmark\\
         & $k$-means & $94.91\pm 0.14$ &$\sim 25M+50K$ & $120(0.24\%)$ & $36,360+208$ & \cmark\\
         & $k$-means & $95.50\pm 0.06$ & $\sim 25M+50K$ & $5,000(10\%)$ & $36,360+1,288$& \cmark\\\hline\hline
         \multirow{6}{*}{\rotatebox{90}{\textsc{ResNet101}}} & Resnet101 & $95.58$ $(84.44^*)$ & $\sim44M$ $(20K)$ & & $36,360$ & \xmark\\
         & random & $95.47\pm 0.06$ & $\sim44M+50K$ & $120 (0.24\%)$ & $36,360+37$ & \cmark\\
         & random & $95.51\pm 0.01$ & $\sim44M+50K$  & $5,000 (10\%)$ & $36,360+37$& \cmark\\
         & xDNN & $95.50\pm 0.10$ & $\sim44M+50K$  & $107 (0.21\%)$ & $36,360+54$&\cmark\\
         & $k$-means & $95.55\pm 0.03$ & $\sim44M+50K$  & $120 (0.24\%)$ & $36,360+231$ &\cmark \\
         & $k$-means & $95.51\pm 0.04$ & $\sim44M+50K$  & $5,000 (10\%)$  & $36,360+1,357$&\cmark \\\hline\hline
         \multirow{6}{*}{\rotatebox{90}{\textsc{VGG-16}}} & VGG-16 & $92.26$ $(83.71^*)$ & $\sim 138M$ $(41K)$ &  & $40,810$ &\xmark \\
         & random & $87.48\pm 0.72$ & $\sim 138M+50K$ & $120(0.24\%)$& $40,810 + 94$&\cmark \\        
         & random & $90.86\pm 0.19$ & $\sim 138M+50K$ & $5,000 (10\%)$ & $40,810 + 94$ & \cmark\\
         & xDNN & $91.42\pm 0.25$ & $\sim 138M+50K$ & $102 (0.20\%)$ & $40,810 + 123$ & \cmark\\
         & $k$-means & $92.24 \pm 0.10$ & $\sim 138M+50K$ & $120(0.24\%)$ &$40,810 + 369$ & \cmark\\
         & $k$-means & $92.55 \pm 0.16$ & $\sim 138M+50K$ & $5,000(10\%)$ & $40,810 + 2,408$ & \cmark \\\hline\hline
         \multirow{6}{*}{\rotatebox{90}{\textsc{ViT}}} &  \textsc{ViT} & $98.51$ $(96.08^*)$ & $\sim 86M$ $(8K)$ &  & $15,282$ $(15,565^*)$ &\xmark \\
         & random & $98.56\pm0.02$  & $\sim 86M+50K$ & $5,000(10\%)$ & $15,282$ + 598&\cmark \\
         & xDNN & $98.00 \pm 0.14$  & $\sim 86M+50K$ & $117(0.23\%)$ & $15,282+607$& \cmark\\
         & $k$-means & $98.53\pm 0.04$  & $\sim 86M+50K$ & $5,000(10\%)$ &$15,282+938$ &\cmark\\\hline    
    \end{tabular}
    \caption{CIFAR-10 classification task comparison for the case of finetuned models ($*$ denotes linear finetuning of the DL model)}
    \label{tab:cifar10_classification_fine-tuning}
\end{table}

\begin{table}[]
    \centering
    \begin{tabular}{|c|c|c|c|c|}\hline
         FE& method & accuracy (\%) & \#prototypes & time, s  \\\hline\hline
         \multirow{5}{*}{\rotatebox{90}{\textsc{ResNet50}}} & random & $41.66\pm 0.74$ &  $1,200(2.4\%)$ & $82$\\ 
         & random & $54.37 \pm 0.43$ &  $10,000(20\%)$ & $82$\\
         & ELM & $57.94 \pm 0.11$ &  $7,524(15.05\%)$  & $129$\\
         & xDNN & $58.25 \pm 0.64$&  $884(1.77\%)$ & $98$\\
         & $k$-means & $62.67\pm 0.26$  & $1,200(2.4\%)$ & $124$ \\
         & $k$-means & $64.07\pm 0.37$  & $10,000(20\%)$ & $258$\\\hline\hline
        \multirow{5}{*}{\rotatebox{90}{\textsc{ResNet101}}} & random & $50.25\pm0.71$  & $1,200(2.4\%)$ & $128$\\
         & random & $61.90\pm0.41$ & $10,000(20\%)$& $128$\\
         & ELM & $64.42 \pm 0.12$ & $4,685(9.37\%)$ & $161$ \\
         & xDNN & $64.60\pm 0.39$ &   $878(1.76\%)$& $143$\\
         & $k$-means & $68.59\pm 0.40$ & $1,200(2.4\%)$ & $170$\\
         & $k$-means & $70.04\pm 0.12$ & $10,000(20\%)$ & $310$\\\hline\hline 
         \multirow{4}{*}{\rotatebox{90}{\textsc{VGG16}}} & random  & $26.16 \pm 0.24$ & $1,200(2.4\%)$  & $94$\\     
         & random & $37.74 \pm 0.48$ & $10,000(20\%)$ & $94$\\
         & ELM & $48.53 \pm 0.05$ &   $2,878(5.76\%)$ & $122$\\
         & xDNN & $47.78\pm 0.41$ & 871 ($1.74\%) $ & $119$ \\  
         & $k$-means & $51.99 \pm 0.24$ & $1,200 (2.4\%)$ & $175$\\
         & $k$-means & $52.55 \pm 0.27$ & $1,200 (2.4\%)$ & $437$\\\hline\hline  
         \multirow{3}{*}{\rotatebox{90}{\textsc{ViT}}}  & random & $72.39 \pm 0.21$ & $10,000(20\%)$ & $604$\\  
         & ELM & $69.94 \pm 0.06$ &  $8,828(17.66\%)$  & $642$ \\
         & xDNN  & $76.24\pm 0.24$ & $830(1.66\%)$ & $613$\\     
         & $k$-means & $79.12 \pm 0.28$ & $10,000(20\%)$ &$673$\\\hline  
        \multirow{2}{*}{{\textsc{ViT-L}}}  & $k$-means & $\mathbf{82.18 \pm 0.14}$ & $10,000(20\%)$ & $3,905$  \\    
         & $k$-means (nearest)  & $78.75\pm 0.29$ & $10,000(20\%)$ & $3,909$\\\hline    
    \end{tabular}
    \caption{CIFAR-100 classification task comparison for the case of no finetuning of the feature extractor}
    \label{tab:cifar100_classification_non_fine-tuned}
\end{table}

\begin{table}[]
    \centering
    \begin{tabular}{|c|c|c|c|c|c|c|c|}\hline
         FE & method & accuracy (\%)  & \#parameters & \#prototypes &time, s & interpretability \\\hline\hline
         \multirow{6}{*}{\rotatebox{90}{\textsc{ResNet50}}}  & ResNet50 & $79.70$ $(56.39^*)$& $\sim 25M$ $(205K)$  &  & $36,360(13,003^*)$ & \xmark \\
         & random & $78.94 \pm 0.17 $& $\sim 25M+50K$  &  $1,200(2.4\%)$ & $36,360+28$& \cmark \\
         & random & $79.52 \pm 0.17 $& $\sim 25M+50K$  &  $10,000(20\%)$ & $36,360+28$& \cmark \\
         & xDNN & $79.75 \pm 0.12$ & $\sim 25M+50K$ &  $859(1.72\%)$ & $36,360+45$& \cmark  \\
          & $k$-means & $79.84 \pm 0.07$  & $\sim 25M+50K$ & $1,200(2.4\%)$ & 36,360+82 & \cmark \\
         & $k$-means & $79.77 \pm 0.07$  & $\sim 25M+50K$ & $10,000(20\%)$ & 36,360+219 & \cmark \\\hline\hline
         \multirow{6}{*}{\rotatebox{90}{\textsc{ResNet101}}}  & ResNet50 & $84.38$ $(63.18^*)$& $\sim 44M$ $(205K)$  &  & $45,619(18,955^*)$ & \xmark \\
         & random & $82.26 \pm 0.15 $& $\sim 44M+50K$  &  $1,200(2.4\%)$ & $45,619+175$& \cmark \\
         & random & $80.75 \pm 0.19 $& $\sim 44M+50K$  &  $10,000(20\%)$ & $45,619+175$& \cmark \\
         & xDNN & $81.13 \pm 0.16$ & $\sim 44M+50K$ &  $831(1.66\%)$ & $45,619+191$& \cmark  \\
          & $k$-means & $83.03 \pm 0.06$  & $\sim 44M+50K$ & $1,200(2.4\%)$ & $45,619+220$ & \cmark \\
         & $k$-means & $83.14 \pm 0.19$  & $\sim 44M+50K$ & $10,000(20\%)$ & $45,619+439$ & \cmark \\\hline\hline
         \multirow{6}{*}{\rotatebox{90}{\textsc{VGG-16}}}  & VGG-16 & $75.08$ $(62.74^*)$& $\sim 138M$ $(410K)$  &  & $41,038(17,098^*)$ & \xmark \\
         & random & $53.83 \pm 0.91 $& $\sim 138M+50K$  &  $1,200(2.4\%)$ & $41,038+92$& \cmark \\
         & random & $64.17 \pm 0.36 $& $\sim 138M+50K$  &  $10,000(20\%)$ & $41,038+92$& \cmark \\
         & xDNN & $72.63 \pm 0.11$ & $\sim 138M+50K$ &  $907(1.81\%)$ & $41,038+120$ & \cmark  \\
          & $k$-means & $73.83 \pm 0.16$  & $\sim 138M+50K$ & $1,200(2.4\%)$ & $41,038+199$ & \cmark \\
         & $k$-means & $73.73 \pm 0.23$  & $\sim 138M+50K$ & $10,000(20\%)$ & $41,038+460$ & \cmark \\\hline\hline
          \multirow{4}{*}{\rotatebox{90}{\textsc{ViT}}} & ViT & $90.29(82.79^*)$ & $\sim 86M$ $(77K)$ & & $15,536(15,423^*)$& \xmark \\ 
          & random & $89.90 \pm 0.10$ & $\sim 86M+50K$ & $10,000(20\%)$ & $15,536+621$& \cmark  \\      
          & xDNN & $89.17 \pm 0.18$ & $\sim 86M+50K$ & $809(1.61\%)$ & $15,536+630$&\cmark\\     
          & $k$-means & $90.48\pm 0.05$ & $\sim 86M+50K$ & $10,000(20\%)$ & $15,536+695$&\cmark\\\hline        
    \end{tabular}
    \caption{CIFAR-100 classification task comparison for the case of finetuned models ($*$ denotes linear finetuning of the DL model)}
    \label{tab:cifar100_classification_fine-tuned}
\end{table}

\begin{table}[]
    \centering
    \begin{tabular}{|c|c|c|c|c|c|}\hline
         FE & method & accuracy (\%) & \#prototypes & time, s \\\hline\hline
         \multirow{3}{*}{\rotatebox{90}{\textsc{ViT}}} & random & $98.55\pm 0.09$ &  $500 (10\%)$ & $61$ \\
         & ELM  & $95.27 \pm 0.03$ & $271 (5.42\%)$  & $63$ \\
         & xDNN  & $98.63 \pm 0.12$ &  $84(1.68\%)$ & $62$\\
         & $k$-means & $99.32\pm 0.03$ & $500(10\%)$ & $65$\\\hline\hline 
         \multirow{2}{*}{{\textsc{ViT-L}}} & $k$-means & $99.71\pm 0.02$ &  $500 (10\%)$ & $377$ \\
         & $k$-means(nearest)  & $99.56 \pm 0.05$ &  $500(10\%)$ & $377$ \\\hline   
    \end{tabular}
    \caption{STL10 classification task comparison for the case of no finetuning (linear finetuning of the ViT gives $98.97\%$)}
    \label{tab:stl10_classification}
\end{table}

\begin{table}[]
    \centering
    \begin{tabular}{|c|c|c|c|c|c|}\hline
         FE & method & accuracy (\%)  & \#prototypes & time, s \\\hline\hline
         \multirow{3}{*}{\rotatebox{90}{\textsc{ViT}}} & random & $90.82\pm 0.53$ &  $365(9.92\%)$ & $48$ \\ 
         & ELM & $90.85 \pm 0.03$ & $122 (3.32\%)$ & $49$\\
         & xDNN & $96.30 \pm 0.23$ & $239(6.49\%)$ & $49$ \\
         & $k$-means & $94.07\pm 0.20$  & $365(9.92\%)$ & $50$\\\hline\hline
         \multirow{2}{*}{{\textsc{ViT-L}}} & $k$-means & $95.78\pm 0.19$ &  $365(9.92\%)$ & $279$ \\ 
         & $k$-means (nearest) & $94.76\pm 0.30$  & $740(9.92\%)$ & $279$\\\hline     
    \end{tabular}
    \caption{OxfordIIITPets classification task comparison for the case of no finetuning (linear finetuning of ViT gives $94.41\%$)}
    \label{tab:OxfordIIITPets_classification}
\end{table}

\begin{table}[]
    \centering
        \begin{tabular}{|c|c|c|c|c|c|}\hline
         FE & method & accuracy (\%) & \#prototypes & time, s  \\\hline
         \multirow{3}{*}{\rotatebox{90}{\textsc{ViT}}} & random & $82.67\pm 0.54$ &  $2,154(9.97\%)$ & $266$\\
         & ELM & $83.69 \pm 0.01$ &  $528(2.44\%)$   & $277$ \\
         & xDNN & $85.24 \pm 1.05$ &   $102(0.47\%)$ & $269$ \\
         & $k$-means & $91.30\pm 0.16$  & $2,154(9.97\%)$ & $330$\\\hline\hline
         \multirow{2}{*}{{\textsc{ViT-L}}} & $k$-means & $88.93\pm 0.22$ &  $2,154(9.97\%)$ & $1685$ \\
         & $k$-means(nearest) & $83.97 \pm 0.16$ &  $2,154(9.97\%)$ & $1685$\\\hline      
    \end{tabular}
    \caption{EuroSAT classification task comparison for the case of no finetuning (linear finetuning gives $95.17\%$)}
    \label{tab:EuroSAT_classification}
\end{table}
\begin{table}[]
    \centering
        \begin{tabular}{|c|c|c|c|c|c|}\hline
         FE & method & accuracy (\%) & \#prototypes & time, s \\\hline
        \multirow{3}{*}{\rotatebox{90}{\textsc{ViT}}} & random & $89.42 \pm 0.32$ &  $649(9.35\%)$ & $96$ \\
         & ELM  & $91.12 \pm 0.07$ &  $516(7.43\%)$  & $97$ \\
         & xDNN  & $94.61 \pm 0.94$ &   $579(8.34\%)$ & $97$ \\
         & $k$-means & $94.46\pm 0.44$  & $649(9.35\%)$ & $99$ \\\hline\hline
         \multirow{2}{*}{{\textsc{ViT-L}}} & $k$-means & $96.08 \pm 0.34$ &  $649(9.35\%)$ & $515$\\
         & $k$-means (nearest) & $93.74\pm 0.42$  & $649(9.35\%)$ & $517$\\\hline   
    \end{tabular}
    \caption{CalTech101 classification task comparison  (linear finetuning gives $96.26\%$)}
    \label{tab:CalTech101_classification}
\end{table}

\section{Sensitivity analysis for the number of prototypes}
\label{sensitivity_number_of_prototypes}
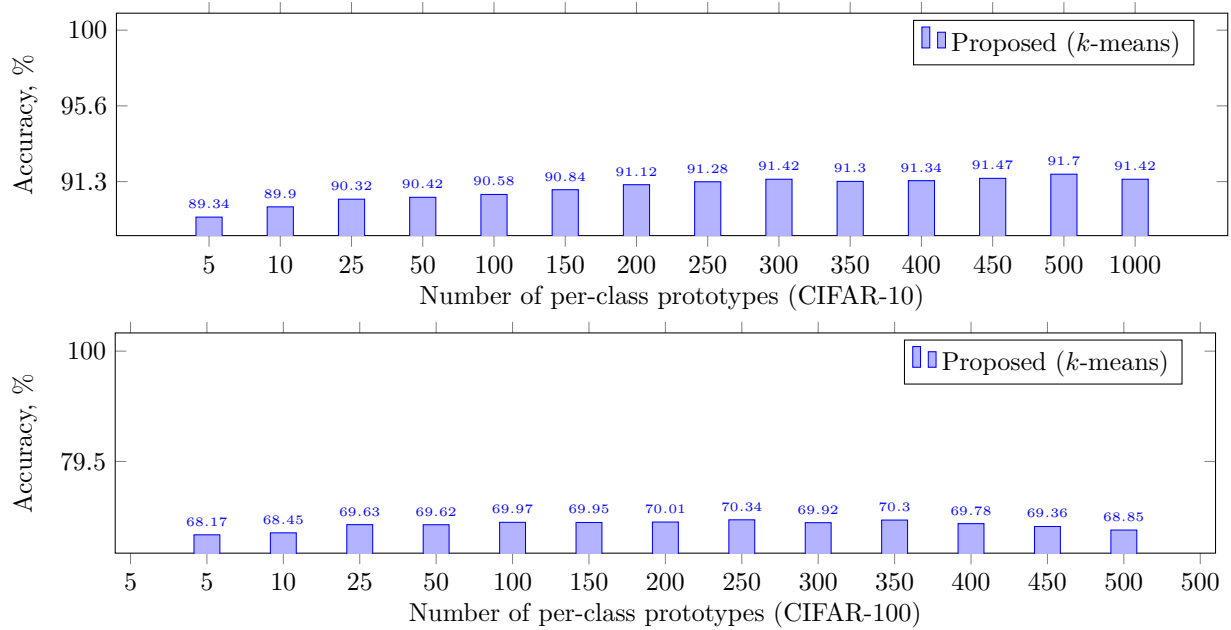
\begin{figure}
    \centering
    
\resizebox{\textwidth}{!}{
\begin{tikzpicture}[domain=0:4]
\begin{axis}[legend pos=north east, ymode=log,log ticks with fixed point,ymin=0,ymax=100,enlargelimits=true,xlabel={Number of per-class prototypes (CIFAR-10)},point meta=rawy, symbolic x coords={5,10,25,50,100,150,200, 250, 300,350,400,450,500,1000},nodes near coords, nodes near coords style={font=\tiny}, xtick = data,
    ylabel={Accuracy, $\%$},ybar=2*\pgflinewidth,width=\textwidth, height=0.2\textheight]
  \addplot coordinates {
                (5,89.34) (10,89.90) (25,90.32) (50,90.42)(100,90.58) (150,90.84) (200,91.12) (250,91.28) (300,91.42) (350,91.30) (400,91.34) (450,91.47) (500,91.70) (1000,91.42)  
            };
 \addlegendentry{Proposed ($k$-means)};
 \addlegendentry{ResNet101};
\end{axis}
\end{tikzpicture}
}
\resizebox{\textwidth}{!}{
\begin{tikzpicture}[domain=0:4]
\begin{axis}[legend pos=north east, ymode=log,log ticks with fixed point,ymin=0,ymax=100,point meta=rawy, symbolic x coords={5,10,25,50,100,150,200,250,300,350,400,450,500},nodes near coords, nodes near coords style={font=\tiny},enlargelimits=true,xlabel={Number of per-class prototypes (CIFAR-100)},ylabel={Accuracy, $\%$},
nodes near coords,ybar=2*\pgflinewidth,width=\textwidth, height=0.2\textheight]
  \addplot coordinates {
                (5,68.17) (10,68.45) (25,69.63) (50,69.62)(100,69.97) (150,69.95) (200,70.01) (250,70.34) (300,69.92) (350,70.30) (400,69.78) (450,69.36) (500,68.85)   
            };
 \addlegendentry{Proposed ($k$-means)}
\end{axis}
\end{tikzpicture}
}

    \caption{Accuracy sensitivity to the number of per-class prototypes ($k$-means, ResNet101, no finetuning)}
    \label{fig:performance_sensitivity_number_of_per_class_prototypes}
\end{figure}

Figure \ref{fig:performance_sensitivity_number_of_per_class_prototypes} further backs up the previous evidence that even with a small number of prototypes, the accuracy is still high. It shows, however, that there is a trade-off between the number of prototypes and accuracy.  It also shows, that after a few hundred prototypes per class on CIFAR-10 and CIFAR-100 tasks, the performance does not increase and may even slightly decrease, indicating saturation. 

\section{Linguistic interpretability of the proposed framework outputs}
\label{linguistic_interpretability_of_IDEAL_outputs}

To back up interpretability claim, we present two additional interpretability scenarios complementing the one in Figure 12 of the main text.

\begin{figure}
\centering{
\resizebox{\textwidth}{!}{
\begin{tikzpicture}[domain=0:15]

\node at (0,0) {IF $\ \left(\begin{array}{l}
\\
Q \sim\\
\\
 \end{array}\right.$ \adjustbox{raise=-2.5ex}{\includegraphics[height=1cm]{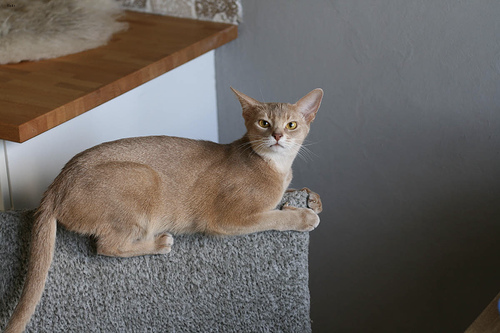}} {$ \left.\begin{array}{l}
\\
\\
\\
 \end{array}\right)$\ OR$\ \left(\begin{array}{l}
\\
Q \sim\\
\\
 \end{array}\right.$ \adjustbox{raise=-2.5ex}{\includegraphics[height=1cm]{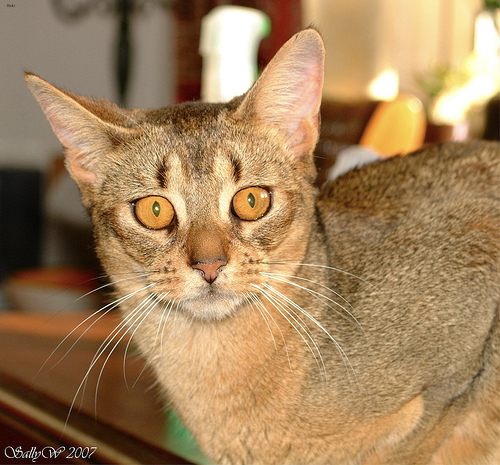}} {$ \left.\begin{array}{l}
\\
\\
\\
 \end{array}\right)$}\ OR$\ \left(\begin{array}{l}
\\
Q \sim\\
\\
 \end{array}\right.$ \adjustbox{raise=-2.5ex}{\includegraphics[height=1cm]{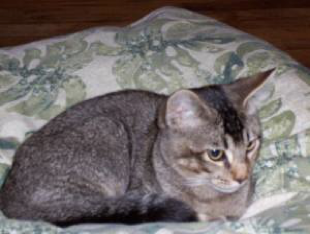}} {$ \left.\begin{array}{l}
\\
\\
\\
 \end{array}\right)$}\  \ THEN 'Abyssinian'}};

\node at (0,-1.5) {IF $\ \left(\begin{array}{l}
\\
Q \sim\\
\\
 \end{array}\right.$ \adjustbox{raise=-2.5ex}{\includegraphics[height=1cm]{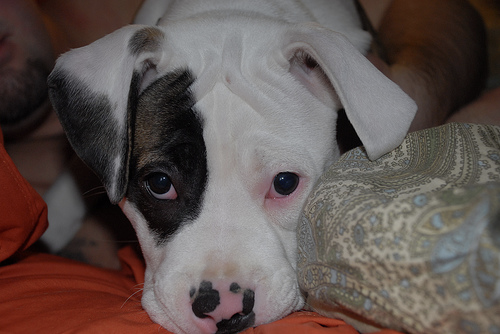}} {$ \left.\begin{array}{l}
\\
\\
\\
 \end{array}\right)$\ OR$\ \left(\begin{array}{l}
\\
Q \sim\\
\\
 \end{array}\right.$ \adjustbox{raise=-2.5ex}{\includegraphics[height=1cm]{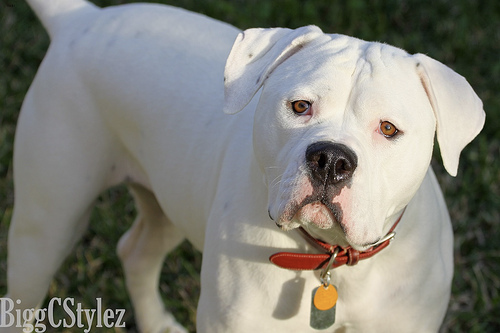}} {$ \left.\begin{array}{l}
\\
\\
\\
 \end{array}\right)$}\ OR$\ \left(\begin{array}{l}
\\
Q \sim\\
\\
 \end{array}\right.$ \adjustbox{raise=-2.5ex}{\includegraphics[height=1cm]{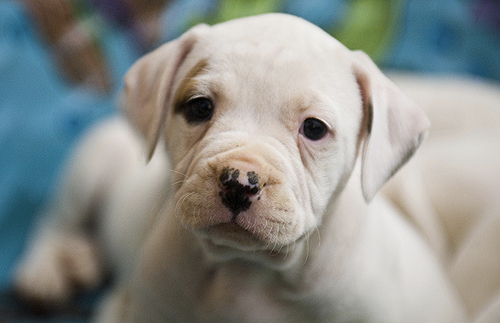}} {$ \left.\begin{array}{l}
\\
\\
\\
 \end{array}\right)$}\  \ THEN 'American Bulldog'}};

\end{tikzpicture}
}
}  
    \caption{An example of symbolic decision rules (OxfordIIITPets), $Q$ denotes the query image}
    \label{fig:symbolic_decision_rules}
\end{figure}

First, we show the symbolic decision rules in  Figure \ref{fig:symbolic_decision_rules}. These symbolic rules are created using ViT-L backbone, with the prototypes selected using the nearest real image to $k$-means cluster centroids, in a no-finetuning scenario for OxfordIIITPets dataset.

\begin{figure}
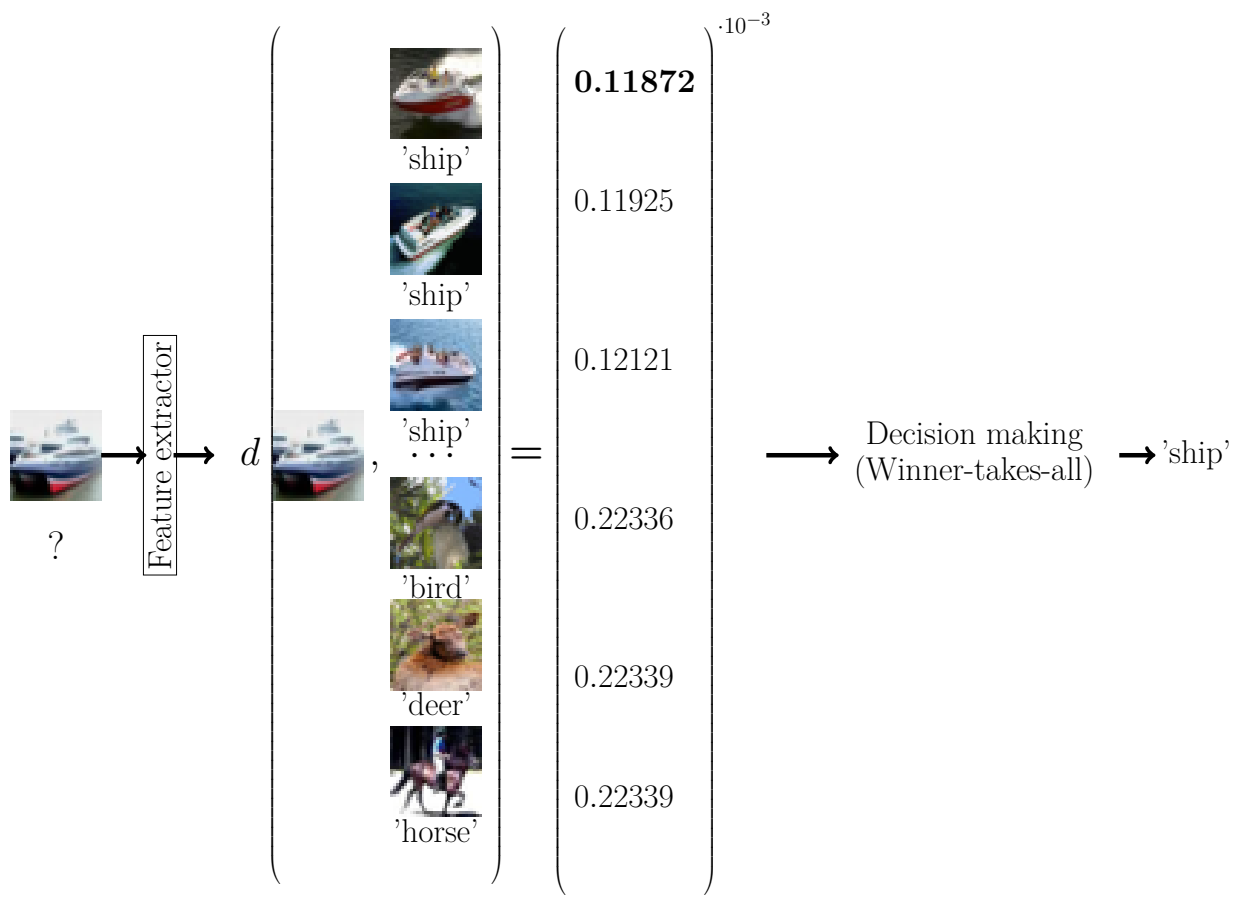

\centering{
\resizebox{\textwidth}{!}{
\begin{tikzpicture}[domain=0:15]
\node at (-1,0) {\includegraphics[width=2cm]{interpretation_example/1.png}};
\node at (-1,-2) {\Huge{?}};
\draw[->, line width=1mm] (0,0) -- (1,0);
\node at (1.3,0) [rotate=90,draw] {\huge{Feature extractor}};
\draw[->, line width=1mm] (1.6,0) -- (2.5,0);
\node at (6,0)  {\Huge{$d\left(\ \ \cdots\ , \begin{array}{l}
\\
 \\
\\
\\
\\
 \\
\\
 \\
\\
 \\
 \\
\\
\\
 \\
 \\
\\
 \\
\\
 \end{array}\cdots\ \ \ \right)$}};
\node at (7.4,8) {\includegraphics[width=2cm]{interpretation_example/39278.png}};
\node at (7.4,5) {\includegraphics[width=2cm]{interpretation_example/15416.png}};
\node at (7.4,2) {\includegraphics[width=2cm]{interpretation_example/19357.png}};
\node at (7.4,-1.5) {\includegraphics[width=2cm]{interpretation_example/19287.png}};
\node at (7.4,-4.2) {\includegraphics[width=2cm]{interpretation_example/28509.png}};
\node at (7.4,-7) {\includegraphics[width=2cm]{interpretation_example/14561.png}};

\node at (4.8,0) {\includegraphics[width=2cm]{interpretation_example/1.png}};

\node at (9.4,0) {\Huge{$ \textbf{ = } $}};
\node at (12.4,0)  {\huge{$\left(\begin{array}{l}
\\
\textbf{0.11872}\\
\\
\\
0.11925 \\
\\
\\
\\
0.12121\\
\\
\\
 \\
0.22336\\
 \\
 \\
 \\
0.22339 \\
\\
\\
0.22339 \\
 \\
\\
 \end{array}\right)^{\cdot 10^{-3}}$}};
 \draw[->, line width=1mm] (14.7,0) -- (16.3,0);`
\node at (19.3,0)  {\huge{\shortstack{Decision making \\(Winner-takes-all)}}};
 \draw[->, line width=1mm] (22.5,0) -- (23.3,0);
\node at (24.2,0)  {\huge{'ship'}};
  
\node at (7.4,6.5) {\huge{'ship'}};
\node at (7.4,3.5) {\huge{'ship'}};
\node at (7.4,0.5) {\huge{'ship'}};
\node at (7.4,-2.9) {\huge{'bird'}};
\node at (7.4,-5.5) {\huge{'deer'}};
\node at (7.4,-8.3) {\huge{'horse'}};

\end{tikzpicture}
}
}
\caption{Interpreting the model predictions ($k$-means (nearest), 500 clusters per class, CIFAR-10, ViT)}
\label{fig:visual_interpretability_ideal}
\end{figure}

Second, in Figure  \ref{fig:visual_interpretability_ideal} we show how the overall pipeline of the proposed method can be summarised in interpretable-through-prototypes fashion. We show the normalised distance obtained through dividing by the sum of distances to all prototypes. This is to improve the perception and give relative, bound between $0$ and $1$, numbers for the prototype images.

\end{document}